\documentclass[letterpaper,twocolumn,10pt]{article}
\usepackage{usenix}

\usepackage{tikz}
\usepackage{amsmath}

\usepackage{filecontents}

\date{}
\usepackage{times}
\usepackage{soul}
\usepackage{adjustbox}
\usepackage{url}
\usepackage{amsmath}
\usepackage{amsthm}
\usepackage{booktabs}
\usepackage{algorithm}
\usepackage{algorithmic}
\usepackage{multirow}
\usepackage[switch]{lineno}
\usepackage{placeins}
\usepackage{algorithm}
\usepackage{algorithmic}
\usepackage{dsfont}
\usepackage{array}
\usepackage{adjustbox}
\usepackage{rotating} 
\usepackage{caption}
\usepackage{array}
\usepackage{float}
\usepackage{makecell}
\usepackage{subcaption}
\usepackage[table]{xcolor}   
\definecolor{oursblue}{RGB}{210,235,255}
\definecolor{lightgray}{gray}{0.92}
\newcolumntype{Y}{c}

\begin{document}

\title{\Large \bf Single-Sample Black-Box Membership Inference Attack against 
\\Vision-Language Models via Cross-modal Semantic Alignment}

\author{
Jiaqing Li$^{1}$, 
Yajuan Lu$^{1}$, 
Xiaochuan Shi$^{1}$, 
Gang Wu$^{2}$, 
ZhongYuan Wang$^{1}$, 
Chao Liang$^{1}$ \\
$^{1}$Wuhan University \\
$^{2}$Tarim University
} 

\maketitle
\begin{abstract}
   Vision-Language Models (VLMs) have achieved remarkable success, yet their reliance on massive datasets and unintended memorization of training data raise significant data security risk. Membership Inference Attacks (MIAs) aim to assess these risks by determining whether a data sample was included in a model’s training set. However, existing MIA methods against VLMs face critical bottlenecks: gray-box method relies on internal logits that are typically restricted in real-world Application Programming Interfaces (APIs), while black-box method depends on large-scale statistical distributions, which struggle in single-sample scenarios. To this end, we investigate MIAs from the perspective of cross-modal semantic alignment, and observe that member images exhibit significantly stronger image-caption alignment due to training memorization, whereas generated captions for non-members may deviate from the original visual content. Leveraging this insight, we propose a novel MIA framework designed for strict black-box and single-sample setting that quantifies such alignment within a joint embedding space, thereby bypassing these unrealistic assumptions. We conducted extensive experiments on three open-source and two closed-source VLMs. On the VL-MIA/Flicker dataset, our method achieves an AUC of 0.821 against LLaVA-1.5, significantly outperforming existing baselines. Furthermore, it remains robust under diverse image perturbations, highlighting its practicality.
\end{abstract}         
\section{Introduction}
\label{sec:intro}

Vision-Language Models (VLMs)~\cite{bai2025qwen3vltechnicalreport,liu2023visual,wang2024cogvlmvisualexpertpretrained} have gained significant prominence due to their cross-modal reasoning prowess. They exhibit remarkable generalization across diverse tasks, including zero-shot classification~\cite{saha2024improved}, visual question answering~\cite{Hartsock_2024,dai2023instructblipgeneralpurposevisionlanguagemodels}, and image-based knowledge extraction~\cite{hu2024bliva}. Yet their formidable capabilities are rooted in an extensive reliance on massive training data, which often contains unauthorized information and thereby introduces a series of critical privacy and security risks~\cite{Carlini0EKS19,Mireshghallah0Z24,CarliniHNJSTBIW23,devlin-etal-2019-bert}. For instance, in domains involving sensitive personal data, such as healthcare~\cite{hu2022m}, VLMs risk inadvertently exposing private patient details learned during training.

Membership Inference Attacks (MIAs), pioneered by Shokri~\cite{shokri2017membership}, is initially developed to assess privacy risks in classification models by discerning whether a specific data sample is utilized during training. Early MIAs commonly adopt shadow training~\cite{shokri2017membership, liu2021encodermi}, where attackers train shadow models to approximate the target model's behavior and simulate the difference between member and non-member samples. The outputs of these shadow models are labeled according to sample membership and used to train a binary membership inference classifier. While effective for simple models, this paradigm cannot scale to complex models like Large Language Models (LLMs)~\cite{openai2024gpt4technicalreport
,touvron2023llamaopenefficientfoundation,zhao2025surveylargelanguagemodels,10.5555/3495724.3495883,chiang2023vicuna} and VLMs, because the immense parameter count and vast data requirements make shadow training prohibitively expensive~\cite{duan2024do,HuYYLHZDY24}. To address the above issue, recent MIAs against LLMs rely solely on the outputs of the target model for membership inference, such as output token probabilities or log-likelihoods~\cite{zhang2025mink,shi2024detecting,ren2025self,mattern2023membership}.

Building upon such efforts, a limited number of studies explore membership inference for visual data in VLMs. Existing approaches primarily leverage generated image captions to conduct attacks, yet they still suffer from two major limitations. The first limitation arises from the strong dependence on gray-box assumptions. Following prior work on MIAs against LLMs~\cite{shi2024detecting,mattern2023membership}, Li \textit{et al.}~\cite{li2024membership} construct discriminative statistics by accessing the logits of generated image captions. However, in real-world VLM applications, prediction evidence such as logits is typically inaccessible to users. This makes logit-based MIA methods impractical in strict black-box scenarios.

\begin{figure}[t!]
     \centering
     \includegraphics[width=\linewidth]{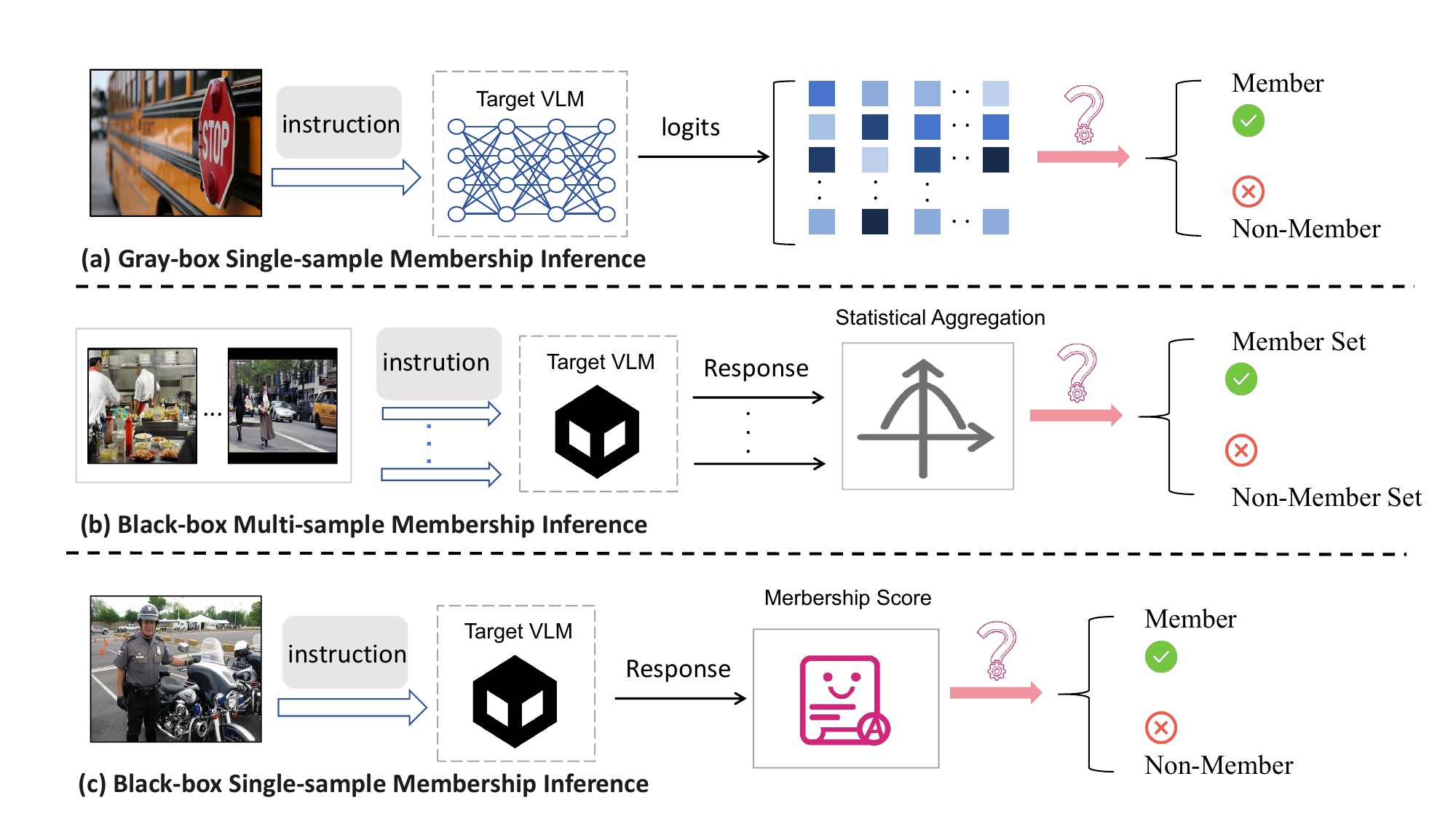} 
  \caption{The difference among 3 scenarios.}
  \label{fig:duibi}
\end{figure}

The second limitation is that existing black-box MIA methods against VLMs struggle to maintain effectiveness under limited sample conditions. 
For example, Hu \textit{et al.}~\cite{hu2025membership} propose a dataset-level membership inference approach under a strict black-box setting. 
Given a collection of images, their method relies on unimodal textual signals and infers membership by evaluating the semantic consistency of generated captions aggregated at the dataset level. Although effective when a sufficiently large dataset is available, the attack suffers severe performance degradation in few-shot or single-sample scenarios, where the lack of a reliable dataset-level statistical reference renders it impractical for black-box MIA applications.

In this paper, we investigate a more challenging and realistic scenario:
\textit{conducting MIA on a single-sample solely from the VLM's textual response under black-box conditions}. As illustrated in Figure \ref{fig:duibi}, we distinguish MIA against VLMs in this setting from other scenarios: gray-box attacks typically rely on model logits, while black-box multi-sample MIA requires a collection of multiple samples for inference. In contrast, the black-box single-sample setting represents the most stringent and restrictive condition, reflecting typical interactions with deployed VLM APIs where the attacker has access only to model outputs and few input samples. 
Studying membership inference under this setting highlights the inherent challenges of performing MIA with minimal information and demonstrates the practical significance of evaluating privacy risks in black-box scenarios.

\begin{figure}[t]
     \centering
     \includegraphics[width=\linewidth]{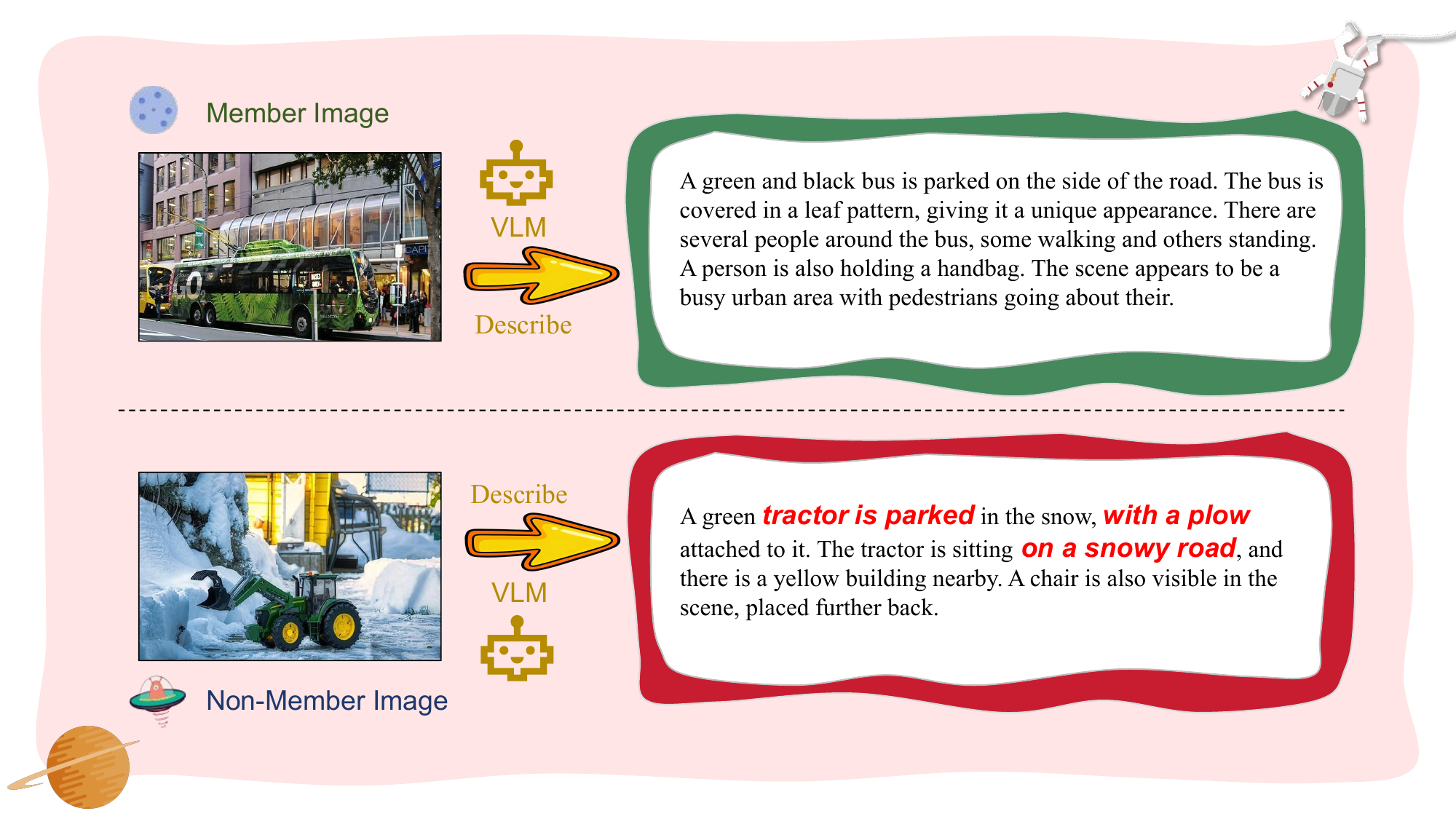} 
  \caption{Qualitative analysis of VLM (LLaVA-1.5) description on member vs. non-member image. Red text highlights factual hallucinations, including misidentified objects, incorrect components, and erroneous scene contexts.}
  \label{fig:huanjue}
\end{figure}

To address this problem, we investigate MIAs from the perspective of cross-modal semantic alignment. As illustrated in Figure~\ref{fig:huanjue}, we present a representative example showing that VLMs exhibit a distinct ``memorization effect'' on member images, generating descriptions with more accurate and specific attributes such as precise bus features and meticulous descriptions of scene details. In contrast, outputs for non-member images often contain semantic hallucinations. For example, a toy tractor is misidentified as a real one, and its state is erroneously described as ``parked'' rather than active or posed. Furthermore, the model fails to recognize specific components, such as mistaking a ``front loader bucket'' for a ``plow,'' and fails to capture the correct environment, incorrectly placing the object on a ``snowy road'' instead of a backyard path. This observation is not an isolated instance; we consistently find this phenomenon in various VLMs. More representative examples across additional models are detailed in Appendix~\ref{app:shili}.

Based on this insight, we propose Cross-modal Semantic Alignment Membership Inference Attack (CSA-MIA). Under a strict black-box single-sample setting, CSA-MIA queries the target VLM to generate a textual description for an input image. It then employs an external pre-trained cross-modal encoder, independent of the target VLM, to extract embeddings for the image and the text. The cosine similarity between the image and text embeddings serves as the membership signal, enabling a strict black-box attack without access to internal logits. While the existing black-box VLM membership inference method relies on unimodal textual signals, which provide limited discriminability at the level of individual samples, CSA-MIA explicitly uses the original image as a reference to evaluate semantic consistency between the image and the generated text. By leveraging the contrast between the model's precise memorization of member images and the potential semantic deviations in non-members, this approach produces a more accurate prediction of membership status for each individual query. To make the final membership decision, we set the decision threshold as the average alignment score computed over an auxiliary non-member reference set, which does not need to follow the training data distribution and can be easily obtained from synthesized or newly released data.

Experiments on a comprehensive suite of VLMs, including open-source models (LLaVA-1.5~\cite{liu2023visual}, MiniGPT-4~\cite{zhu2023minigpt}, and LLaMA Adapter v2~\cite{gao2023llamaadapterv2parameterefficientvisual}) and leading commercial APIs (GPT-4\cite{achiam2023gpt}, Claude-3~\cite{anthropic2024claude}), demonstrate that CSA-MIA enables effective membership inference from a single query without access to internal logits. To thoroughly analyze the method's dynamic properties, we conduct extensive ablation studies across different cross-modal encoders, generated text lengths, temperatures, and prompting strategies. Furthermore, CSA-MIA remains robust to diverse image perturbations and distribution shifts~\cite{takahashi2019data,hendrycks2019benchmarking,Xie_2020_CVPR}, demonstrating its utility against adversarial usage of unauthorized data in real-world scenarios.

In summary, the contributions of this paper are as follows:
\begin{itemize}
    \item We investigate a challenging yet realistic scenario: MIA against VLMs in strict black-box, single-sample setting. We find that member images exhibit significantly higher cross-modal semantic alignment than non-member images, allowing membership inference to be framed as detecting image-text alignment for individual samples.

    \item We propose CSA-MIA to address this challenging setting. CSA-MIA quantifies cross-modal alignment via an external encoder to conduct MIA and investigates decision threshold settings.

    \item We conduct extensive experiments on three datasets involving three open-source and two closed-source VLMs. The results demonstrate that CSA-MIA consistently achieves strong performance and remains robust to diverse image perturbations.
\end{itemize}

\section{Related Works}
In this section, we review the development of MIAs. We first discuss foundational methods targeting traditional models in Sec~\ref{Tradition}, followed by metric-based methods for LLMs in Sec~\ref{LLMs}. Finally, we analyze the emerging research and existing limitations regarding VLMs in Sec~\ref{VLMs}.

\subsection{MIA against Traditional Machine Learning}
\label{Tradition}
The concept of MIA is originally formalized by Shokri \textit{et al.}~\cite{shokri2017membership} in the context of discriminative classification tasks. Their work aims to determine whether a specific data sample is included in a target model's training dataset. To achieve this, they introduce a shadow model-based method, where multiple local models are trained to mimic the target model's behavior, generating a dataset of confidence vectors to train a binary attack classifier. Following this, a substantial body of work continues to refine this shadow training paradigm to achieve higher attack precision~\cite{liu2022membershipinferenceattacksexploiting,WatsonGCS22,9833649,Chaudhuri}. While effective, this method incurs high computational costs. Consequently, other studies aim to simplify the attack pipeline. Salem \textit{et al.}~\cite{salem2018mlleaksmodeldataindependent} demonstrate that shadow models are often unnecessary and that simple thresholding on the maximum prediction confidence suffices to distinguish members. Yeom \textit{et al.}~\cite{Yeom2017PrivacyRI} further analyze the theoretical connection between MIA and overfitting, proposing metric-based methods that leverage prediction loss to exploit the generalization gap between member and non-member samples. Furthermore, researchers investigate strictly black-box scenarios where confidence scores are unavailable. Choquette-Choo \textit{et al.}~\cite{pmlr-v139-choquette-choo21a} and Li \textit{et al.}~\cite{LiZ21} introduce label-only attacks that measure the model's robustness to input perturbations or the distance to decision boundaries to infer membership. However, these methods are primarily tailored for discriminative classification tasks and cannot be directly applied to the open-ended generation paradigm of modern LLMs and VLMs. Furthermore, adopting the shadow training ideology typically used in these methods is unrealistic in this context, as the prohibitive computational cost renders the training of such large-scale reference models infeasible~\cite{mattern-etal-2023-membership,duan2024do,HuYYLHZDY24}.

\subsection{MIA against LLMs.} 
\label{LLMs}
Extensive research scrutinizes MIAs across a diverse spectrum of machine learning architectures, spanning classification models~\cite{Yunhui,SongSM19}, generative models~\cite{hayes2017logan,hilprecht2019monte,chen2020gan}, and embedding models~\cite{mahloujifar2021membership,song2020information}. With the rapid development of LLMs, interest in MIA increases~\cite{mireshghallah2022empirical,fu2023practical,carlini2021extracting,shi2024detecting,ren2025self}. Due to the computational impracticality of shadow training in this domain, contemporary research shifts toward metric-based methods~\cite{carlini2021extracting,shi2024detecting,mattern2023membership,ren2025self} that leverage output distributions to identify member data. For example, Carlini \textit{et al.}~\cite{carlini2021extracting} observe that member data typically exhibit lower perplexity, Shi \textit{et al.}~\cite{shi2024detecting} note that the $k$\% most improbable tokens tend to have higher likelihood for members compared to non-members, Mattern \textit{et al.}~\cite{mattern2023membership} evaluate discrepancies in confidence under synonym substitutions, and Ren \textit{et al.}~\cite{ren2025self} assess fluctuations in model likelihood when sequences are paraphrased. Despite their effectiveness, most of these methods operate under a gray-box assumption, requiring full access to model logits. This constraint substantially limits their applicability in practical scenarios, a limitation that similarly affects large vision-language models.

\subsection{MIA against VLMs.} 
\label{VLMs}
Research on the privacy and security of VLMs largely focuses on poisoning~\cite{xu2024shadowcaststealthydatapoisoning} and adversarial or jailbreaking attacks~\cite{qi2023visualadversarialexamplesjailbreak,liu2024arondightredteaminglarge}, while membership inference for these models remains underexplored. To extend MIA research from LLMs to VLMs, recent studies apply text-based methods to multi-modal models by analyzing the captions generated for target images. Li \textit{et al.}~\cite{li2024membership} compute the MaxRényi-K\% metric over the complete logits produced by the model for the generated caption of each image, using this metric to distinguish member from non-member data. However, such gray-box methods rely heavily on access to internal probability distributions, limiting their applicability in real-world API scenarios. Hu \textit{et al.}~\cite{hu2025membership} propose a black-box method that calculates the self-similarity of multiple responses generated for each image within a target set and identifies member data based on the aggregate mean similarity. While effective at the dataset level, this method is inherently a dataset-level inference that requires multiple samples, which limits its usefulness in strict single-sample scenarios. In contrast, our proposed CSA-MIA overcomes these limitations by leveraging an external cross-modal encoder to quantify the semantic alignment consistency between generated responses and input images. This method establishes a strictly black-box and single-sample inference framework, achieving robustness without relying on internal logits or large datasets.

\section{Motivation}
\label{sec:prob}

In this section, we first present the problem formulation in Section~\ref{Threat}. Then, we detail the copilot experiment and analysis in Section~\ref{subsec:intuition} that motivate our proposed CSA-MIA.

\begin{figure}[t!]
     \centering
     \includegraphics[width=\linewidth]{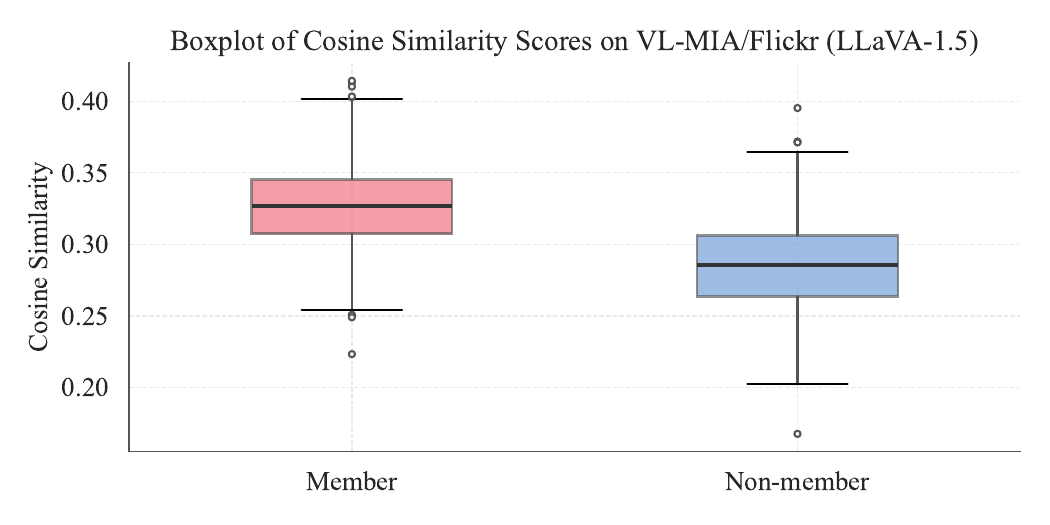} 
     \caption{Boxplots of CLIP cosine similarity scores between image embeddings and LLaVA-1.5 generated description embeddings for 300 member and 300 non-member images.}
    \label{fig:similarity_score_boxplot}
\end{figure}

\subsection{Threat Model}
\label{Threat}
\paragraph{Problem Formulation.}
The attacker aims to determine whether a single queried image was included in the training set of a target VLM. We study instance-level membership inference and formulate the attack as a binary classification problem. Let $\mathcal{D}$ denote the set of images used to train the VLM $M$.
Given a queried image $x$, the attacker seeks to decide whether $x \in \mathcal{D}$ or not. 


To perform membership inference, the attacker queries the target VLM $M$ with an input image $x$ and a textual instruction $T_{\text{in}}$ (e.g., ``Please describe the image as accurately and specifically as possible.''), and observes a generated textual description $T_{\text{out}}$.
Based on the image $x$ and the generated description
$T_{\text{out}}$, the attacker computes a scalar score
$\text{Score}(x, T_{\text{out}})$. The membership inference decision is obtained by comparing the score with a threshold $\lambda$, i.e.,
\begin{equation}
A(x) =
\begin{cases}
1, & \text{if } \text{Score}(x, T_{\text{out}}) > \lambda, \\
0, & \text{if } \text{Score}(x, T_{\text{out}}) \le \lambda.
\end{cases}
\end{equation}

\paragraph{Attacker's Knowledge.}
Unlike prior work that assumes access to token-level probabilities or logits, we adopt a black-box single-sample setting in which the attacker can only query the target VLM $M$ with an input image $x$ and a textual instruction $T_{\text{in}}$, and observe only one generated textual description $T_{\text{out}}$.
The attacker has no access to the architectural design, model parameters, or training algorithm of the target model $M$.
We note that this represents the most challenging and general setting for the attacker.

\begin{figure}[t!]
     \centering
     \includegraphics[width=\linewidth]{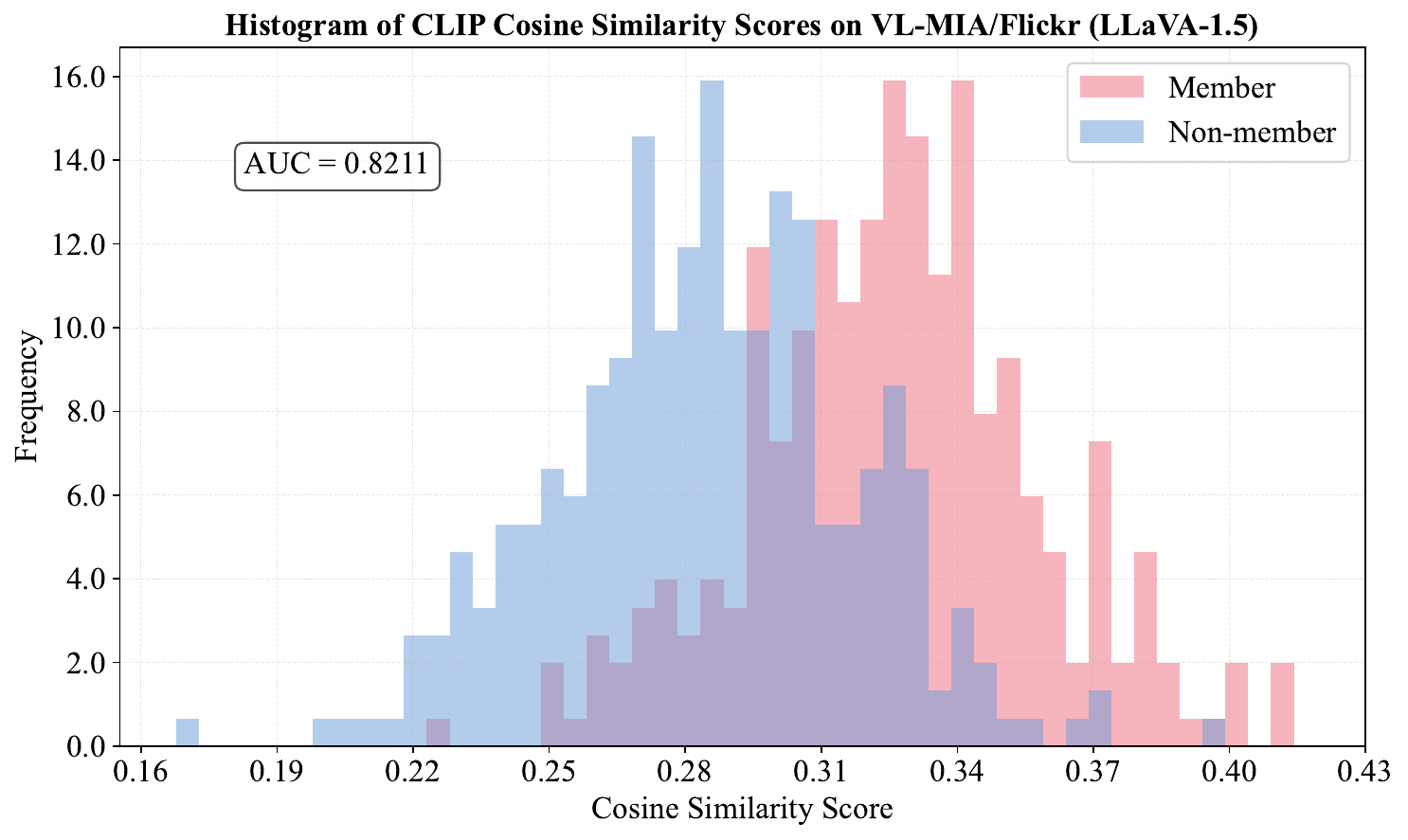} 
  \caption{Histogram of CLIP cosine similarity scores between image embeddings and LLaVA-1.5 generated description embeddings for 300 member and 300 non-member images.}
  \label{fig:similarity_score}
\end{figure}

\subsection{Copilot Experiment and Analysis}
\label{subsec:intuition}

Membership inference exploits the implicit memorization effects induced during training, which lead to observable deviations in output characteristics between member and non-member images. To investigate whether VLMs exhibit such ``memorization" effects, we conduct a comparative analysis on the LLaVA-1.5 model using the \texttt{Img\_Flickr} dataset constructed by Li \textit{et al.}~\cite{li2024membership}. Specifically, we sample 300 member images and 300 non-member images from the dataset to examine the model's generation behaviors under a unified textual instruction $T_{in}$: ``Please describe the image as accurately and specifically as possible.'' For each input image $x$ paired with $T_{in}$, we obtain the generated textual description $T_{out}$. Given the constraints of a black-box scenario where internal logits or confidence scores are typically inaccessible, we utilize a pre-trained CLIP model to extract embeddings for both the image and the generated description. We then compute the cosine similarity between them to characterize the cross-modal semantic consistency between the visual content and the generated description.


\begin{figure*}[t]
    \centering
    \includegraphics[width=0.90\linewidth, trim=0 1cm 0 1cm, clip]{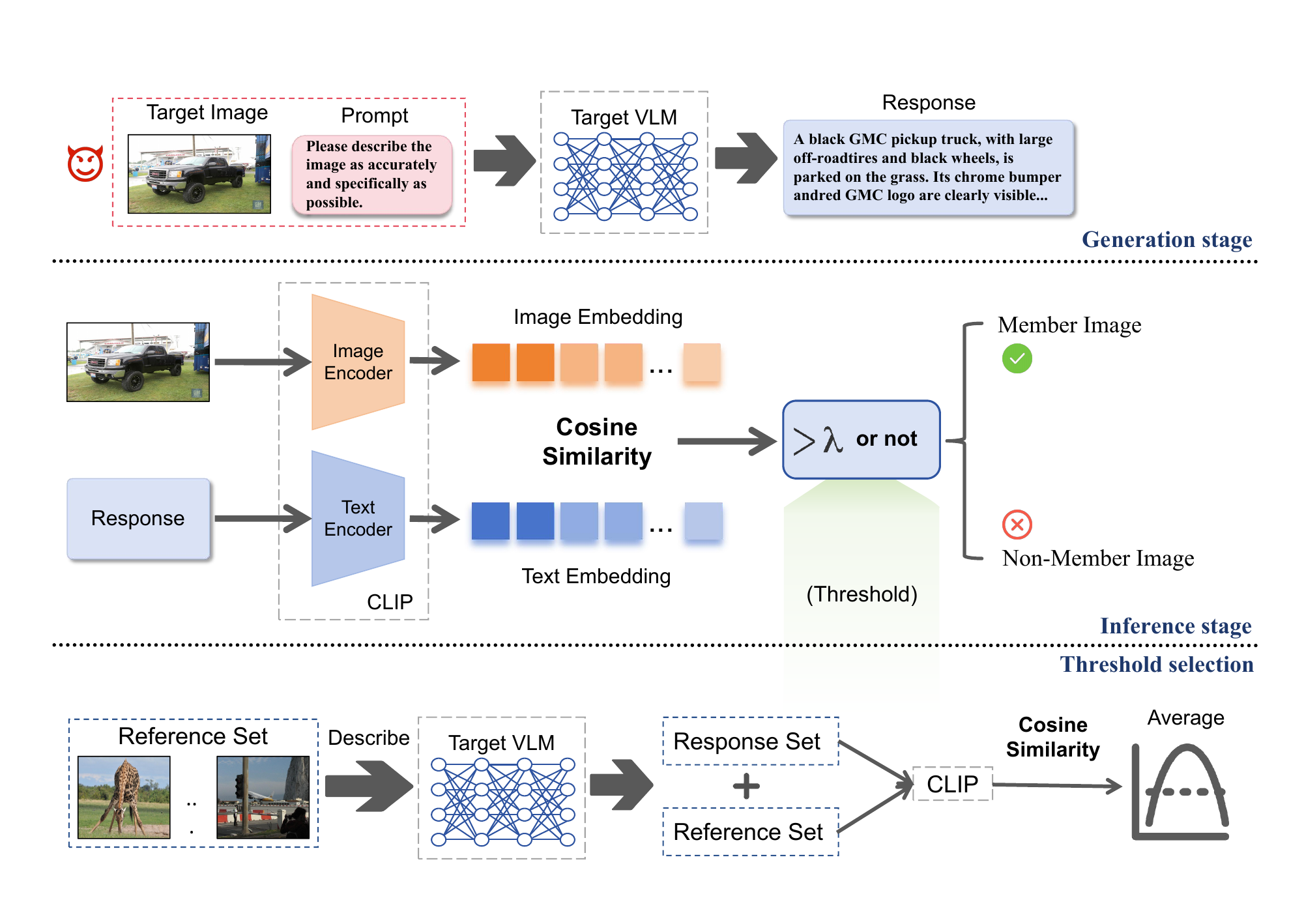}
    \caption{Overview of the proposed CSA-MIA.}
    \label{fig:overview}
\end{figure*}


The box plot in Figure~\ref{fig:similarity_score_boxplot} illustrates the distributions of cosine similarity for member and non-member images, highlighting a clear statistical separation where the median and interquartile range (IQR) for member images are higher. This trend is further corroborated by the histogram (Figure~\ref{fig:similarity_score}), which reveals that the density of member images is concentrated in the higher similarity range while the non-member images distribution is shifted toward lower values. Notably, this distributional discrepancy yields an AUC score of 0.8211, confirming that the similarity gap provides a highly discriminative signal for membership inference. Furthermore, this trend is not an isolated instance; we consistently observe this disparity across other models and images. More statistical visualizations are provided in the Appendix~\ref{app:tongji}. This gap in cross-modal semantic alignment directly inspires our methodology: by quantifying the degree of alignment between $x$ and $T_{\text{out}}$, we can potentially distinguish training members from non-members.

\section{Methodology}
\label{sec:method}
In this section, we present the implementation of our proposed method. Section~\ref{csa} details the procedure of CSA-MIA, and Section~\ref{subsec:threshold_calibration} introduces the threshold selection strategy.

\subsection{CSA-MIA}
\label{csa}

Building upon the insights detailed in Section \ref{subsec:intuition}, we formalize our black-box single-sample MIA method, termed CSA-MIA. The core of CSA-MIA leverages the discrepancy in cross-modal alignment between member and non-member data, which we categorize into two distinct stages:
\begin{itemize}
    \item Generation stage. The attacker queries the target VLM $M$ with an input image $x$ and a textual instruction $T_{\text{in}}$ (e.g., ``Please describe the image as accurately and specifically as possible.''). This interaction is formulated as $T_{\text{out}} = M(x, T_{\text{in}})$, where $T_{\text{out}}$ represents the textual description generated by $M$.
    \item Inference stage. We exploit the model's training memory to distinguish membership. We posit that for images seen during training, the model tends to generate descriptions with finer-grained details that exhibit higher semantic alignment in the visual-language joint embedding space. To quantify the degree of alignment, we employ a pretrained cross-modal encoder (e.g., CLIP) to extract the image embedding $\Phi_{\text{img}}(x)$ and the text embedding $\Phi_{\text{txt}}(T_{\text{out}})$. The cosine similarity $\text{Score}(x, T_{\text{out}})$ is then computed to measure the alignment quality:
   \begin{equation}
   \text{Score}(x, T_{\text{out}}) = \frac{\Phi_{\text{img}}(x) \cdot \Phi_{\text{txt}}(T_{\text{out}})}{\|\Phi_{\text{img}}(x)\| \|\Phi_{\text{txt}}(T_{\text{out}})\|}
   \end{equation}
   The computed score $\text{Score}(x, T_{\text{out}})$ serves as the foundational metric for single-sample membership detection, which is subsequently compared against the threshold $\lambda$ to yield a binary decision. The complete procedure of CSA-MIA is summarized in Algorithm ~\ref{alg:single_image_mia}.
\end{itemize}

\begin{algorithm}[t]
\caption{CSA-MIA}
\label{alg:single_image_mia}
\textbf{Input:} Target image $x$, target VLM $M$, instruction $T_{\text{in}}$, threshold $\lambda$.

\begin{algorithmic}[1]
\STATE Query $M$ with $x$ and $T_{\text{in}}$ to obtain output text $T_{\text{out}}$
\STATE Compute image embedding $\Phi_{\text{img}}(x)$ and text embedding $\Phi_{\text{txt}}(T_{\text{out}})$ using a pretrained cross-modal model
\STATE Compute the cosine similarity score $\text{Score}(x, T_{\text{out}})$ between the embeddings
\IF{$\text{Score}(x, T_{\text{out}}) > \lambda$}
    \STATE Conclude that $A(x) = 1$ ($x$ is a member)
\ELSE
    \STATE Conclude that $A(x) = 0$ ($x$ is a non-member)
\ENDIF
\end{algorithmic}
\textbf{Output:} Membership status $A(x) \in \{0, 1\}$
\end{algorithm}

\subsection{Threshold Selection via Reference Set}
\label{subsec:threshold_calibration}

To further enhance the accuracy of membership inference, we introduce an additional reference dataset $\mathcal{D}_{\text{ref}}$ for threshold selection. This reference set consists of confirmed non-member samples that do not overlap with the training set $\mathcal{D}$. For example, $\mathcal{D}_{\text{ref}}$ can be easily collected from synthesized data, unpublished data, or data released after the model's release date.

Specifically, we determine the decision threshold $\lambda$ by computing the average cosine similarity score over a reference dataset $\mathcal{D}_{\text{ref}}$. For each reference image $x^{i}_{\text{ref}} \in \mathcal{D}_{\text{ref}}$, we follow the same generation and inference procedure described in Section~\ref{csa}. We query the target VLM $M$ with the input pair $(x^{i}_{\text{ref}}, T_{\text{in}})$ to obtain the generated description $T^{i}_{\text{ref}}$, and compute the $\text{Score}(x^{i}_{\text{ref}}, T^{i}_{\text{ref}})$.
This process yields the set of $\text{Score}(x^i_{\text{ref}}, T^i_{\text{ref}})$ for all $i = 1, \dots, L$, where $L$ denotes the number of samples in $\mathcal{D}_{\text{ref}}$.
We define the decision threshold $\lambda$ as the empirical mean of these reference scores:
\begin{equation}
\lambda = \frac{1}{L} \sum_{i=1}^{L} \text{Score}(x^{i}_{\text{ref}}, T^{i}_{\text{ref}}).
\end{equation}
The rationale behind this strategy is to establish an empirical baseline for cross-modal alignment. Since $\mathcal{D}_{\text{ref}}$ consists exclusively of unseen samples, the derived threshold $\lambda$ effectively represents the model's expected alignment quality when generalizing to new data. Consequently, any sample yielding a score higher than this average indicates a degree of alignment that exceeds the model's normal generalization ability, thereby serving as a strong indicator of training data memorization.

By introducing $\mathcal{D}_{\text{ref}}$ to determine a concrete value for $\lambda$, our method enables a complete and practical black-box single-sample MIA pipeline. This allows the attacker to output a definitive binary membership status for any individual image $x$.

  \section{Experiments}
In this section, we present a comprehensive set of experiments to demonstrate the effectiveness of CSA-MIA. We first introduce the experimental setup and evaluation metrics in Section~\ref{setup}. Then, in Section~\ref{open}, we evaluate CSA-MIA across three open-source models and in Section~\ref{close} across two close-source models. In Section~\ref{threshold}, we investigate the impact of reference set size on the stability of threshold selection and verify the effectiveness of our strategy across different attack settings. Section~\ref{ablation} presents ablation studies that analyze the impact of key factors, including the length of description, embedding model, temperature, prompt. Finally, in Section~\ref{robustness}, we conduct a series of robustness experiments to examine the performance of CSA-MIA under more general infringement scenarios.
\subsection{Experimental Setup}
\label{setup}

\paragraph{Evaluation Metric.}
Following standard evaluation protocols in membership inference studies, we comprehensively assess attack performance using the Area Under the ROC Curve (AUC) and detailed utility metrics at strict low False Positive Rate thresholds (i.e., TPR @ $\alpha$\% FPR, where $\alpha \in \{1, 5, 10\}$).
AUC measures the overall discriminative capability across all decision thresholds.
Crucially, to quantify the specific privacy risk, we adopt the Membership Advantage (Adv) metric proposed by Yeom et al.~\cite{Yeom2017PrivacyRI}. 
Advantage is defined as the difference between the true positive rate and the false positive rate ($\text{Adv} = \text{TPR} - \text{FPR}$). 
It formally measures the adversary's predictive gain over random guessing.
Complementary to Adv, we also report Recall, Precision, and Accuracy at each threshold to provide a holistic view of the attack's effectiveness under stringent constraints. Detailed definitions and computation procedures for all evaluation metrics are presented in Appendix~\ref{sec:Metrics}.

\paragraph{Models and Datasets.}We evaluate our method on both open-source and closed-source models. The open-source models include LLaVA-1.5~\cite{liu2023visual}, MiniGPT-4~\cite{zhu2023minigpt}, and LLaMA Adapter v2~\cite{gao2023llamaadapterv2parameterefficientvisual}. In addition, we conduct experiments on closed-source models, including GPT-4~\cite{achiam2023gpt} and Claude-3~\cite{anthropic2024claude}. A detailed description of these open-source model is provided in Appendix~\ref{sec:Models}. We evaluate our method on three public benchmark datasets with different scales~\cite{li2024membership},
namely VL-MIA/Flickr, VL-MIA/Flickr-2k, and VL-MIA/Flickr-10k.

\begin{table*}[t]
\centering
\caption{Comprehensive evaluation of CSA-MIA versus baselines on the VL-MIA/Flickr dataset using LLaVA-1.5. We report AUC and detailed utility metrics at strict False Positive Rate thresholds of 1\%, 5\%, and 10\%. The metrics Adv, Rec, Prec, and Acc correspond to Advantage, Recall, Precision, and Accuracy, respectively. \textbf{Bold} indicates the best AUC within each column and \underline{underline} indicates the second best result.}
\label{tab:llava_600}
\resizebox{\textwidth}{!}{%
\begin{tabular}{ll c cccc cccc cccc}
\toprule
\multirow{3}{*}[-1.0ex]{\textbf{Type}} & 
\multirow{3}{*}[-1.0ex]{\textbf{Method}} & 
\multirow{3}{*}[-1.0ex]{\textbf{AUC}} 
& \multicolumn{12}{c}{\textbf{LLaVA-1.5}} \\
\cmidrule(lr){4-15}

& & & \multicolumn{4}{c}{\textbf{TPR @ 1\% FPR}} 
    & \multicolumn{4}{c}{\textbf{TPR @ 5\% FPR}} 
    & \multicolumn{4}{c}{\textbf{TPR @ 10\% FPR}} \\
\cmidrule(lr){4-7} \cmidrule(lr){8-11} \cmidrule(lr){12-15}

& & & Adv & Rec & Prec & Acc 
    & Adv & Rec & Prec & Acc 
    & Adv & Rec & Prec & Acc \\
\midrule

\multirow{13}{*}{Gray-Box}
& Aug\_KL             & 0.627 & -0.010 & 0.000 & 0.000 & 0.495 & 0.017 & 0.063 & 0.576 & 0.508 & 0.227 & 0.157 & 0.618 & 0.530 \\
& Perplexity          & 0.691 & 0.063 & 0.073 & 0.880 & 0.532 & 0.110 & 0.157 & 0.771 & 0.555 & 0.300 & 0.260 & 0.722 & 0.580 \\
& Max\_Prob\_Gap      & 0.673 & 0.033 & 0.043 & 0.813 & 0.517 & 0.123 & 0.173 & 0.776 & 0.562 & 0.300 & 0.320 & 0.762 & 0.610 \\
& Min\_0\% Prob       & 0.638 & 0.047 & 0.057 & 0.850 & 0.523 & 0.077 & 0.123 & 0.726 & 0.538 & 0.210 & 0.227 & 0.694 & 0.563 \\
& Min\_10\% Prob      & 0.670 & 0.050 & 0.060 & 0.857 & 0.525 & 0.120 & 0.167 & 0.781 & 0.560 & 0.280 & 0.220 & 0.695 & 0.562 \\
& Min\_20\% Prob      & 0.674 & 0.063 & 0.073 & 0.880 & 0.532 & 0.113 & 0.163 & 0.766 & 0.557 & 0.273 & 0.250 & 0.721 & 0.577 \\
& MaxRényi\_Max\_0\%  & 0.689 & 0.057 & 0.067 & 0.870 & 0.528 & 0.167 & 0.217 & 0.813 & 0.583 & 0.317 & 0.267 & 0.727 & 0.583 \\
& MaxRényi\_Max\_10\% & 0.719 & \textbf{0.077} & \underline{0.087} & \underline{0.897} & \underline{0.538} & 0.110 & 0.153 & 0.780 & 0.555 & 0.350 & 0.307 & 0.754 & 0.603 \\
& MaxRényi\_Max\_100\%& \underline{0.723} & 0.063 & 0.073 & 0.880 & 0.532 & \underline{0.177} & \underline{0.227} & \underline{0.819} & \underline{0.588} & \underline{0.360} & \underline{0.330} & \underline{0.767} & \underline{0.615} \\
& ModRényi            & 0.678 & \underline{0.067} & 0.077 & 0.885 & 0.533 & 0.077 & 0.127 & 0.717 & 0.538 & 0.283 & 0.233 & 0.700 & 0.567 \\
& Min-0\% ++          & 0.608 & 0.037 & 0.047 & 0.824 & 0.518 & 0.023 & 0.070 & 0.600 & 0.512 & 0.197 & 0.180 & 0.643 & 0.540 \\
& Min-10\% ++         & 0.638 & 0.027 & 0.037 & 0.786 & 0.513 & 0.080 & 0.130 & 0.722 & 0.540 & 0.260 & 0.220 & 0.688 & 0.560 \\
& Min-20\% ++         & 0.647 & 0.017 & 0.027 & 0.727 & 0.508 & 0.157 & 0.207 & 0.805 & 0.578 & 0.250 & 0.300 & 0.750 & 0.600 \\
\cmidrule(lr){1-15} 

\multirow{2}{*}{Black-Box}
& Image-only Inference & 0.587 & 0.000 & 0.000 & 0.000 & 0.500 & -0.003 & 0.047 & 0.467 & 0.497 & 0.003 & 0.103 & 0.508 & 0.502 \\
& \cellcolor{oursblue}\textbf{CSA-MIA (Ours)} 
& \cellcolor{oursblue}\textbf{0.821} 
& \cellcolor{oursblue} 0.008 & \cellcolor{oursblue}\textbf{0.090} & \cellcolor{oursblue}\textbf{0.900} & \cellcolor{oursblue}\textbf{0.540} 
& \cellcolor{oursblue}\textbf{0.287} & \cellcolor{oursblue}\textbf{0.337} & \cellcolor{oursblue}\textbf{0.871} & \cellcolor{oursblue}\textbf{0.643} 
& \cellcolor{oursblue}\textbf{0.401} & \cellcolor{oursblue}\textbf{0.510} & \cellcolor{oursblue}\textbf{0.832} & \cellcolor{oursblue}\textbf{0.703} \\

\bottomrule
\end{tabular}
}
\end{table*}

\paragraph{Baselines.}We compare our method against representative membership inference attacks that have been widely used in VLM settings. These baselines can be broadly categorized into gray-box and black-box approaches. Gray-box baselines include: (i) \textbf{Aug-KL}~\cite{li2024membership,liu2021encodermi}: This method extends the feature-consistency idea to LVLMs by comparing the output logit distributions of the original image and its augmented variants, using the KL divergence to quantify their discrepancy; (ii) \textbf{Perplexity} (Loss-based attack)~\cite{li2024membership,carlini2021extracting}: This approach performs membership inference based on the model loss, which reduces to perplexity in language modeling scenarios. Prior work shows that member samples tend to exhibit lower perplexity than non-members; (iii) \textbf{Max\_Prob\_Gap}~\cite{li2024membership}: This metric captures extreme token-level confidence by computing, at each token position, the difference between the highest and second-highest predicted probabilities, and averaging this gap across all positions; (iv) \textbf{Min-K\% Prob}\cite{shi2024detecting}: Originally proposed for LLMs, Min-K\% computes statistics over the lowest K\% predicted probabilities corresponding to ground-truth tokens; (v) \textbf{Min-K\% ++}~\cite{zhang2025mink}: An improved variant of Min-K\% that introduces theoretically motivated selection terms to better identify pre-training data; (vi) \textbf{MaxR\'enyi-Max-K\%}~\cite{li2024membership}: This method selects the top K\% tokens with the highest R\'enyi entropy values and uses the average of these entropies as the attack statistic; (vii) \textbf{ModR\'enyi}~\cite{li2024membership}: ModR\'enyi is an extended variant of MaxR\'enyi-Max-K\% designed for target-based scenarios. Black-box baselines include: (i) \textbf{Image-only Inference}~\cite{hu2025membership}: A strict black-box membership inference baseline that relies solely on the model’s final generated outputs. The method performs membership inference by analyzing similarity-based statistical discrepancies across multiple generated responses.

\paragraph{Implementation details.}All Gray-box baselines require access to the target model’s fine-grained output statistics. For the Min-K\% and Min-K\%++ baselines, we select K = 0, 10, and 20 to capture their best-performing configurations. For R\'enyi-based baselines (MaxR\'enyi-Max-K\% and ModR\'enyi), we choose K = 0, 10, and 100 to cover their optimal operating points, and fix the R\'enyi order to $\alpha = 0.5$ in all experiments. For the Image-only Inference baseline, we similarly adopt the best-performing setting by setting the number of repeated queries to K = 10 and using ROUGE-2 as the similarity metric.

\subsection{CSA-MIA on Open-source Models}
\label{open}
In this subsection, we show the effectiveness of our method in different VLMs and datasets. Due to space limitations, we only present a subset of the comparative results here. Specifically, Fig.~\ref{tab:llava_600} presents the results of LLaVA-1.5 on VL-MIA/Flickr, and the results on larger-scale datasets are summarized in Table~\ref{tab:full_complete_table}. The complete comparative results are detailed in Appendix~\ref{app:acr}. We get three observations from the results.
\begin{table*}[t]
\centering
\caption{Comprehensive evaluation of CSA-MIA versus baselines on the VL-MIA/Flickr-2k and VL-MIA/Flickr-10k datasets using LLaVA-1.5, MiniGPT-4 and LLaMA Adapter v2. We report AUC and detailed utility metrics at 5\% FPR. The metrics Adv, Rec, Prec, and Acc correspond to Advantage, Recall, Precision, and Accuracy, respectively. \textbf{Bold} indicates the best results, and \underline{underlined} indicates the second-best results.}
\label{tab:full_complete_table}

\resizebox{\textwidth}{!}{%
\begin{tabular}{@{\hspace{5pt}} lll YYYYY YYYYY YYYYY @{\hspace{5pt}}}
\toprule
\multirow{3}{*}[-0.5ex]{\textbf{Type}} & 
\multirow{3}{*}[-0.5ex]{\textbf{Method}} & 
\multirow{3}{*}[-0.5ex]{\textbf{Data}} & 
\multicolumn{5}{c}{\textbf{LLaVA-1.5}} & 
\multicolumn{5}{c}{\textbf{MiniGPT-4}} & 
\multicolumn{5}{c}{\textbf{LLaMA Adapter v2}} \\

\cmidrule(r){4-8} \cmidrule(lr){9-13} \cmidrule(l){14-18}

& & & \multirow{2}{*}{\textbf{AUC}} & \multicolumn{4}{c}{\textbf{TPR @ 5\% FPR}} 
    & \multirow{2}{*}{\textbf{AUC}} & \multicolumn{4}{c}{\textbf{TPR @ 5\% FPR}} 
    & \multirow{2}{*}{\textbf{AUC}} & \multicolumn{4}{c}{\textbf{TPR @ 5\% FPR}} \\

\cmidrule(r){5-8} \cmidrule(lr){10-13} \cmidrule(l){15-18}

& & & & Adv & Rec & Prec & Acc 
      & & Adv & Rec & Prec & Acc 
      & & Adv & Rec & Prec & Acc \\
\midrule

\multirow{26}{*}{\shortstack{Gray-\\Box}} 
& \multirow{2}{*}{Aug\_KL} 
  & 2k  & 0.577 & 0.066 & 0.116 & 0.707 & 0.534 & 0.534 & 0.004 & 0.045 & 0.495 & 0.500 & 0.469 & -0.018 & 0.032 & 0.390 & 0.491 \\
& & 10k & 0.599 & 0.072 & 0.122 & 0.711 & 0.536 & 0.506 & 0.018 & 0.068 & 0.578 & 0.509 & 0.448 & -0.016 & 0.034 & 0.402 & 0.492 \\
\cmidrule{2-18}

& \multirow{2}{*}{Perplexity} 
  & 2k  & 0.683 & 0.089 & 0.139 & 0.739 & 0.545 & 0.609 & 0.032 & 0.082 & 0.621 & 0.516 & 0.649 & 0.068 & 0.117 & 0.705 & 0.534 \\
& & 10k & 0.695 & 0.107 & 0.157 & 0.759 & 0.554 & 0.614 & 0.045 & 0.095 & 0.656 & 0.523 & 0.663 & 0.099 & 0.149 & 0.749 & 0.549 \\
\cmidrule{2-18}

& \multirow{2}{*}{Max\_Prob\_Gap} 
  & 2k  & 0.668 & 0.139 & 0.189 & 0.794 & 0.570 & 0.641 & \textbf{0.096} & \textbf{0.146} & \textbf{0.749} & \textbf{0.549} & 0.650 & 0.125 & 0.175 & 0.778 & 0.563 \\
& & 10k & 0.676 & 0.135 & 0.185 & 0.787 & 0.568 & 0.650 & 0.093 & 0.143 & 0.740 & 0.546 & 0.644 & 0.110 & 0.160 & 0.762 & 0.555 \\
\cmidrule{2-18}

& \multirow{2}{*}{Min\_0\% Prob} 
  & 2k  & 0.584 & 0.018 & 0.068 & 0.576 & 0.509 & 0.551 & 0.005 & 0.055 & 0.524 & 0.503 & 0.584 & -0.001 & 0.049 & 0.495 & 0.500 \\
& & 10k & 0.601 & 0.029 & 0.079 & 0.612 & 0.515 & 0.536 & 0.015 & 0.065 & 0.565 & 0.508 & 0.590 & 0.035 & 0.085 & 0.631 & 0.518 \\
\cmidrule{2-18}

& \multirow{2}{*}{Min\_10\% Prob} 
  & 2k  & 0.614 & 0.033 & 0.083 & 0.624 & 0.517 & 0.579 & -0.002 & 0.044 & 0.489 & 0.499 & 0.609 & 0.033 & 0.083 & 0.624 & 0.517 \\
& & 10k & 0.633 & 0.047 & 0.096 & 0.660 & 0.523 & 0.565 & 0.024 & 0.074 & 0.597 & 0.512 & 0.624 & 0.049 & 0.099 & 0.665 & 0.525 \\
\cmidrule{2-18}

& \multirow{2}{*}{Min\_20\% Prob} 
  & 2k  & 0.629 & 0.053 & 0.103 & 0.673 & 0.527 & 0.592 & 0.025 & 0.075 & 0.600 & 0.513 & 0.619 & 0.029 & 0.078 & 0.614 & 0.515 \\
& & 10k & 0.649 & 0.059 & 0.108 & 0.688 & 0.530 & 0.548 & 0.035 & 0.085 & 0.630 & 0.518 & 0.637 & 0.062 & 0.112 & 0.691 & 0.531 \\
\cmidrule{2-18}

& \multirow{2}{*}{\small MaxRényi\_0\%} 
  & 2k  & 0.651 & 0.115 & 0.165 & 0.767 & 0.558 & 0.562 & -0.014 & 0.035 & 0.417 & 0.493 & 0.612 & 0.071 & 0.121 & 0.708 & 0.536 \\
& & 10k & 0.661 & 0.057 & 0.107 & 0.682 & 0.529 & 0.546 & 0.002 & 0.052 & 0.512 & 0.501 & 0.632 & 0.106 & 0.156 & 0.758 & 0.553 \\
\cmidrule{2-18}

& \multirow{2}{*}{\small MaxRényi\_10\%} 
  & 2k  & 0.672 & 0.052 & 0.102 & 0.671 & 0.526 & 0.667 & 0.076 & 0.126 & 0.716 & 0.538 & \underline{0.716} & 0.134 & 0.184 & 0.786 & 0.567 \\
& & 10k & 0.679 & 0.069 & 0.118 & 0.705 & 0.534 & 0.703 & \textbf{0.140} & \textbf{0.190} & \textbf{0.791} & \textbf{0.570} & 0.658 & 0.098 & 0.148 & 0.748 & 0.549 \\
\cmidrule{2-18}

& \multirow{2}{*}{\small MaxRényi\_100\%} 
  & 2k  & \underline{0.712} & \underline{0.162} & \underline{0.212} & \underline{0.809} & \underline{0.581} & \underline{0.679} & 0.076 & 0.126 & 0.724 & 0.539 & 0.713 & \underline{0.153} & \underline{0.203} & \underline{0.809} & \underline{0.578} \\
& & 10k & \underline{0.707} & 0.113 & 0.163 & 0.766 & 0.557 & \underline{0.707} & \underline{0.138} & 0.188 & \underline{0.790} & \underline{0.569} & \underline{0.709} & \textbf{0.166} & \textbf{0.216} & \textbf{0.812} & \textbf{0.583} \\
\cmidrule{2-18}

& \multirow{2}{*}{ModRényi} 
  & 2k  & 0.688 & 0.127 & 0.177 & 0.780 & 0.564 & 0.602 & 0.032 & 0.082 & 0.621 & 0.516 & 0.641 & 0.071 & 0.121 & 0.708 & 0.536 \\
& & 10k & 0.703 & \underline{0.144} & \underline{0.194} & \underline{0.795} & \underline{0.572} & 0.602 & 0.040 & 0.090 & 0.643 & 0.520 & 0.654 & 0.089 & 0.139 & 0.737 & 0.545 \\
\cmidrule{2-18}

& \multirow{2}{*}{Min-0\% ++} 
  & 2k  & 0.588 & -0.012 & 0.038 & 0.432 & 0.494 & 0.621 & 0.035 & 0.085 & 0.634 & 0.518 & 0.612 & 0.061 & 0.110 & 0.692 & 0.531 \\
& & 10k & 0.548 & -0.012 & 0.038 & 0.434 & 0.494 & 0.634 & 0.040 & 0.090 & 0.642 & 0.520 & 0.608 & 0.052 & 0.102 & 0.671 & 0.526 \\
\cmidrule{2-18}

& \multirow{2}{*}{Min-10\% ++} 
  & 2k  & 0.650 & 0.078 & 0.128 & 0.723 & 0.540 & 0.621 & 0.035 & 0.085 & 0.634 & 0.518 & 0.655 & 0.086 & 0.136 & 0.731 & 0.543 \\
& & 10k & 0.626 & 0.041 & 0.091 & 0.645 & 0.521 & 0.634 & 0.040 & 0.090 & 0.642 & 0.520 & 0.648 & 0.076 & 0.126 & 0.716 & 0.538 \\
\cmidrule{2-18}

& \multirow{2}{*}{Min-20\% ++} 
  & 2k  & 0.666 & 0.141 & 0.191 & 0.793 & 0.571 & 0.649 & 0.021 & 0.071 & 0.587 & 0.511 & 0.679 & 0.101 & 0.150 & 0.754 & 0.551 \\
& & 10k & 0.663 & 0.113 & 0.163 & 0.765 & 0.556 & 0.682 & 0.088 & 0.138 & 0.735 & 0.544 & 0.670 & 0.084 & 0.134 & 0.729 & 0.542 \\
\midrule

\multirow{4}{*}{\shortstack{Black-\\Box}} 
& \multirow{2}{*}{\shortstack[l]{Image-only\\Inference}} 
  & 2k  & 0.556 & 0.017 & 0.067 & 0.573 & 0.509 & 0.557 & 0.028 & 0.078 & 0.614 & 0.515 & 0.386 & -0.039 & 0.011 & 0.183 & 0.481 \\
& & 10k & 0.578 & 0.011 & 0.061 & 0.548 & 0.505 & 0.486 & -0.023 & 0.027 & 0.479 & 0.499 & 0.388 & -0.025 & 0.025 & 0.334 & 0.488 \\

\cmidrule{2-18}

& \cellcolor{oursblue} & \cellcolor{oursblue} 2k  
  & \cellcolor{oursblue}\textbf{0.814} & \cellcolor{oursblue}\textbf{0.342} & \cellcolor{oursblue}\textbf{0.392} & \cellcolor{oursblue}\textbf{0.887} & \cellcolor{oursblue}\textbf{0.671} 
  & \cellcolor{oursblue}\textbf{0.722} & \cellcolor{oursblue}\underline{0.092} & \cellcolor{oursblue}\underline{0.142} & \cellcolor{oursblue}\underline{0.740} & \cellcolor{oursblue}\underline{0.546} 
  & \cellcolor{oursblue}\textbf{0.755} & \cellcolor{oursblue}\textbf{0.180} & \cellcolor{oursblue}\textbf{0.230} & \cellcolor{oursblue}\textbf{0.821} & \cellcolor{oursblue}\textbf{0.590} \\

& \cellcolor{oursblue}\multirow{-2}{*}{\textbf{CSA-MIA (Ours)}}  & \cellcolor{oursblue} 10k 
  & \cellcolor{oursblue}\textbf{0.760} & \cellcolor{oursblue}\textbf{0.163} & \cellcolor{oursblue}\textbf{0.213} & \cellcolor{oursblue}\textbf{0.810} & \cellcolor{oursblue}\textbf{0.582} 
  & \cellcolor{oursblue}\textbf{0.712} & \cellcolor{oursblue}0.102 & \cellcolor{oursblue} \underline{0.152} & \cellcolor{oursblue} 0.752 & \cellcolor{oursblue}0.551 
  & \cellcolor{oursblue}\textbf{0.751} & \cellcolor{oursblue}\underline{0.159} & \cellcolor{oursblue}\underline{0.209} & \cellcolor{oursblue} \underline{0.807} & \cellcolor{oursblue}\underline{0.579} \\

\bottomrule
\end{tabular}
}
\end{table*}

\textbf{Comparison with the black-box baseline method, CSA-MIA demonstrates superior performance across various models and datasets.} As summarized in Table~\ref{tab:llava_600} and Table~\ref{tab:full_complete_table}, in a strictly black-box environment, where an attacker can only query the model and observe its output, CSA-MIA consistently outperforms image-only inference baseline methods on all evaluated models. In summary, these results demonstrate that even under conditions of limited black-box access, CSA-MIA significantly enhances member inference capabilities, outperforming existing image-only baseline methods, and exhibits clear and consistent improvements across various experimental settings.

\textbf{Comparison with gray-box methods, CSA-MIA achieves competitive and often superior performance across different models and datasets.}
As summarized in Table~\ref{tab:llava_600} and Table~\ref{tab:full_complete_table}, despite operating in a more rigorous black-box environment, CSA-MIA consistently performs comparably to gray-box membership inference methods that utilize additional internal information.  These results demonstrate that despite its more rigorous threat modeling, CSA-MIA remains highly effective, matching or even surpassing gray-box baseline methods across different model architectures and dataset sizes, exhibiting a clear and consistent performance trend.

\textbf{The strength of membership inference signals correlates with the scale of trainable parameters.} 
As shown in the Figure~\ref{tab:llava_600} and Figure~\ref{tab:full_complete_table}, our results indicate that the effectiveness of image-based MIA is closely related to the number of trainable parameters involved in model training. Models that update more parameters tend to exhibit a stronger memory effect, thus producing more discriminative membership signals. This trend is consistent with the training pipelines used by different visual language models. Specifically, MiniGPT-4 updates only the image projection layer during image training, while LLaMA Adapter v2 relies on a parameter-efficient fine-tuning strategy. In contrast, LLaVA-1.5 updates both the image projection layer and the underlying language model simultaneously. Therefore, models with a wider range of parameter updates generally exhibit stronger membership inference signals, i.e., increasing the number of trainable parameters enhances the memory of training data. Conversely, models trained using limited or parameter-efficient update methods tend to produce weaker membership cues, making inference more challenging.

\subsection{CSA-MIA on Closed-source Models}
\label{close}
In this section, we demonstrate the effectiveness of CSA-MIA on closed-source models, including GPT-4 and Claude-3. 

\begin{table}[t]
    \centering
    \caption{Release information and knowledge cut-off dates of evaluated
    closed-source models.}
    \label{tab:closed_source_models_imformation}
    \begin{tabular}{lcc}
        \toprule
        \textbf{Model} & \textbf{Release} & \textbf{Knowledge Cut-off Date} \\
        \midrule
        GPT-4 & May 2024 & October 2023 \\
        Claude-3-haiku    & March 2024 & August 2023 \\
        \bottomrule
    \end{tabular}
\end{table}

\begin{table}[t]
\centering
\caption{CSA-MIA on \textbf{GPT-4}. \textbf{Bold} indicates the best AUC within each column and \underline{underline} indicates the second best result.
}
\label{tab:gpt4_results}
\resizebox{\linewidth}{!}{
\begin{tabular}{llccccc}
\toprule
\multirow{2}{*}{\textbf{Type}} 
& \multirow{2}{*}{\textbf{Method}} 
& \multirow{2}{*}{\textbf{AUC}} 
& \multicolumn{4}{c}{\textbf{TPR@5\%FPR}} \\
\cmidrule(lr){4-7}
& & & \textbf{Adv} & \textbf{Recall} & \textbf{Precision} & \textbf{Acc} \\
\midrule

\multirow{13}{*}{Gray-Box}
& Aug-KL              & 0.483 & -0.030 & 0.020 & 0.333 & 0.490 \\
& Perplexity          & 0.358 & -0.030 & 0.020 & 0.286 & 0.485 \\

& Max-Prob-Gap        
& \underline{0.616} 
& \underline{0.219} 
& \underline{0.182} 
& \underline{0.783} 
& \underline{0.566} \\

& Min-0\% Prob        & 0.413 & -0.010 & 0.040 & 0.444 & 0.495 \\
& Min-10\% Prob       & 0.408 & -0.030 & 0.020 & 0.250 & 0.480 \\
& Min-20\% Prob       & 0.382 & 0.011  & 0.061 & 0.600 & 0.510 \\
& MaxRényi-Max-0\%    & 0.375 & -0.020 & 0.030 & 0.375 & 0.490 \\
& MaxRényi-Max-10\%   & 0.362 & -0.010 & 0.040 & 0.364 & 0.485 \\
& MaxRényi-Max-100\%  & 0.306 & -0.040 & 0.010 & 0.167 & 0.480 \\
& ModRényi            & 0.407 & -0.010 & 0.040 & 0.571 & 0.505 \\
& Min-0\%++           & 0.440 & -0.010 & 0.040 & 0.444 & 0.495 \\
& Min-10\%++          & 0.423 & -0.010 & 0.040 & 0.444 & 0.495 \\
& Min-20\%++          & 0.403 & -0.030 & 0.020 & 0.286 & 0.485 \\

\midrule
\multirow{2}{*}{Black-Box}
& Image-only Inference 
& 0.658 
& 0.040 
& 0.090 
& 0.643 
& 0.520 \\

& \cellcolor{oursblue}\textbf{CSA-MIA (Ours)} 
& \cellcolor{oursblue}\textbf{0.736} 
& \cellcolor{oursblue}\textbf{0.220} 
& \cellcolor{oursblue}\textbf{0.270} 
& \cellcolor{oursblue}\textbf{0.844} 
& \cellcolor{oursblue}\textbf{0.608} \\

\bottomrule
\end{tabular}
}
\end{table}

\begin{table}[t]
\centering
\caption{Performance comparison of black-box MIA methods on \textbf{Claude-3}.
Results are reported in terms of AUC and TPR at 5\% FPR.
\textbf{Bold} indicates the best result.}
\label{tab:claude3_results}

\resizebox{\columnwidth}{!}{
\begin{tabular}{cccccc}
\toprule
\multirow{2}{*}{\raggedright\textbf{Method}}
& \multirow{2}{*}{\raggedright\textbf{AUC}}
& \multicolumn{4}{c}{\textbf{TPR@5\%FPR}} \\
\cmidrule(lr){3-6}
& & \textbf{Adv} & \textbf{Recall} & \textbf{Precision} & \textbf{Acc} \\
\midrule

Image-only Inference
& 0.668 
& 0.010 
& 0.060 
& 0.546 
& 0.505 \\

\rowcolor{oursblue}
\textbf{CSA-MIA (Ours)}
& \textbf{0.738}
& \textbf{0.220}
& \textbf{0.270}
& \textbf{0.844}
& \textbf{0.610} \\

\bottomrule
\end{tabular}
}
\end{table} 

As summarized in Table~\ref{tab:closed_source_models_imformation}, both GPT-4 and Claude-3-haiku have a knowledge cutoff in 2023. Consequently, these models could not have been exposed during pretraining to the non-member images collected after January 2024 in the VL-MIA/Flickr dataset. For evaluation, we randomly sample an equal number of member and non-member images from the VL-MIA/Flickr dataset, resulting in a total of 200 test images. We conduct experiments using the GPT-4 and claude-3-haiku APIs. Each image is prompted to generate a textual description with a maximum length of 70 tokens.

Due to API access limitations, GPT-4 only exposes partial token-level probability information, returning the top-20 token probabilities at each generation step. Consequently, the evaluation of GPT-4 falls into the gray-box setting, where MIA methods that require access to the full vocabulary distribution cannot be applied directly. To enable a unified and fair evaluation across all gray-box methods, we assume a fixed vocabulary size of 32,000 and approximate the probability mass of the remaining 31,980 tokens using a uniform distribution. In contrast, Claude-3 does not provide any token-level probability information.
As a result, gray-box MIA methods are not applicable to Claude-3, and we restrict
our evaluation on Claude-3 to black-box attack settings.

The results on closed-source models, as reported in Tables~\ref{tab:gpt4_results} and~\ref{tab:claude3_results}, reveal consistent advantages of CSA-MIA under two practical access scenarios. 

As shown in Table~\ref{tab:gpt4_results}, when partial token-level probability information is available in GPT-4, CSA-MIA achieves the best metrics among all methods. In contrast, most gray-box baselines perform poorly in this setting, with AUC values below 0.5, indicating performance even worse than random guessing. This highlights the difficulty of extracting reliable membership signals from limited probability information and underscores the advantage of leveraging cross-modal semantic alignment. As shown in Table~\ref{tab:claude3_results}, in the pure black-box setting on Claude-3, CSA-MIA consistently outperforms the image-only inference baseline across all metrics.

\begin{table*}[t]
\centering
\caption{Effect of reference set size on threshold selection performance. For each reference set size, we \textbf{randomly sample} the reference subset \textbf{100 times} and report the \textbf{mean performance} across runs. \textbf{Bold} indicates the best AUC within each column and \underline{underline} indicates the second best result.}
\label{tab:threshold_refsize}

\resizebox{\textwidth}{!}{
\begin{tabular}{llcccccccccccccccc}
\toprule
\multirow{2}{*}{\textbf{Type}} 
& \multirow{2}{*}{\textbf{Method}} 
& \multicolumn{4}{c}{\textbf{Reference Set = 1}}
& \multicolumn{4}{c}{\textbf{Reference Set = 60}}
& \multicolumn{4}{c}{\textbf{Reference Set = 120}}
& \multicolumn{4}{c}{\textbf{Reference Set = 180}} \\
\cmidrule(lr){3-6} \cmidrule(lr){7-10} \cmidrule(lr){11-14} \cmidrule(lr){15-18}

& 
& Adv & Recall & Precision & Acc
& Adv & Recall & Precision & Acc
& Adv & Recall & Precision & Acc
& Adv & Recall & Precision & Acc \\
\midrule

\multirow{13}{*}{Gray-Box}
& Aug-KL
& 0.127 & 0.609 & 0.552 & 0.564
& 0.205 & 0.675 & 0.590 & 0.603
& 0.206 & 0.672 & 0.591 & 0.603
& 0.206 & 0.670 & 0.591 & 0.603 \\

& Perplexity
& 0.201 & 0.712 & 0.615 & 0.601
& 0.260 & 0.804 & 0.597 & 0.630
& 0.260 & 0.802 & 0.597 & 0.630
& 0.260 & 0.803 & 0.597 & 0.630 \\

& Max\_Prob\_Gap
& 0.186 & 0.655 & 0.623 & 0.593
& 0.231 & 0.725 & 0.596 & 0.616
& 0.232 & 0.721 & 0.596 & 0.616
& 0.233 & 0.724 & 0.596 & 0.617 \\

& Min-0\% Prob
& 0.139 & 0.658 & 0.589 & 0.570
& 0.199 & 0.715 & 0.581 & 0.600
& 0.199 & 0.726 & 0.580 & 0.600
& 0.200 & 0.730 & 0.580 & 0.600 \\

& Min-10\% Prob
& 0.163 & 0.627 & {0.620} & 0.581
& 0.263 & 0.785 & 0.601 & 0.631
& 0.266 & 0.783 & 0.603 & 0.633
& 0.267 & 0.785 & 0.603 & 0.633 \\

& Min-20\% Prob
& 0.177 & 0.694 & 0.607 & 0.588
& 0.253 & 0.771 & 0.599 & 0.626
& 0.258 & 0.767 & 0.601 & 0.629
& 0.260 & 0.768 & 0.602 & 0.630 \\

& MaxRényi-Max-0\%
& 0.179 & 
\underline{0.739} & 0.602 & 0.589
& 0.300 & 0.752 & 0.625 & 0.650
& 0.302 & 0.755 & 0.625 & 0.651
& 0.304 & 0.755 & 0.626 & 0.652 \\

& MaxRényi-Max-10\%
& \underline{0.229} & 0.651 & \underline{0.650} & \underline{0.615}
& \underline{0.330} & 0.806 & \underline{0.629} & \underline{0.665}
& \underline{0.333} & 0.809 & \underline{0.630} & \underline{0.666}
& \underline{0.334} & 0.808 & \underline{0.631} & \underline{0.667} \\

& MaxRényi-Max-100\%
& 0.225 & 0.621 & \textbf{0.666} & 0.612
& 0.313 & 0.782 & 0.626 & 0.656
& 0.312 & 0.791 & 0.623 & 0.656
& 0.311 & 0.791 & 0.622 & 0.655 \\

& ModRényi
& 0.178 & 0.682 & 0.609 & 0.589
& 0.248 & 0.765 & 0.598 & 0.624
& 0.248 & 0.768 & 0.597 & 0.624
& 0.253 & 0.767 & 0.599 & 0.626 \\

& Min-0\%++
& 0.106 & 0.602 & 0.573 & 0.553
& 0.133 & 0.760 & 0.549 & 0.567
& 0.136 & 0.757 & 0.550 & 0.568
& 0.137 & 0.757 & 0.550 & 0.568 \\

& Min-10\%++
& 0.146 & 0.599 & 0.583 & 0.573
& 0.125 & \underline{0.807} & 0.543 & 0.563
& 0.120 & \underline{0.814} & 0.540 & 0.560
& 0.117 & \underline{0.815} & 0.541 & 0.561 \\

& Min-20\%++
& 0.152 & 0.623 & 0.616 & 0.576
& 0.155 & 0.749 & 0.558 & 0.577
& 0.156 & 0.749 & 0.558 & 0.578
& 0.159 & 0.752 & 0.558 & 0.579 \\

\midrule
\multirow{2}{*}{Black-Box}
& Image-only Inference
& 0.082 & 0.595 & 0.536 & 0.541
& 0.135 & 0.523 & 0.574 & 0.568
& 0.136 & 0.757 & 0.550 & 0.568
& 0.137 & 0.757 & 0.550 & 0.568 \\

& \cellcolor{oursblue}\textbf{CSA-MIA (Ours)}
& \cellcolor{oursblue}\textbf{0.303} & \cellcolor{oursblue}\textbf{0.860} & \cellcolor{oursblue}{0.639} & \cellcolor{oursblue}\textbf{0.615}
& \cellcolor{oursblue}\textbf{0.397} & \cellcolor{oursblue}\textbf{0.903} & \cellcolor{oursblue}\textbf{0.642} & \cellcolor{oursblue}\textbf{0.698}
& \cellcolor{oursblue}\textbf{0.393} & \cellcolor{oursblue}\textbf{0.903} & \cellcolor{oursblue}\textbf{0.639} & \cellcolor{oursblue}\textbf{0.697}
& \cellcolor{oursblue}\textbf{0.392} & \cellcolor{oursblue}\textbf{0.903} & \cellcolor{oursblue}\textbf{0.639} & \cellcolor{oursblue}\textbf{0.696} \\

\bottomrule
\end{tabular}
}

\vspace{2em} 

\caption{Threshold selection performance on VL-MIA benchmarks.
The best results are shown in \textbf{bold}, and the second best are \underline{underlined}.}
\label{tab:threshold_result_ref_set_60}

\resizebox{0.85\linewidth}{!}{
\begin{tabular}{llcccccccccccc}
\toprule
\multirow{3}{*}{Type} &
\multirow{3}{*}{Method} &
\multicolumn{12}{c}{\textbf{VL-MIA/Flickr}} \\
\cmidrule(lr){3-14}
& &
\multicolumn{4}{c}{LLaVA-1.5} &
\multicolumn{4}{c}{Minigpt4} &
\multicolumn{4}{c}{LLaMA Adapter v2} \\
\cmidrule(lr){3-6} \cmidrule(lr){7-10} \cmidrule(lr){11-14}
& & Adv & Recall & Precision & Acc
& Adv & Recall & Precision & Acc
& Adv & Recall & Precision & Acc \\
\midrule

\multirow{13}{*}{Gray-Box}
& Aug-KL
& 0.205 & 0.675 & 0.590 & 0.603
& 0.040 & 0.477 & 0.522 & 0.520
& -0.020 & 0.400 & 0.488 & 0.490 \\

& Perplexity
& 0.260 & \underline{0.804} & 0.597 & 0.630
& 0.190 & 0.670 & 0.583 & 0.595
& 0.157 & 0.717 & 0.561 & 0.578 \\

& Max-Prob-Gap
& 0.231 & 0.725 & 0.596 & 0.616
& 0.137 & 0.623 & 0.562 & 0.568
& 0.163 & 0.700 & 0.566 & 0.582 \\

& Min-0\% Prob
& 0.199 & 0.715 & 0.581 & 0.600
& 0.040 & 0.623 & 0.517 & 0.520
& 0.167 & 0.703 & 0.567 & 0.583 \\

& Min-10\% Prob
& 0.263 & 0.785 & 0.601 & 0.631
& 0.123 & 0.627 & 0.555 & 0.562
& 0.150 & 0.660 & 0.564 & 0.575 \\

& Min-20\% Prob
& 0.253 & 0.771 & 0.599 & 0.626
& 0.163 & 0.690 & 0.567 & 0.582
& \underline{0.187} & 0.677 & \underline{0.580} & \underline{0.593} \\

& MaxRényi-Max-0\%
& 0.300 & 0.752 & \underline{0.625} & 0.650
& -0.063 & 0.437 & 0.466 & 0.468
& 0.150 & 0.597 & 0.572 & 0.575 \\

& MaxRényi-Max-10\%
& \underline{0.330} & \underline{0.806} & \underline{0.629} & \underline{0.665}
& 0.090 & 0.553 & 0.544 & 0.545
& 0.163 & 0.683 & 0.568 & 0.582 \\

& MaxRényi-Max-100\%
& 0.313 & 0.782 & 0.626 & 0.656
& \underline{0.197} & \underline{0.773} & 0.573 & \underline{0.598}
& 0.183 & 0.720 & 0.573 & 0.592 \\

& ModRényi
& 0.248 & 0.765 & 0.598 & 0.624
& 0.123 & 0.643 & 0.553 & 0.562
& 0.170 & 0.693 & 0.570 & 0.585 \\

& Min-0\%++
& 0.133 & 0.760 & 0.549 & 0.567
& 0.050 & 0.190 & 0.576 & 0.525
& 0.060 & 0.700 & 0.522 & 0.530 \\

& Min-10\%++
& 0.125 & 0.807 & 0.543 & 0.563
& 0.077 & 0.407 & 0.552 & 0.538
& 0.120 & \underline{0.800} & 0.541 & 0.560 \\

& Min-20\%++
& 0.155 & 0.749 & 0.558 & 0.577
& 0.133 & 0.703 & 0.552 & 0.567
& 0.123 & 0.683 & 0.550 & 0.562 \\

\midrule
\multirow{2}{*}{Black-Box}
& Image-only Inference
& 0.135 & 0.523 & 0.574 & 0.568
& 0.183 & 0.631 & \underline{0.585} & 0.592
& 0.022 & 0.474 & 0.512 & 0.511 \\

& \cellcolor{oursblue}\textbf{CSA-MIA(Ours)}
& \cellcolor{oursblue}\textbf{0.397}
& \cellcolor{oursblue}\textbf{0.903}
& \cellcolor{oursblue}\textbf{0.642}
& \cellcolor{oursblue}\textbf{0.698}
& \cellcolor{oursblue}\textbf{0.344}
& \cellcolor{oursblue}\textbf{0.859}
& \cellcolor{oursblue}\textbf{0.626}
& \cellcolor{oursblue}\textbf{0.672}
& \cellcolor{oursblue}\textbf{0.402}
& \cellcolor{oursblue}\textbf{0.903}
& \cellcolor{oursblue}\textbf{0.644}
& \cellcolor{oursblue}\textbf{0.701} \\

\bottomrule
\end{tabular}
}
\end{table*}

\subsection{Analysis of Threshold selection} 
\label{threshold}
To analyze the effect of threshold selection via the reference set, we conduct experiments on the LLaVA-1.5 model with reference subsets of different sizes. Specifically, we vary the number of samples in the reference set $\mathcal{D}_{ref}$ and evaluate the performance of CSA-MIA under different selection scales. This setting allows us to examine how the choice of reference set size influences the stability and effectiveness of the calibrated decision threshold.

All reference sets are randomly sampled from the non-member portion of the VL-MIA/Flickr dataset. For each reference set size, we repeat the sampling process 100 times to reduce the impact of randomness. As shown in Table~\ref{tab:threshold_refsize}, CSA-MIA consistently achieves
the best performance under threshold selection with reference sets of
different sizes. Notably, even when the reference set consists of only a single non-member sample, CSA-MIA attains a high recall of 0.860, indicating strong capability in identifying member samples in practical scenarios. 

As the reference set size increases, the performance of CSA-MIA gradually converges. In particular, when the reference set size reaches 60, the performance metrics become relatively stable, and further increasing the reference set size does not lead to consistent improvements. Based on this observation, we adopt a reference set size of 60 as the default choice for threshold selection in subsequent experiments. More fine-grained analyses on the impact of reference set size are provided in Appendix~\ref{sec:fine_grained_impact_ref_set_size}.

We further analyze the effectiveness of threshold selection across different attack settings and model architectures. As shown in Table~\ref{tab:threshold_result_ref_set_60}, CSA-MIA consistently outperforms existing baselines under both gray-box and black-box settings on all VL-MIA benchmarks. A notable observation is that threshold selection brings particularly significant improvements in recall, which is critical for membership inference attacks. For completeness, we report the full experimental results on all datasets, models, and selection settings in Appendix~\ref{sec:complete_threshold_results}.

\begin{table}[h]
\centering
\caption{Performance comparison of different encoders across LLaVA-1.5, MiniGPT-4, and LLaMA Adapter v2. \textbf{Bold} indicates the best result.}
\label{tab:wide_comparison}
\resizebox{\linewidth}{!}{
\begin{tabular}{l cc cc cc}
\toprule
\multirow{2}{*}{\textbf{Encoder}} 
& \multicolumn{2}{c}{\textbf{LLaVA-1.5}} 
& \multicolumn{2}{c}{\textbf{MiniGPT-4}} 
& \multicolumn{2}{c}{\textbf{LLaMA Adapter v2}} \\

\cmidrule(lr){2-3} \cmidrule(lr){4-5} \cmidrule(lr){6-7} 

& \textbf{AUC} & \textbf{TPR@5\%FPR} & \textbf{AUC} & \textbf{TPR@5\%FPR} & \textbf{AUC} & \textbf{TPR@5\%FPR} \\
\midrule

CLIP   & \textbf{0.821} & 0.337          & \textbf{0.773} & \textbf{0.347} & \textbf{0.812} & 0.257 \\
SLIP   & 0.816          & \textbf{0.363} & 0.750          & 0.260          & 0.806          & \textbf{0.380} \\
FLIP   & 0.807          & 0.320          & 0.760          & 0.253          & 0.784          & 0.287 \\
DeCLIP & 0.813          & 0.340          & 0.747          & 0.287          & 0.730          & 0.167 \\

\bottomrule
\end{tabular}
}
\end{table}

\begin{figure*}[htb]
    \centering
    \captionsetup{
        font=small,
        labelfont=small
    }

    \begin{subfigure}{0.32\textwidth}
        \centering
        \includegraphics[width=\linewidth]{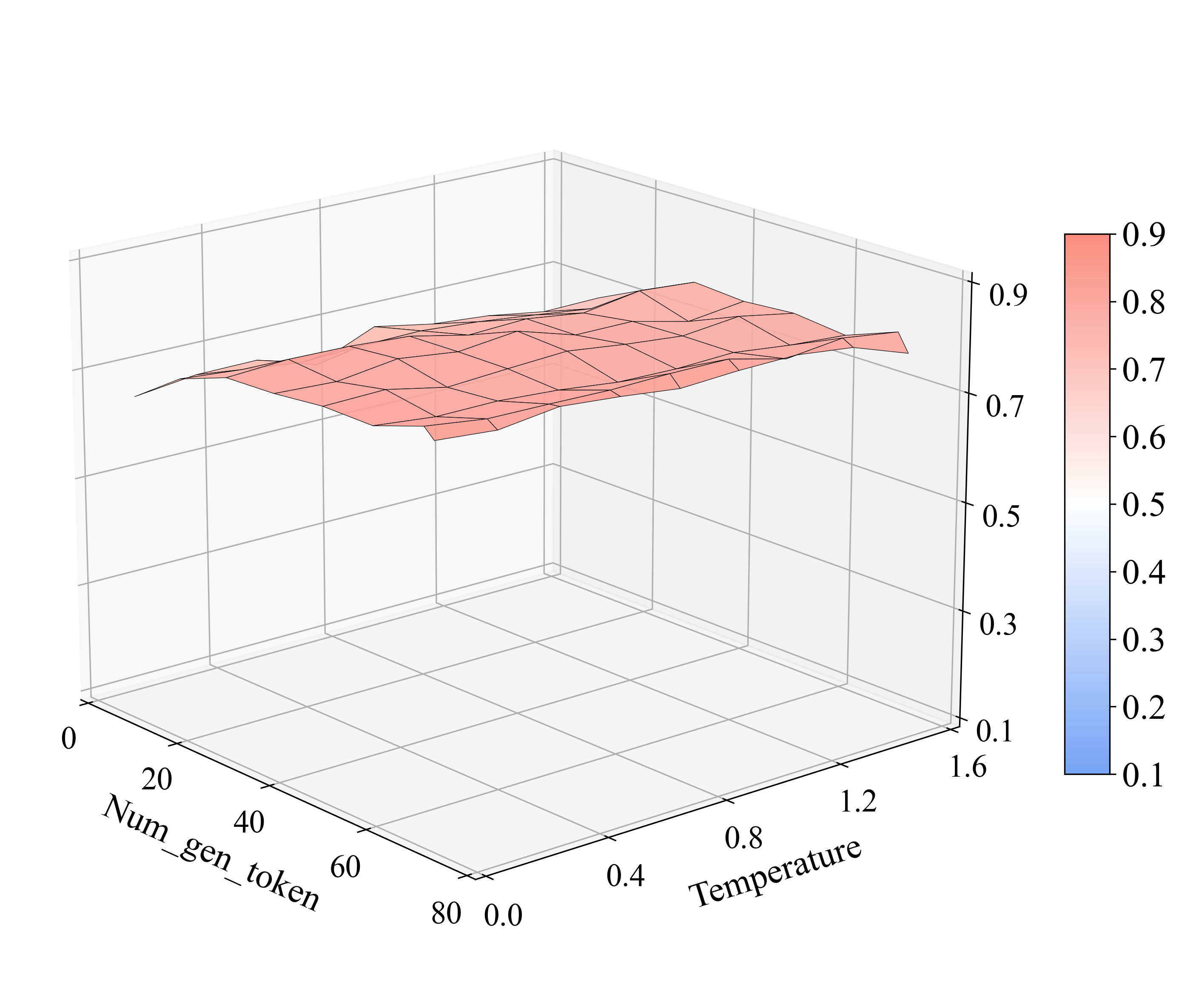}
        \caption{LLaVA-1.5 (AUC)}
        \label{fig:LLaVA_AUC}
    \end{subfigure}
    \hfill
    \begin{subfigure}{0.32\textwidth}
        \centering
        \includegraphics[width=\linewidth]{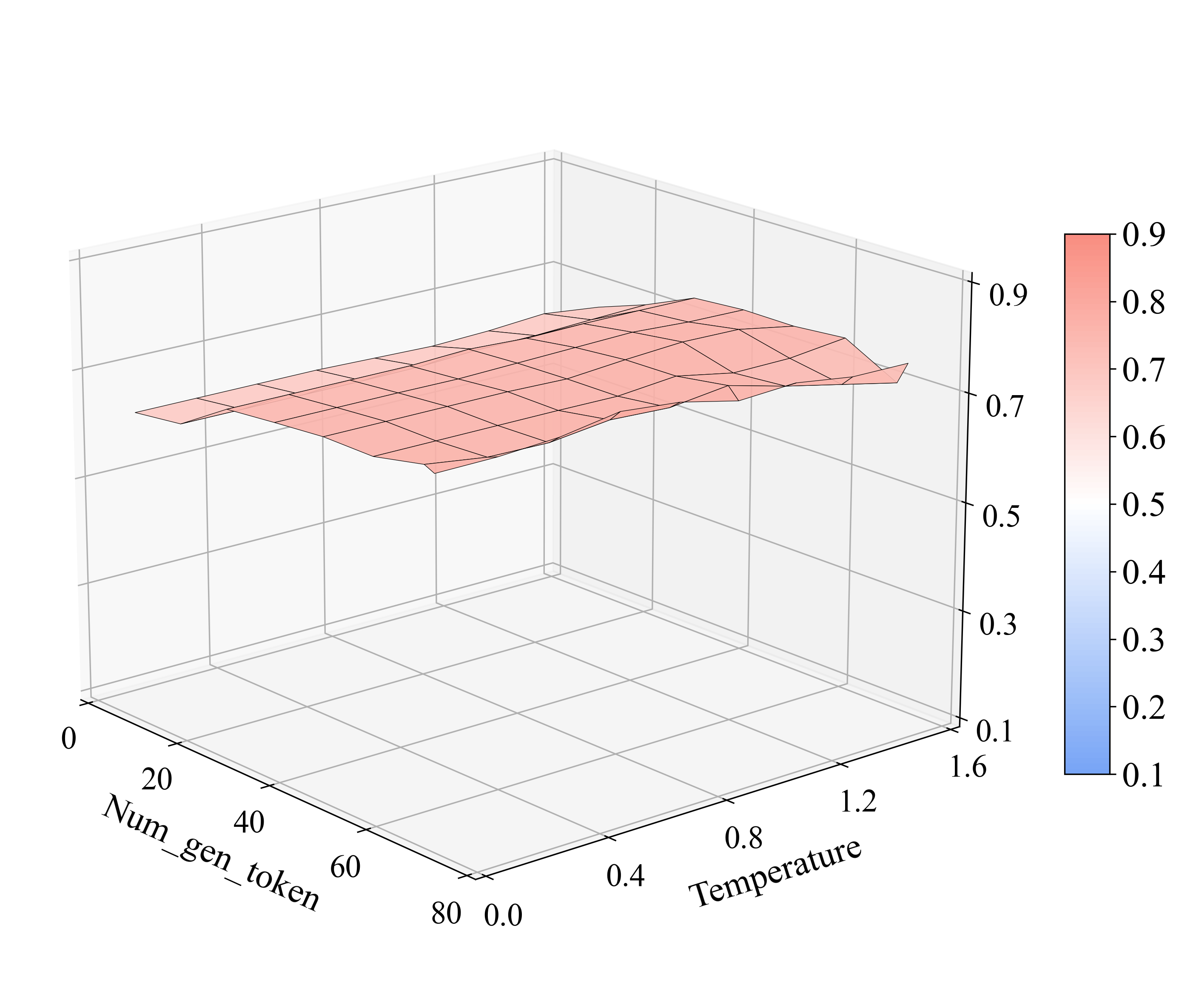}
        \caption{MiniGPT-4 (AUC)}
        \label{fig:MiniGPT-4_AUC}
    \end{subfigure}
    \hfill
    \begin{subfigure}{0.32\textwidth}
        \centering
        \includegraphics[width=\linewidth]{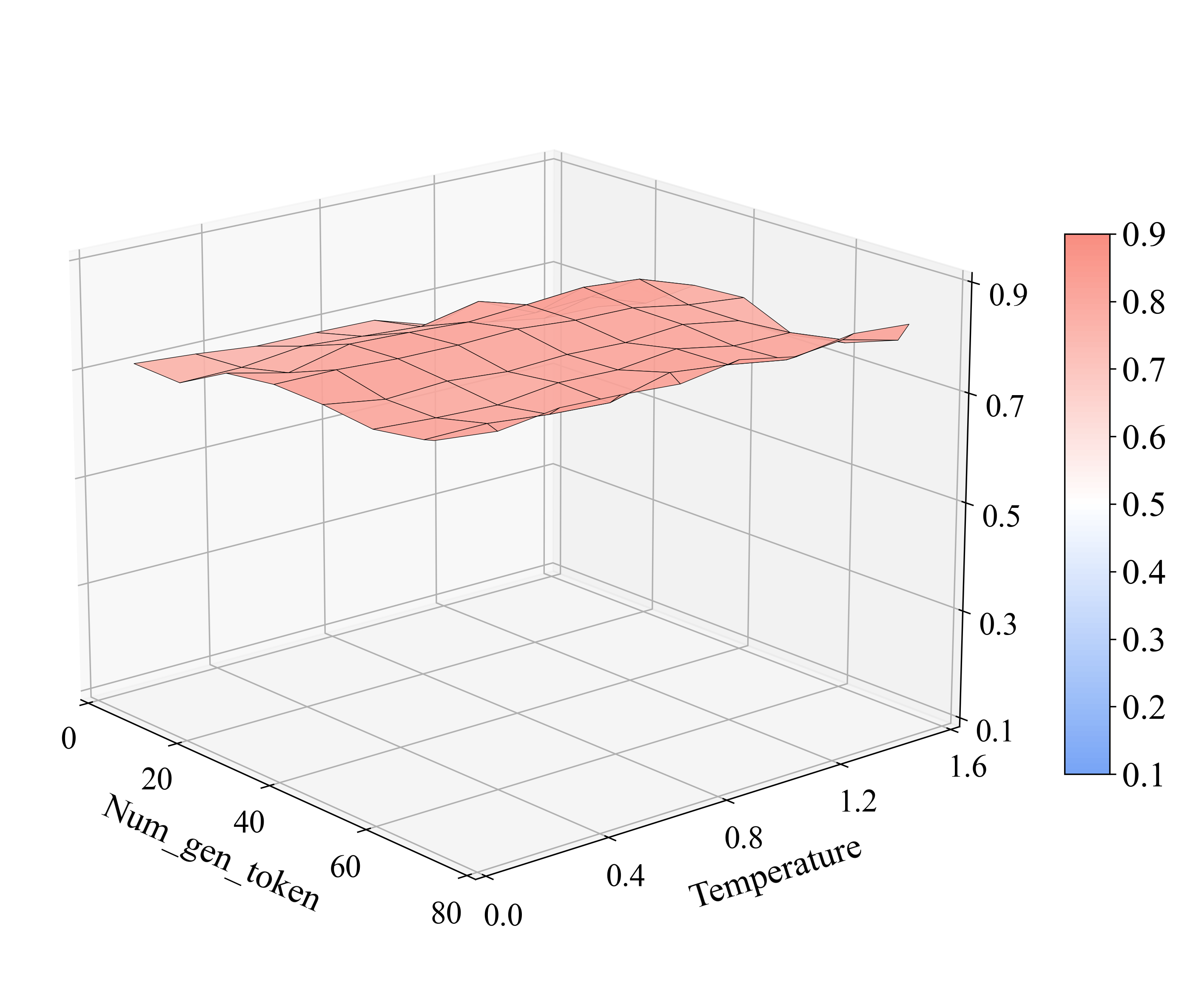}
        \caption{LLaMA Adapter v2 (AUC)}
        \label{fig:LLaMA_adapter_v2_auc}
    \end{subfigure}

    \vspace{2mm}

    \begin{subfigure}{0.32\textwidth}
        \centering
        \includegraphics[width=\linewidth]{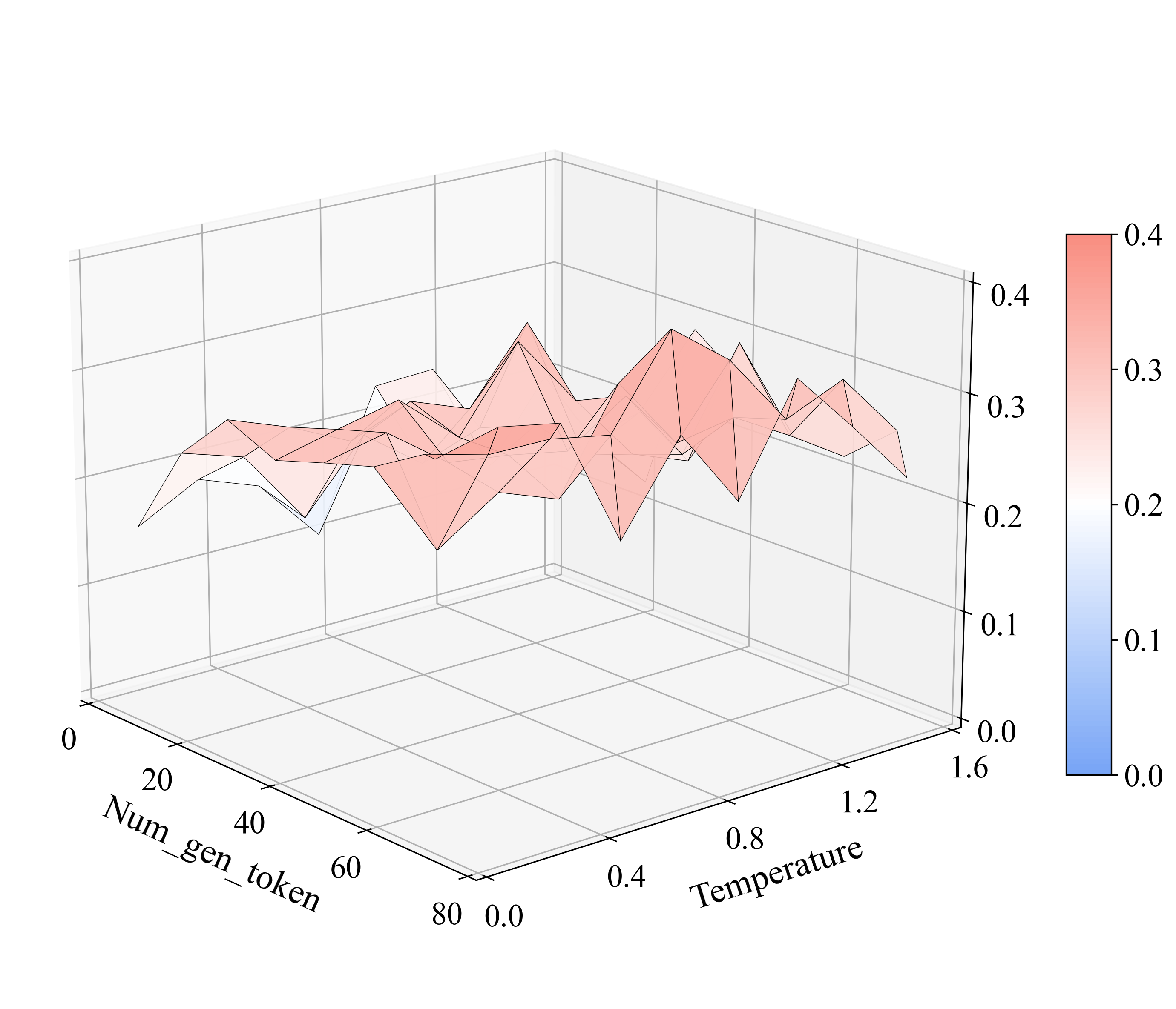}
        \caption{LLaVA-1.5 (TPR@5\%FPR)}
        \label{fig:LLaVA_tpr}
    \end{subfigure}
    \hfill
    \begin{subfigure}{0.32\textwidth}
        \centering
        \includegraphics[width=\linewidth]{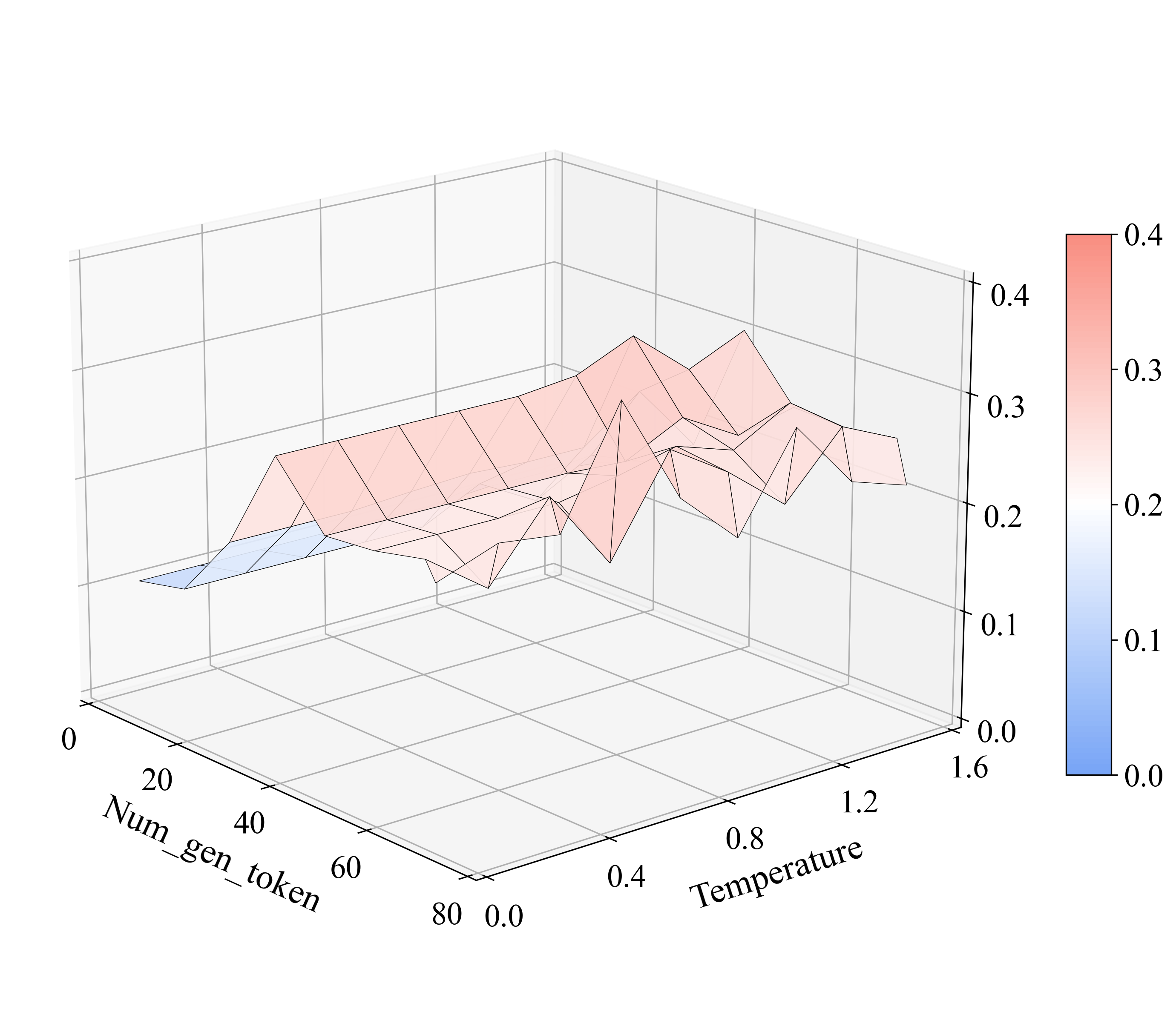}
        \caption{MiniGPT-4 (TPR@5\%FPR)}
        \label{fig:MiniGPT-4_tpr}
    \end{subfigure}
    \hfill
    \begin{subfigure}{0.32\textwidth}
        \centering
        \includegraphics[width=\linewidth]{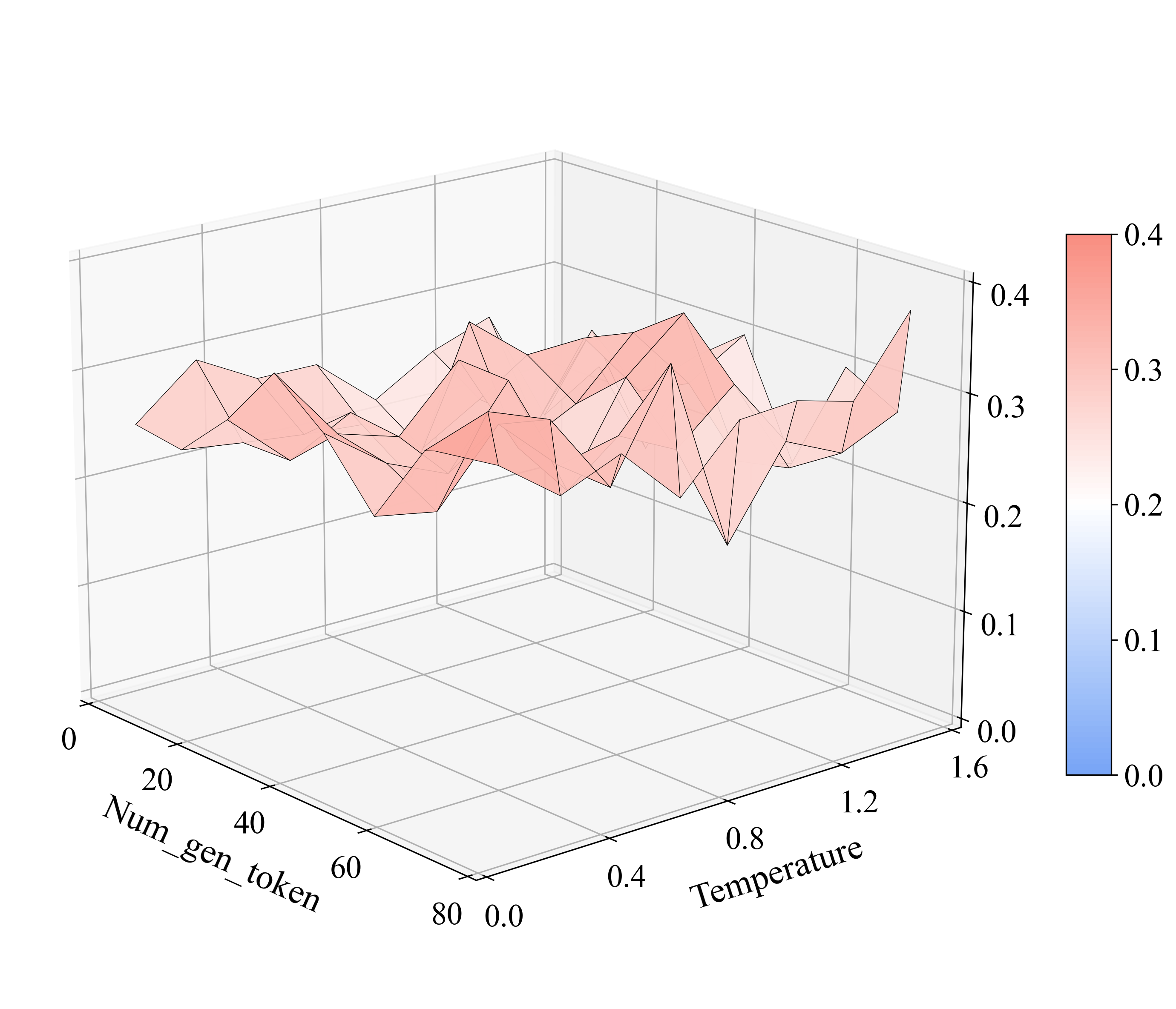}
        \caption{LLaMA Adapter v2 (TPR@5\%FPR)}
        \label{fig:LLaMA_adapter_v2_tpr}
    \end{subfigure}

    \caption{
    \textbf{Main results on different models.}
    Top row reports AUC scores, while the bottom row shows TPR at 5\% FPR.
    }
    \label{fig:main_results_grid}
\end{figure*}

\subsection{Ablation Studies}
\label{ablation}

\noindent\textbf{Does the embedding model affect the CSA-MIA performance?}
We further conduct experiments to study how the choice of cross-modal embedding model influences CSA-MIA. In our pipeline, the embedding model is used to project the target image and the generated description into a shared feature space, where we compute their similarity as the core membership signal. Keeping the settings fixed, we replace the encoder with CLIP\cite{pmlr-v139-radford21a}, SLIP\cite{DBLP:conf/eccv/MuK0X22}, FLIP\cite{Li_2023_CVPR}, and DeCLIP\cite{DBLP:conf/cvpr/WangCLKCT25}, which all adopt a dual-tower architecture trained with contrastive learning to align image–text representations. As shown in Table~\ref{tab:wide_comparison}, CLIP achieves the best performance among all embedding models. Meanwhile, other models also deliver strong results, suggesting that CSA-MIA is relatively robust to the choice of dual-tower contrastive encoder.

\begin{table}[htbp]
\centering
\caption[Image description prompts]{Different image description prompts used for generation.}
\label{tab:prompt_design}
\small
\setlength{\tabcolsep}{6pt}
\begin{tabular}{l p{\dimexpr\columnwidth-2.6cm}}
\toprule
\textbf{Prompt ID} & \textbf{Prompt Content} \\
\midrule
\makecell[c]{Prompt 1} &
Give a short, detail-rich caption describing only what is visible. Include the main subjects, the setting, any actions, notable colors, and any readable text. \\

\makecell[c]{Prompt 2} &
What’s happening in this image? Describe the visible content with concrete details. \\

\makecell[c]{Prompt 3} &
Give a short, detail-rich caption for the image. \\
\bottomrule
\end{tabular}
\end{table}

\noindent\textbf{Does the length of description affect the CSA-MIA performance?}
We conduct experiments across multiple vision--language models to investigate the impact of generated description length on the performance of CSA-MIA. During the generation stage, we control the length of the output descriptions by adjusting the \texttt{Num\_gen\_token} parameter, thereby producing descriptions of varying lengths. As shown in Figure~\ref{fig:main_results_grid}, the attack performance consistently improves with increasing description length at first, but declines when the descriptions become excessively long. The best performance is achieved at an intermediate description length. These results indicate that overly short descriptions may fail to capture sufficient discriminative semantic information, whereas overly long descriptions tend to introduce redundant or less relevant content, which weakens semantic alignment and degrades membership inference performance.

\begin{table}[h]
\centering
\caption{Comparison of AUC performance on LLaVA-1.5 under different prompts. \textbf{Bold} indicates the best result, and \underline{underline} indicates the second best result.}
\label{tab:prompt_combined_results}
\resizebox{\linewidth}{!}{
\begin{tabular}{ll ccc}
\toprule
\textbf{Type} & \textbf{Method} & \textbf{Prompt 1} & \textbf{Prompt 2} & \textbf{Prompt 3} \\
\midrule
\multirow{13}{*}{Gray-Box}
& Aug\_KL              & 0.582 & 0.611 & 0.607 \\
& Perplexity           & 0.647 & 0.672 & 0.675 \\
& Max\_Prob\_Gap       & 0.638 & 0.673 & 0.670 \\
& Min\_0\% Prob        & 0.589 & 0.623 & 0.570 \\
& Min\_10\% Prob       & 0.636 & 0.652 & 0.629 \\
& Min\_20\% Prob       & 0.651 & 0.659 & 0.668 \\
& MaxRényi\_Max\_0\%   & \underline{0.728} & 0.690 & 0.719 \\
& MaxRényi\_Max\_10\%  & 0.721 & 0.734 & 0.733 \\
& MaxRényi\_Max\_100\% & \underline{0.728} & \underline{0.772} & \underline{0.752} \\
& ModRényi             & 0.642 & 0.658 & 0.664 \\
& Min-0\% ++           & 0.577 & 0.616 & 0.587 \\
& Min-10\% ++          & 0.626 & 0.640 & 0.633 \\
& Min-20\% ++          & 0.628 & 0.667 & 0.637 \\
\midrule
\multirow{2}{*}{Black-Box}
& Image-only Inference & 0.616 & 0.606 & 0.620 \\
& \cellcolor{oursblue}\textbf{CSA-MIA (Ours)} 
& \cellcolor{oursblue}\textbf{0.802} & \cellcolor{oursblue}\textbf{0.786} & \cellcolor{oursblue}\textbf{0.790} \\
\bottomrule
\end{tabular}
}
\end{table}

\begin{figure*}[!t]
    \centering
    \includegraphics[width=0.8\textwidth]{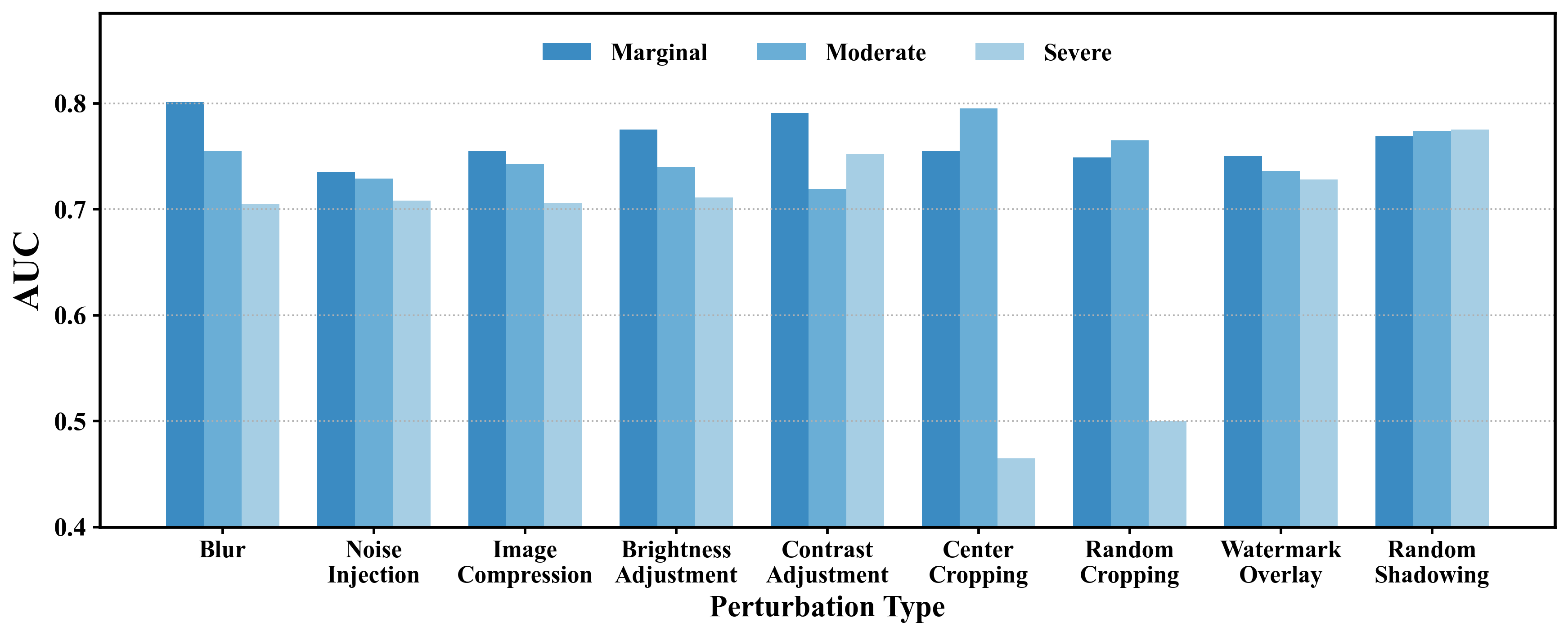}
    \caption{Robustness evaluation under nine common image perturbations at three severity levels (marginal, moderate, and severe), measured by AUC.}
    \label{fig:robustness_bar}

    \vspace{0.5cm}
    \centering
\small
\captionof{table}{Image perturbations used in robustness evaluation.}
\label{tab:perturbations}
\setlength{\tabcolsep}{6pt}
\begin{tabular}{llll}
\toprule
\rowcolor{lightgray}
\multicolumn{4}{l}{\textbf{Pixel-level perturbations}} \\
\midrule
Perturbation & Method & Parameter & Parameters (Marginal / Moderate / Severe)
 \\
\midrule
Blur &
Gaussian smoothing &
Blur radius &
1.0 / 3.0 / 5.0 \\
Noise Injection &
Additive Gaussian noise &
Noise level &
0.02 / 0.08 / 0.14 \\
Image Compression &
JPEG lossy encoding &
JPEG quality &
90 / 50 / 10 \\
\midrule
\rowcolor{lightgray}
\multicolumn{4}{l}{\textbf{Photometric transforms}} \\
\midrule
Perturbation & Method & Parameter & Parameters (Marginal / Moderate / Severe)
 \\
\midrule
Brightness Adjustment &
Global intensity scaling &
Brightness factor &
0.9 / 0.5 / 0.1 \\
Contrast Adjustment &
Global contrast scaling &
Contrast factor &
0.7 / 0.4 / 0.2 \\
\midrule
\rowcolor{lightgray}
\multicolumn{4}{l}{\textbf{Geometric / spatial transforms}} \\
\midrule
Perturbation & Method & Parameter & Parameters (Marginal / Moderate / Severe)
 \\
\midrule
Center Cropping &
Center crop and resize &
Crop ratio &
0.9 / 0.5 / 0.1 \\
Random Cropping &
Random crop and resize &
Crop ratio &
0.9 / 0.5 / 0.1 \\
\midrule
\rowcolor{lightgray}
\multicolumn{4}{l}{\textbf{Content-agnostic occlusions}} \\
\midrule
Perturbation & Method & Parameter & Parameters (Marginal / Moderate / Severe)
 \\
\midrule
Watermark Overlay &
Semi-transparent overlay &
Watermark size ratio &
0.4 / 0.6 / 0.8 \\
Random Shadowing &
Random shadow occlusion &
Shadow intensity &
0.3 / 0.6 / 0.9 \\
\bottomrule
\end{tabular}
\end{figure*}

\noindent\textbf{Does the sampling temperature affect the CSA-MIA performance?}

We conducted further experiments to examine the impact of sampling \texttt{temperature} on CSA-MIA performance. During the generation phase, we controlled the degree of randomness in the target model's decoding process by adjusting the sampling temperature parameter. As shown in the Figure~\ref{fig:main_results_grid}, while performance remained relatively stable over a wide range of non-zero values, the model exhibited greater stability at lower temperatures. This trend suggests that lower temperatures produce more deterministic and consistent descriptions, thus contributing to clearer image-text similarity signals and improving the distinguishability between member and non-member samples. Conversely, higher temperatures introduce additional randomness into the generated descriptions, increasing variability and partially masking discriminative cues. However, as long as the generated content remains semantically consistent with the image, CSA-MIA demonstrates strong robustness to variations in sampling temperature.

\noindent\textbf{Does the generation prompt affect CSA-MIA performance?} We evaluate the sensitivity of CSA-MIA to different generation prompts by comparing its performance under multiple prompts. The detailed descriptions of the prompts used in this study are summarized in Table~\ref{tab:prompt_design}. As reported in Table~\ref{tab:prompt_combined_results}, CSA-MIA consistently achieves strong performance across all prompt settings, demonstrating limited sensitivity to the specific prompt formulation. While minor performance variations can be observed across different prompts, the overall ranking remains stable, and CSA-MIA either achieves the best or second-best results in most cases. These results further indicate that CSA-MIA is robust to prompt variations and can generalize well across different prompting strategies in practical deployment scenarios. Due to space limitations, more extensive results are provided in Appendix~\ref{app:prompt}.

\subsection{Robustness Study}
\label{robustness}

In real-world infringement and unauthorized usage scenarios, images are often modified through various perturbations and editing operations before redistribution. To assess whether our method remains effective under such realistic conditions, we conduct a robustness evaluation.
Specifically, we apply a variety of common content perturbations and editing operations to the input images. As illustrated in Table~\ref{tab:perturbations}, our experiments cover nine representative scenarios, including blur, noise injection, image compression, brightness adjustment, contrast adjustment, center cropping, random cropping, watermark overlay, and random shadowing. These perturbations frequently occur in real-world infringement and redistribution processes and can significantly alter the visual appearance and local semantic structures of images. For each perturbation, we evaluate three severity levels: marginal, moderate, and severe, by varying the corresponding control parameters. The specific parameter settings for each perturbation and severity level are summarized in Table~\ref{tab:perturbations}. Examples illustrating the visual effects of these perturbations are provided in Appendix~\ref{sec:robustness_examples}.

The results are reported in Figure~\ref{fig:robustness_bar}. Overall, our method remains effective under these realistic infringement scenarios, achieving an AUC above 0.7 in all cases except for severe cropping, where only 10\% of the image content is retained. We attribute this sharp decline to the fact that such extreme cropping removes the primary semantic subject making it nearly impossible to establish a valid cross-modal alignment. In contrast, other perturbations like noise or blur maintain high AUCs even at severe levels because they leave the global semantic structure largely intact.

\section{Conclusion}
In this work, we investigate MIA against VLMs under the most restrictive yet realistic setting, namely black-box and single-sample settings. To address this setting, we propose CSA-MIA, a membership inference framework that quantifies cross-modal semantic alignment in a joint embedding space without accessing the target VLM's internal logits or relying on large-scale statistical references. Experimental results on both open-source and closed-source VLMs suggest that CSA-MIA enables effective membership inference and maintains robustness under diverse image perturbations. These findings shed light on training data leakage risks in deployed VLMs and may support the detection and mitigation of unauthorized or infringing image usage in real-world applications.

\cleardoublepage
\cleardoublepage
\appendix
\section*{Ethical Considerations}
\textbf{This work does not involve human subjects, user interaction, the collection of personal data or other identifiable ethical concerns. All experiments are conducted using publicly available or synthetic datasets in accordance with their respective licenses. The research is intended to assess potential privacy risks and support responsible and secure deployment of VLMs. We commit to conducting and disseminating this research in accordance with established ethical guidelines and responsible disclosure practices, and to avoiding any use of the proposed techniques that could intentionally harm individuals, organizations, or data owners.}

\section*{Open Science}
\textbf{We will release all artifacts needed to evaluate the contributions of this paper through an anonymous repository: \footnote{\url{https://anonymous.4open.science/r/CSA-MIA-F2B200110/}}. This repository contains: (1) the source code for our proposed method and all baseline methods used in this paper; (2) scripts for preparing datasets used in the experiments; (3) scripts for reproducing the main results; (4) configuration files for running experiments on different VLMs; and (5) utilities for calculating metrics. Reviewers can access these artifacts via the anonymous link above without registration.}
\cleardoublepage
\bibliographystyle{plainurl}
\bibliography{main}

\clearpage 

\appendix
\section*{Appendix}
\addcontentsline{toc}{section}{Appendix}
\appendix

\section{Observation}
\subsection{Qualitative Analysis}
\label{app:shili}
\begin{figure}[H]
    \centering
    \includegraphics[width=\linewidth, trim=50 0 300 0, clip]{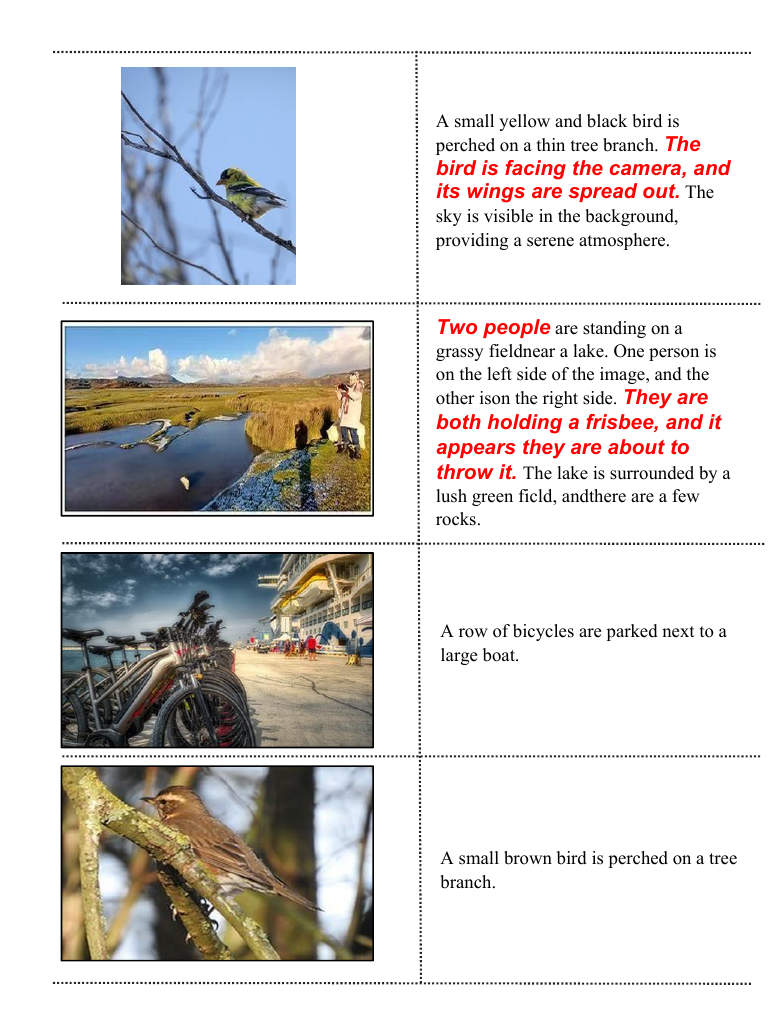}
    
    \caption{Qualitative examples of VLM descriptions on non-member images. The model fails to provide accurate details, exhibiting issues such as \textbf{hallucinations} (highlighted in red, e.g., describing non-existent actions or attributes) or generating \textbf{overly brief} captions that lack semantic depth.}
    \label{fig:vlm_limitations}
\end{figure}

Figure \ref{fig:vlm_limitations} presents qualitative examples illustrating the limitations of VLM descriptions on non-member images. As observed in the first two rows, the model frequently exhibits hallucinations and semantic errors. For instance, it fabricates non-existent objects (e.g., mentioning a camera not present in the scene), inaccurately describes the bird's actions, and suffers from recognition failures—specifically, misidentifying a camera as a frisbee and miscounting the number of people. Conversely, the bottom two rows demonstrate the issue of over-simplification, where the generated captions are excessively brief. 

\subsection{Quantitative Evaluation}
\label{app:tongji}

\begin{figure}[H]
    \centering
    \begin{subfigure}[b]{\linewidth}
        \centering
        \includegraphics[width=\linewidth]{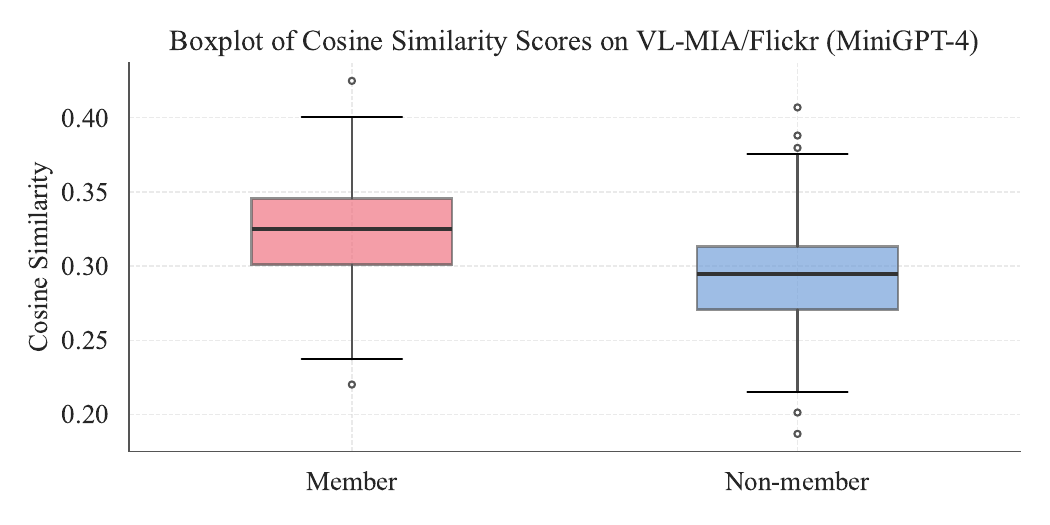}
        \caption{Boxplots of CLIP cosine similarity scores between image embeddings and Minigpt-4 generated description embeddings for 300 member and 300 non-member images.}
        \label{fig:minigpt_boxplot} 
    \end{subfigure}
    
    \vspace{0.5cm} 
    
    \begin{subfigure}[b]{\linewidth}
        \centering
        \includegraphics[width=\linewidth]{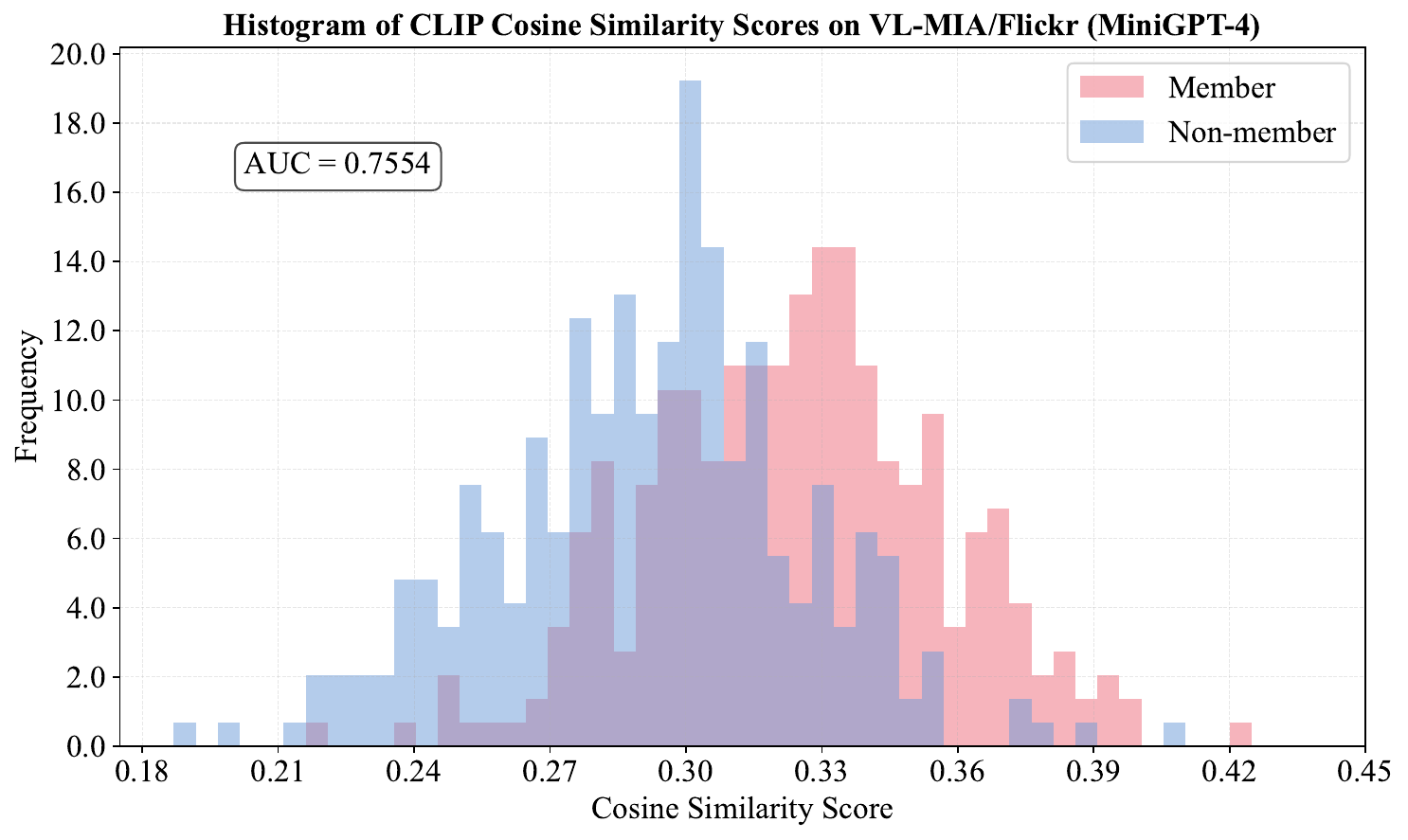}
        \caption{Histogram of CLIP cosine similarity scores between image embeddings and Minigpt-4 generated description embeddings for 300 member and 300 non-member images.}
        \label{fig:minigpt_histogram} 
    \end{subfigure}
    
    \caption{Visualization of CLIP cosine similarity scores for MiniGPT-4.}
    \label{fig:minigpt_combined}
\end{figure}

\begin{figure}[H]
    \centering
    \begin{subfigure}[b]{\linewidth}
        \centering
        \includegraphics[width=\linewidth]{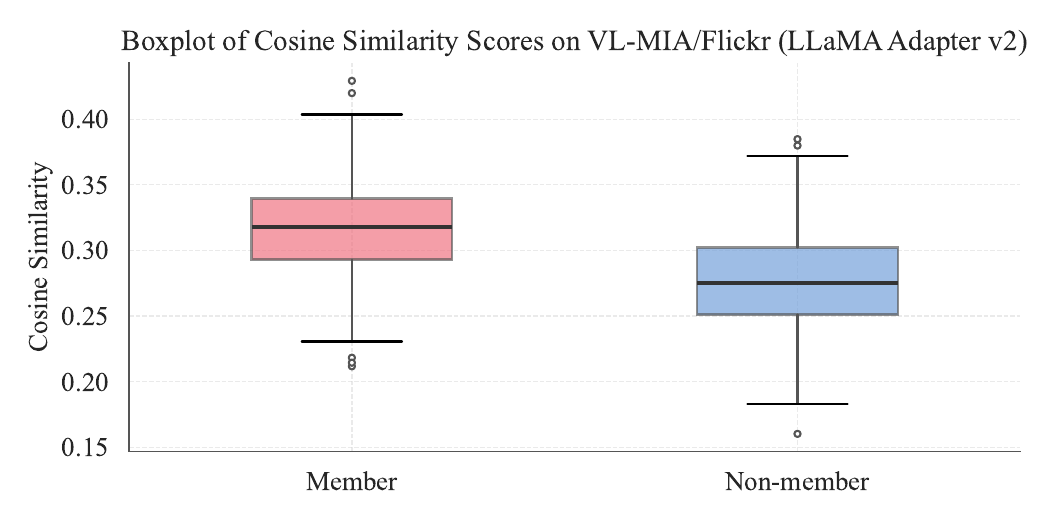}
        \caption{Boxplots of CLIP cosine similarity scores between image embeddings and LLaMA Adapter v2 generated description embeddings for 300 member and 300 non-member images.}
        \label{fig:llama_boxplot} 
    \end{subfigure}
    
    \vspace{0.5cm} 
    
    \begin{subfigure}[b]{\linewidth}
        \centering
        \includegraphics[width=\linewidth]{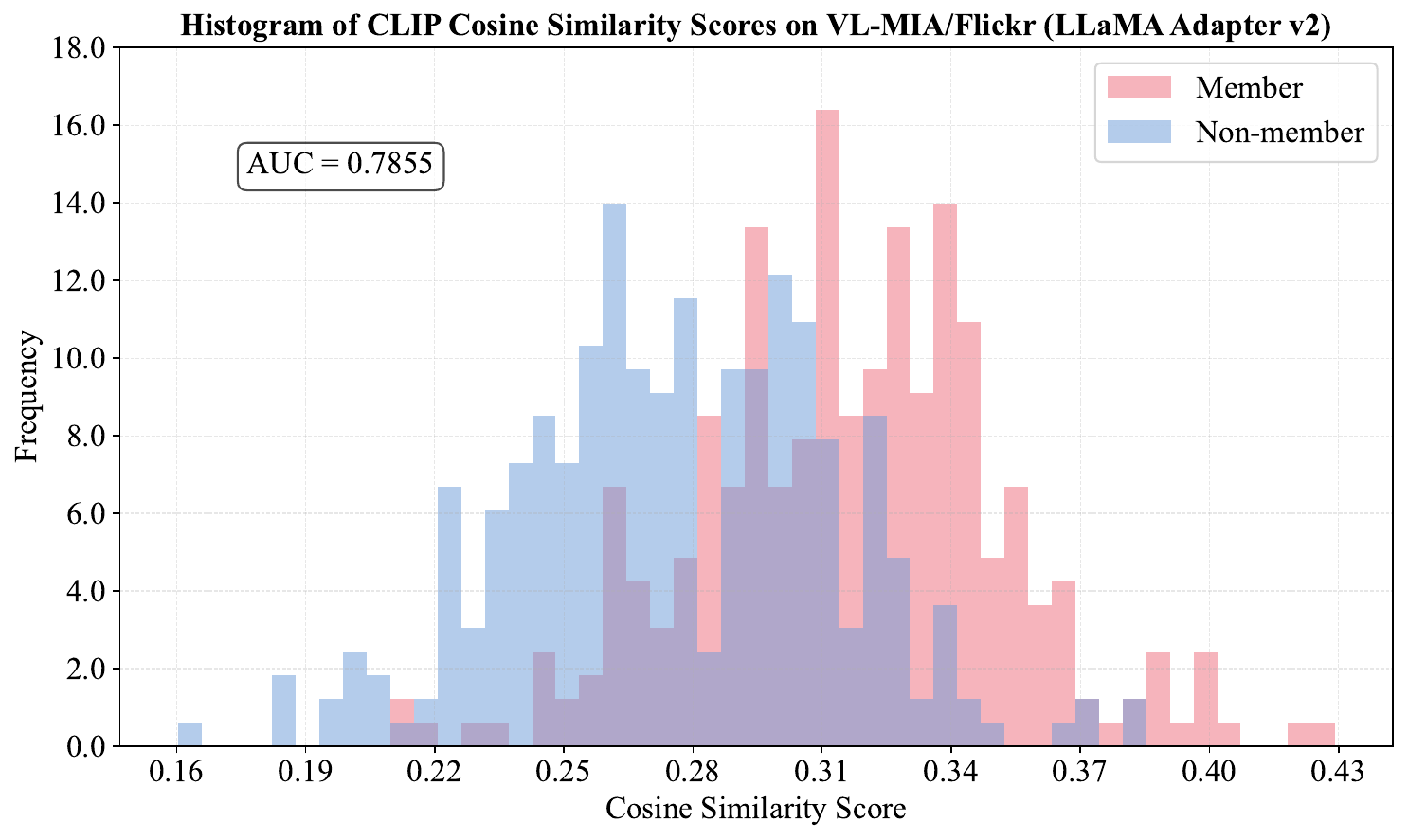}
        \caption{Histogram of CLIP cosine similarity scores between image embeddings and LLaMA Adapter v2 generated description embeddings for 300 member and 300 non-member images.}
        \label{fig:llama_adapter} 
    \end{subfigure}
    \caption{Visualization of CLIP cosine similarity scores for LLaMA Adapter v2.}
\end{figure}

We randomly selected 300 member and 300 non-member images from the VL-MIA/Flickr dataset. For each image, we generated descriptions using MiniGPT-4 and LLaMA Adapter v2, and subsequently calculated the cosine similarity between the image embeddings and the generated text embeddings extracted by CLIP. The visualization results for MiniGPT-4 are presented in Figures \ref{fig:minigpt_boxplot} and \ref{fig:minigpt_histogram}, while the corresponding results for LLaMA Adapter v2 are shown in Figures \ref{fig:llama_boxplot} and \ref{fig:llama_adapter}. Across both models, member images demonstrate a higher degree of cross-modal alignment with their generated descriptions compared to non-member images.

\section{Metrics}
\label{sec:Metrics}

\noindent \textbf{Evaluation Metrics.} We employ two standard metrics to quantitatively evaluate the performance of the membership inference attacks:

\begin{itemize}
    \item \textbf{AUC}: The Area Under the Curve (AUC) measures the entire two-dimensional area underneath the Receiver Operating Characteristic (ROC) curve. Formally, it is calculated by integrating the True Positive Rate (TPR) over the False Positive Rate (FPR) range:
    \begin{equation}
        \text{AUC} = \int_{0}^{1} \text{TPR}(\text{FPR}) \, d\text{FPR}
    \end{equation}
    The AUC value represents the probability that a randomly chosen member sample will be assigned a higher membership score than a randomly chosen non-member sample.

    \item \textbf{TPR@5\%FPR}: This metric evaluates the attack's effectiveness at a specific, stringent error tolerance level. It is calculated by identifying the classification threshold $\tau$ that results in a False Positive Rate of exactly 5\%:
    \begin{equation}
        \text{FPR}(\tau) = \frac{FP}{FP + TN} = 0.05
    \end{equation}
    Then, the corresponding True Positive Rate is measured as $\text{TPR}(\tau) = \frac{TP}{TP + FN}$. This metric is particularly crucial in practical privacy auditing where minimizing false alarms is a priority.
\end{itemize}

\section{Details of Models}
\label{sec:Models}
We evaluate our method on three representative large multimodal models. The versions, base models and training datasets of target VLMs are listed in Table~\ref{tab:model_details}.

\begin{table*}[t]
\centering
\caption{Open-sourced model details used in this work.}
\label{tab:model_details}
\setlength{\tabcolsep}{10pt}
\renewcommand{\arraystretch}{1.2}

\resizebox{\textwidth}{!}{
\begin{tabular}{lccc}
\toprule
\textbf{Model} 
& \textbf{LLaVA-1.5} 
& \textbf{MiniGPT-4} 
& \textbf{LLaMA Adapter v2} \\
\midrule

Base LLM 
& Vicuna v1.5 7B 
& LLaMA 2 Chat 7B 
& LLaMA 7B \\

Vision processor 
& CLIP-ViT-L 
& BLIP2 / eva\_vit\_g\textsuperscript{5} 
& CLIP-ViT-L \\

Image-text pretraining 
& BLIP\_LAION\_CCS 
& BLIP\_LAION\_CCS 
& Image-Text-V1\textsuperscript{6} \\

Instruction tuning 
& LLaVA-1.5 mix665k\textsuperscript{7} 
& CC\_SBU\_Align 
& GPT4LLM, LLaVA Instruct 150k, VQAv2 \\

\bottomrule
\end{tabular}
}
\end{table*}

\section{Additional Comparative Results}
\label{app:acr}
Table \ref{tab:minigpt4_600} presents the detailed experimental results on the VL-MIA/Flickr dataset using MiniGPT-4. Additionally, we report the evaluation metrics for the LLaMA Adapter v2 model on the same dataset in Table~\ref{tab:llama_adapter_performance}.
\begin{table*}[t]
\centering
\caption{Comprehensive evaluation of CSA-MIA versus baselines on the VL-MIA/Flickr dataset using MiniGPT-4. We report AUC and detailed utility metrics at strict False Positive Rate thresholds of 1\%, 5\%, and 10\%. The metrics Adv, Rec, Prec, and Acc correspond to Advantage, Recall, Precision, and Accuracy, respectively. \textbf{Bold} indicates the best AUC within each column and \underline{underline} indicates the second best result.}
\label{tab:minigpt4_600}
\resizebox{\textwidth}{!}{%
\begin{tabular}{ll c cccc cccc cccc}
\toprule
\multirow{3}{*}[-1.0ex]{\textbf{Type}} & 
\multirow{3}{*}[-1.0ex]{\textbf{Method}} & 
\multirow{3}{*}[-1.0ex]{\textbf{AUC}} 
& \multicolumn{12}{c}{\textbf{MiniGPT-4}} \\
\cmidrule(lr){4-15}
& & & \multicolumn{4}{c}{\textbf{TPR @ 1\% FPR}} 
    & \multicolumn{4}{c}{\textbf{TPR @ 5\% FPR}} 
    & \multicolumn{4}{c}{\textbf{TPR @ 10\% FPR}} \\
\cmidrule(lr){4-7} \cmidrule(lr){8-11} \cmidrule(lr){12-15}
& & & Adv & Rec & Prec & Acc 
    & Adv & Rec & Prec & Acc 
    & Adv & Rec & Prec & Acc \\
\midrule
\multirow{13}{*}{Gray-Box}
& Aug\_KL             & 0.528 & 0.013 & 0.023 & 0.700 & 0.507 & 0.030 & 0.080 & 0.632 & 0.517 & 0.007 & 0.107 & 0.525 & 0.505 \\
& Perplexity          & 0.621 & 0.027 & 0.037 & 0.786 & 0.513 & 0.060 & 0.103 & 0.705 & 0.530 & 0.093 & 0.180 & 0.675 & 0.547 \\
& Max\_Prob\_Gap      & 0.678 & 0.023 & 0.033 & 0.769 & 0.512 & 0.103 & 0.153 & 0.754 & 0.552 & 0.153 & 0.250 & 0.721 & 0.577 \\
& Min\_0\% Prob       & 0.525 & 0.007 & 0.013 & 0.667 & 0.503 & 0.000 & 0.050 & 0.500 & 0.500 & 0.013 & 0.110 & 0.532 & 0.507 \\
& Min\_10\% Prob      & 0.587 & 0.007 & 0.013 & 0.667 & 0.503 & 0.013 & 0.063 & 0.559 & 0.507 & 0.053 & 0.153 & 0.605 & 0.527 \\
& Min\_20\% Prob      & 0.614 & 0.000 & 0.010 & 0.500 & 0.500 & 0.030 & 0.070 & 0.636 & 0.515 & 0.080 & 0.180 & 0.643 & 0.540 \\
& MaxRényi\_Max\_0\%  & 0.630 & 0.007 & 0.013 & 0.800 & 0.505 & 0.017 & 0.067 & 0.571 & 0.508 & 0.037 & 0.130 & 0.582 & 0.518 \\
& MaxRényi\_Max\_10\% & 0.749 & \underline{0.053} & \underline{0.063} & 0.864 & \underline{0.527} & \underline{0.157} & \underline{0.207} & \underline{0.805} & \underline{0.578} & 0.227 & 0.323 & 0.770 & 0.613 \\
& MaxRényi\_Max\_100\%& \textbf{0.793} & \textbf{0.133} & \textbf{0.143} & \textbf{0.935} & \textbf{0.567} & \underline{0.157} & \underline{0.207} & \underline{0.805} & \underline{0.578} & \underline{0.297} & \underline{0.397} & \underline{0.799} & \underline{0.648} \\
& ModRényi            & 0.606 & 0.020 & 0.030 & 0.750 & 0.510 & 0.030 & 0.080 & 0.615 & 0.515 & 0.073 & 0.170 & 0.638 & 0.537 \\
& Min-0\% ++          & 0.710 & 0.010 & 0.020 & 0.667 & 0.505 & 0.107 & 0.157 & 0.758 & 0.553 & 0.207 & 0.307 & 0.754 & 0.603 \\
& Min-10\% ++         & 0.710 & 0.010 & 0.020 & 0.667 & 0.505 & 0.107 & 0.157 & 0.758 & 0.553 & 0.207 & 0.307 & 0.754 & 0.603 \\
& Min-20\% ++         & 0.769 & 0.023 & 0.027 & \underline{0.889} & 0.512 & 0.150 & 0.200 & 0.800 & 0.575 & 0.253 & 0.353 & 0.779 & 0.627 \\
\cmidrule(lr){1-15} 
\multirow{2}{*}{Black-Box}
& Image-only Inference & 0.617 & -0.007 & 0.003 & 0.250 & 0.497 & 0.027 & 0.077 & 0.622 & 0.515 & 0.087 & 0.187 & 0.651 & 0.543 \\
& \cellcolor{oursblue}\textbf{CSA-MIA (Ours)} 
& \cellcolor{oursblue}\underline{0.773} 
& \cellcolor{oursblue} 0.020 & \cellcolor{oursblue} 0.030 & \cellcolor{oursblue} 0.750 & \cellcolor{oursblue} 0.510 
& \cellcolor{oursblue}\textbf{0.297} & \cellcolor{oursblue}\textbf{0.347} & \cellcolor{oursblue}\textbf{0.865} & \cellcolor{oursblue}\textbf{0.635} 
& \cellcolor{oursblue}\textbf{0.343} & \cellcolor{oursblue}\textbf{0.443} & \cellcolor{oursblue}\textbf{0.816} & \cellcolor{oursblue}\textbf{0.672} \\
\bottomrule
\end{tabular}
}
\end{table*}

\begin{table*}[t]
\centering
\caption{Comprehensive evaluation of CSA-MIA versus baselines on the VL-MIA/Flickr dataset using LLaMA Adapter v2. We report AUC and detailed utility metrics at strict False Positive Rate thresholds of 1\%, 5\%, and 10\%. The metrics Adv, Rec, Prec, and Acc correspond to Advantage, Recall, Precision, and Accuracy, respectively. \textbf{Bold} indicates the best AUC within each column and \underline{underline} indicates the second best result.}
\label{tab:llama_adapter_performance}
\resizebox{\textwidth}{!}{%
\begin{tabular}{ll c cccc cccc cccc}
\toprule
\multirow{3}{*}[-1.0ex]{\textbf{Type}} & 
\multirow{3}{*}[-1.0ex]{\textbf{Method}} & 
\multirow{3}{*}[-1.0ex]{\textbf{AUC}} 
& \multicolumn{12}{c}{\textbf{LLaMA Adapter v2}} \\
\cmidrule(lr){4-15}
& & & \multicolumn{4}{c}{\textbf{TPR @ 1\% FPR}} 
    & \multicolumn{4}{c}{\textbf{TPR @ 5\% FPR}} 
    & \multicolumn{4}{c}{\textbf{TPR @ 10\% FPR}} \\
\cmidrule(lr){4-7} \cmidrule(lr){8-11} \cmidrule(lr){12-15}
& & & Adv & Rec & Prec & Acc 
    & Adv & Rec & Prec & Acc 
    & Adv & Rec & Prec & Acc \\
\midrule
\multirow{13}{*}{Gray-Box}
& Aug\_KL             & 0.491 & -0.003 & 0.000 & 0.000 & 0.498 & -0.010 & 0.033 & 0.435 & 0.495 & -0.007 & 0.093 & 0.483 & 0.497 \\
& Perplexity          & 0.645 & 0.043 & 0.053 & 0.842 & 0.522 & 0.077 & 0.127 & 0.717 & 0.538 & 0.170 & 0.270 & 0.730 & 0.585 \\
& Max\_Prob\_Gap      & 0.627 & 0.033 & 0.043 & 0.813 & 0.517 & 0.127 & 0.173 & 0.788 & 0.563 & 0.160 & 0.260 & 0.722 & 0.580 \\
& Min\_0\% Prob       & 0.612 & -0.007 & 0.003 & 0.250 & 0.497 & -0.003 & 0.047 & 0.483 & 0.498 & 0.090 & 0.190 & 0.655 & 0.545 \\
& Min\_10\% Prob      & 0.622 & -0.003 & 0.007 & 0.400 & 0.498 & 0.047 & 0.097 & 0.659 & 0.523 & 0.057 & 0.157 & 0.610 & 0.528 \\
& Min\_20\% Prob      & 0.630 & 0.013 & 0.023 & 0.700 & 0.507 & 0.070 & 0.120 & 0.706 & 0.535 & 0.083 & 0.183 & 0.647 & 0.542 \\
& MaxRényi\_Max\_0\%  & 0.597 & 0.000 & 0.003 & 0.500 & 0.500 & -0.003 & 0.040 & 0.480 & 0.498 & 0.073 & 0.173 & 0.634 & 0.537 \\
& MaxRényi\_Max\_10\% & 0.641 & 0.023 & 0.033 & 0.769 & 0.512 & 0.047 & 0.097 & 0.674 & 0.525 & 0.093 & 0.193 & 0.659 & 0.547 \\
& MaxRényi\_Max\_100\%& \underline{0.685} & 0.030 & 0.040 & 0.857 & 0.517 & \underline{0.197} & \underline{0.247} & \underline{0.832} & \underline{0.598} & \underline{0.257} & \underline{0.350} & \underline{0.790} & \underline{0.628} \\
& ModRényi            & 0.633 & \underline{0.047} & 0.053 & 0.889 & \underline{0.523} & 0.080 & 0.127 & 0.731 & 0.540 & 0.120 & 0.220 & 0.688 & 0.560 \\
& Min-0\% ++          & 0.537 & 0.003 & 0.003 & \textbf{1.000} & 0.502 & 0.030 & 0.070 & 0.636 & 0.515 & -0.007 & 0.090 & 0.482 & 0.497 \\
& Min-10\% ++         & 0.607 & \underline{0.047} & \underline{0.057} & 0.850 & \underline{0.523} & 0.070 & 0.120 & 0.706 & 0.535 & 0.123 & 0.220 & 0.695 & 0.562 \\
& Min-20\% ++         & 0.622 & 0.007 & 0.017 & 0.625 & 0.503 & 0.120 & 0.170 & 0.773 & 0.560 & 0.140 & 0.240 & 0.706 & 0.570 \\
\midrule
\multirow{2}{*}{Black-Box}
& Image-only Inference & 0.510 & -0.010 & 0.000 & 0.000 & 0.495 & -0.003 & 0.043 & 0.481 & 0.498 & 0.000 & 0.100 & 0.500 & 0.500 \\
& \cellcolor{oursblue}\textbf{CSA-MIA (Ours)} 
& \cellcolor{oursblue}\textbf{0.812} 
& \cellcolor{oursblue}\textbf{0.093} & \cellcolor{oursblue}\textbf{0.103} & \cellcolor{oursblue}\underline{0.912} & \cellcolor{oursblue}\textbf{0.547} 
& \cellcolor{oursblue}\textbf{0.207} & \cellcolor{oursblue}\textbf{0.257} & \cellcolor{oursblue}\textbf{0.837} & \cellcolor{oursblue}\textbf{0.603} 
& \cellcolor{oursblue}\textbf{0.333} & \cellcolor{oursblue}\textbf{0.433} & \cellcolor{oursblue}\textbf{0.813} & \cellcolor{oursblue}\textbf{0.667} \\
\bottomrule
\end{tabular}
}
\end{table*}

\section{Impact of Reference Set Size}
\label{sec:fine_grained_impact_ref_set_size}

In this section, we provide a more fine-grained analysis of how the size of the reference set affects the performance of CSA-MIA. To this end, we systematically vary the reference set size from 1 to 200 and evaluate the corresponding attack performance under different settings. Specifically, we consider reference set sizes of 
$\{1, 20, 40, 60, 80, 100, 120, 140, 160, 180, 200\}$, covering both small and large reference regimes.

The results reported in Table~\ref{tab:ref_size_20_200} show that CSA-MIA performance improves steadily as the reference set size increases, but gradually stabilizes beyond a moderate scale. 

\begin{table*}[!t]
\centering
\caption{Fine-grained impact of reference set size on CSA-MIA performance.}
\label{tab:ref_size_20_200}

\resizebox{0.95\textwidth}{!}{
\begin{tabular}{llcccccccccccccccccccc}
\toprule
\multirow{3}{*}{Type} &
\multirow{3}{*}{Method} &
\multicolumn{20}{c}{\textbf{Reference Set Size}} \\
\cmidrule(lr){3-22}
& &
\multicolumn{4}{c}{20} &
\multicolumn{4}{c}{40} &
\multicolumn{4}{c}{60} &
\multicolumn{4}{c}{80} &
\multicolumn{4}{c}{100} \\
\cmidrule(lr){3-6} \cmidrule(lr){7-10} \cmidrule(lr){11-14} \cmidrule(lr){15-18} \cmidrule(lr){19-22}
& & Adv & Recall & Prec & Acc
& Adv & Recall & Prec & Acc
& Adv & Recall & Prec & Acc 
& Adv & Recall & Prec & Acc 
& Adv & Recall & Prec & Acc \\
\midrule

\multirow{13}{*}{Gray-Box}
& Aug-KL
& 0.202 & 0.674 & 0.591 & 0.601
& 0.205 & 0.669 & 0.592 & 0.603
& 0.205 & 0.675 & 0.590 & 0.603
& 0.206 & 0.678 & 0.590 & 0.603
& 0.205 & 0.669 & 0.591 & 0.603 \\

& Perplexity
& 0.255 & 0.801 & 0.597 & 0.628
& 0.263 & 0.795 & 0.600 & 0.631
& 0.260 & 0.804 & 0.597 & 0.630
& 0.261 & 0.802 & 0.598 & 0.630
& 0.259 & 0.806 & 0.596 & 0.629 \\

& Max-Prob-Gap
& 0.230 & 0.724 & 0.596 & 0.617
& 0.231 & 0.723 & 0.597 & 0.617
& 0.233 & 0.724 & 0.596 & 0.616
& 0.231 & 0.724 & 0.595 & 0.616
& 0.231 & 0.720 & 0.596 & 0.616 \\

& Min-0\% Prob
& 0.190 & 0.732 & 0.577 & 0.596
& 0.197 & 0.717 & 0.580 & 0.598
& 0.199 & 0.715 & 0.581 & 0.600
& 0.199 & 0.728 & 0.579 & 0.599
& 0.199 & 0.722 & 0.580 & 0.599 \\

& Min-10\% Prob
& 0.251 & 0.769 & 0.599 & 0.626
& 0.258 & 0.782 & 0.600 & 0.629
& 0.263 & 0.785 & 0.601 & 0.631
& 0.264 & 0.786 & 0.601 & 0.632
& 0.266 & 0.784 & 0.602 & 0.633 \\

& Min-20\% Prob
& 0.246 & 0.762 & 0.598 & 0.623
& 0.252 & 0.770 & 0.599 & 0.626
& 0.253 & 0.771 & 0.599 & 0.626
& 0.255 & 0.773 & 0.600 & 0.628
& 0.258 & 0.763 & 0.602 & 0.629 \\

& MaxRényi-Max-0\%
& 0.284 & 0.739 & 0.621 & 0.642
& 0.290 & 0.747 & 0.622 & 0.645
& 0.300 & 0.752 & 0.625 & 0.650
& 0.301 & 0.754 & 0.625 & 0.650
& 0.300 & 0.750 & 0.626 & 0.650 \\

& MaxRényi-Max-10\%
& 0.325 & 0.786 & 0.632 & 0.663
& 0.329 & 0.798 & 0.631 & 0.665
& 0.330 & 0.806 & 0.629 & 0.665 
& 0.331 & 0.805 & 0.630 & 0.665 
& 0.331 & 0.813 & 0.628 & 0.665 \\

& MaxRényi-Max-100\%
& 0.312 & 0.795 & 0.624 & 0.656
& 0.313 & 0.797 & 0.623 & 0.656
& 0.313 & 0.782 & 0.626 & 0.656
& 0.313 & 0.795 & 0.623 & 0.656
& 0.312 & 0.793 & 0.623 & 0.656 \\

& ModRényi
& 0.247 & 0.748 & 0.602 & 0.624
& 0.245 & 0.769 & 0.596 & 0.623
& 0.248 & 0.765 & 0.598 & 0.624
& 0.249 & 0.768 & 0.597 & 0.625
& 0.249 & 0.770 & 0.597 & 0.625 \\

& Min-0\%++
& 0.135 & 0.753 & 0.552 & 0.568
& 0.140 & 0.749 & 0.553 & 0.570
& 0.133 & 0.760 & 0.549 & 0.567
& 0.137 & 0.752 & 0.551 & 0.569
& 0.137 & 0.754 & 0.551 & 0.569 \\

& Min-10\%++
& 0.129 & 0.799 & 0.546 & 0.564
& 0.129 & 0.807 & 0.544 & 0.565
& 0.125 & 0.807 & 0.543 & 0.563
& 0.122 & 0.814 & 0.541 & 0.561
& 0.126 & 0.811 & 0.543 & 0.563 \\

& Min-20\%++
& 0.160 & 0.743 & 0.562 & 0.580
& 0.159 & 0.744 & 0.560 & 0.579
& 0.155 & 0.749 & 0.558 & 0.577
& 0.159 & 0.744 & 0.560 & 0.579
& 0.158 & 0.750 & 0.559 & 0.579 \\

\midrule
\multirow{2}{*}{Black-Box}
& Image-only Inference
& 0.139 & 0.545 & 0.573 & 0.570
& 0.136 & 0.524 & 0.574 & 0.568
& 0.135 & 0.523 & 0.574 & 0.568
& 0.133 & 0.518 & 0.573 & 0.567
& 0.132 & 0.513 & 0.573 & 0.566 \\

& \cellcolor{oursblue}\textbf{CSA-MIA (Ours)}
& \cellcolor{oursblue}\textbf{0.386}
& \cellcolor{oursblue}\textbf{0.905}
& \cellcolor{oursblue}\textbf{0.638}
& \cellcolor{oursblue}\textbf{0.693}
& \cellcolor{oursblue}\textbf{0.395}
& \cellcolor{oursblue}\textbf{0.903}
& \cellcolor{oursblue}\textbf{0.641}
& \cellcolor{oursblue}\textbf{0.698}
& \cellcolor{oursblue}\textbf{0.397}
& \cellcolor{oursblue}\textbf{0.903}
& \cellcolor{oursblue}\textbf{0.642}
& \cellcolor{oursblue}\textbf{0.698}
& \cellcolor{oursblue}\textbf{0.396}
& \cellcolor{oursblue}\textbf{0.903}
& \cellcolor{oursblue}\textbf{0.641}
& \cellcolor{oursblue}\textbf{0.698}
& \cellcolor{oursblue}\textbf{0.396}
& \cellcolor{oursblue}\textbf{0.902}
& \cellcolor{oursblue}\textbf{0.641}
& \cellcolor{oursblue}\textbf{0.698}\\


\midrule
\multirow{3}{*}{Type} &
\multirow{3}{*}{Method} &
\multicolumn{20}{c}{\textbf{Reference Set Size}} \\
\cmidrule(lr){3-22}
& &
\multicolumn{4}{c}{120} &
\multicolumn{4}{c}{140} &
\multicolumn{4}{c}{160} &
\multicolumn{4}{c}{180} &
\multicolumn{4}{c}{200} \\
\cmidrule(lr){3-6} \cmidrule(lr){7-10} \cmidrule(lr){11-14}
\cmidrule(lr){15-18} \cmidrule(lr){19-22}

\multicolumn{2}{c}{} &
Adv & Recall & Prec & Acc
& Adv & Recall & Prec & Acc
& Adv & Recall & Prec & Acc
& Adv & Recall & Prec & Acc
& Adv & Recall & Prec & Acc \\
\midrule

\multirow{13}{*}{Gray-Box}
& Aug-KL
& 0.206 & 0.672 & 0.591 & 0.603 & 0.206 & 0.676 & 0.590 & 0.603 & 0.205 & 0.673 & 0.590 & 0.603 & 0.206 & 0.670 & 0.591 & 0.603 & 0.208 & 0.680 & 0.590 & 0.604 \\

& Perplexity
& 0.260 & 0.802 & 0.597 & 0.630 & 0.260 & 0.800 & 0.597 & 0.630 & 0.260 & 0.798 & 0.598 & 0.630 & 0.260 & 0.803 & 0.597 & 0.630 & 0.260 & 0.803 & 0.597 & 0.630 \\

& Max-Prob-Gap
& 0.232 & 0.721 & 0.596 & 0.616 & 0.233 & 0.725 & 0.596 & 0.617 & 0.232 & 0.725 & 0.596 & 0.616 & 0.233 & 0.724 & 0.596 & 0.617 & 0.233 & 0.727 & 0.596 & 0.616 \\

& Min-0\% Prob
& 0.199 & 0.726 & 0.580 & 0.600 & 0.200 & 0.726 & 0.580 & 0.600 & 0.201 & 0.725 & 0.580 & 0.600 & 0.200 & 0.730 & 0.580 & 0.600 & 0.200 & 0.727 & 0.580 & 0.600 \\

& Min-10\% Prob
& 0.266 & 0.783 & 0.603 & 0.633 & 0.267 & 0.788 & 0.602 & 0.633 & 0.268 & 0.791 & 0.602 & 0.634 & 0.267 & 0.785 & 0.603 & 0.633 & 0.268 & 0.790 & 0.602 & 0.634 \\

& Min-20\% Prob
& 0.258 & 0.767 & 0.601 & 0.629 & 0.260 & 0.768 & 0.602 & 0.630 & 0.259 & 0.771 & 0.601 & 0.630 & 0.260 & 0.768 & 0.602 & 0.630 & 0.258 & 0.772 & 0.601 & 0.629 \\

& MaxRényi-Max-0\%
& 0.302 & 0.755 & 0.625 & 0.651 & 0.306 & 0.754 & 0.628 & 0.653 & 0.307 & 0.755 & 0.628 & 0.653 & 0.304 & 0.755 & 0.626 & 0.652 & 0.305 & 0.756 & 0.627 & 0.653 \\

& MaxRényi-Max-10\%
& 0.333 & 0.809 & 0.630 & 0.666 & 0.332 & 0.806 & 0.630 & 0.666 & 0.332 & 0.805 & 0.630 & 0.666 & 0.334 & 0.808 & 0.631 & 0.667 & 0.332 & 0.812 & 0.628 & 0.666 \\ 

& MaxRényi-Max-100\%
& 0.312 & 0.791 & 0.623 & 0.656 & 0.313 & 0.790 & 0.624 & 0.656 & 0.313 & 0.790 & 0.624 & 0.656 & 0.311 & 0.791 & 0.622 & 0.655 & 0.314 & 0.787 & 0.625 & 0.657 \\

& ModRényi
& 0.248 & 0.768 & 0.597 & 0.624 & 0.247 & 0.772 & 0.595 & 0.623 & 0.251 & 0.765 & 0.599 & 0.626 & 0.253 & 0.767 & 0.599 & 0.626 & 0.254 & 0.763 & 0.600 & 0.627 \\

& Min-0\%++
& 0.136 & 0.757 & 0.550 & 0.568 & 0.134 & 0.762 & 0.549 & 0.567 & 0.135 & 0.761 & 0.549 & 0.567 & 0.137 & 0.757 & 0.550 & 0.568 & 0.137 & 0.757 & 0.550 & 0.568 \\

& Min-10\%++
& 0.120 & 0.814 & 0.540 & 0.560 & 0.123 & 0.814 & 0.541 & 0.562 & 0.125 & 0.813 & 0.542 & 0.562 & 0.122 & 0.815 & 0.541 & 0.561 & 0.123 & 0.814 & 0.541 & 0.562 \\

& Min-20\%++
& 0.156 & 0.749 & 0.558 & 0.578 & 0.156 & 0.752 & 0.558 & 0.578 & 0.160 & 0.748 & 0.560 & 0.580 & 0.157 & 0.752 & 0.558 & 0.579 & 0.161 & 0.750 & 0.560 & 0.580 \\

\midrule
\multirow{2}{*}{Black-Box}
& Image-only Inference
& 0.132 & 0.516 & 0.573 & 0.566 & 0.131 & 0.512 & 0.573 & 0.565 & 0.133 & 0.515 & 0.574 & 0.567 & 0.133 & 0.515 & 0.574 & 0.566 & 0.137 & 0.520 & 0.576 & 0.568 \\

& \cellcolor{oursblue}\textbf{CSA-MIA (Ours)}
& \cellcolor{oursblue}\textbf{0.393} & \cellcolor{oursblue}\textbf{0.903} & \cellcolor{oursblue}\textbf{0.639} & \cellcolor{oursblue}\textbf{0.697}
& \cellcolor{oursblue}\textbf{0.393} & \cellcolor{oursblue}\textbf{0.902} & \cellcolor{oursblue}\textbf{0.640} & \cellcolor{oursblue}\textbf{0.697}
& \cellcolor{oursblue}\textbf{0.393} & \cellcolor{oursblue}\textbf{0.903} & \cellcolor{oursblue}\textbf{0.639} & \cellcolor{oursblue}\textbf{0.696}
& \cellcolor{oursblue}\textbf{0.392} & \cellcolor{oursblue}\textbf{0.903} & \cellcolor{oursblue}\textbf{0.639} & \cellcolor{oursblue}\textbf{0.696}
& \cellcolor{oursblue}\textbf{0.392} & \cellcolor{oursblue}\textbf{0.902} & \cellcolor{oursblue}\textbf{0.639} & \cellcolor{oursblue}\textbf{0.696} \\

\bottomrule
\end{tabular}
}
\end{table*}

\section{A Complete Analysis of Threshold Selection Results}
\label{sec:complete_threshold_results}
This appendix details the performance comparisons on the VL-MIA/Flickr-2k and VL-MIA/Flickr-10k benchmarks, substantiating our analysis in the main text regarding the effectiveness of threshold selection across varied model architectures and attack scenarios. The full results provided in Table~\ref{tab:threshold_result_flickr_2k_10k} allow for a detailed examination of the proposed method's performance and robustness across datasets of different scales. Specifically, CSA-MIA achieves the best or runner-up results across both models and diverse evaluation metrics in the majority of cases.

\begin{table*}[t]
\centering
\caption{Threshold selection performance on VL-MIA benchmarks (VL-MIA/Flickr\_2k and VL-MIA/Flickr\_10k).
The best results are shown in \textbf{bold}.}
\label{tab:threshold_result_flickr_2k_10k}

\resizebox{0.85\linewidth}{!}{
\begin{tabular}{llcccccccccccc}
\toprule
\multirow{3}{*}{Type} &
\multirow{3}{*}{Method} &
\multicolumn{12}{c}{\textbf{VL-MIA/Flickr\_2k}} \\
\cmidrule(lr){3-14}
& &
\multicolumn{4}{c}{LLaVA-1.5} &
\multicolumn{4}{c}{Minigpt4} &
\multicolumn{4}{c}{LLaMA adapter v2} \\
\cmidrule(lr){3-6} \cmidrule(lr){7-10} \cmidrule(lr){11-14}
& & Adv & Recall & Precision & Acc
& Adv & Recall & Precision & Acc
& Adv & Recall & Precision & Acc \\
\midrule

\multirow{13}{*}{Gray-Box}
& Aug-KL
& 0.096 & 0.592 & 0.545 & 0.548
& -0.010 & 0.361 & 0.493 & 0.495
& -0.028 & 0.300 & 0.478 & 0.486 \\

& Perplexity
& 0.251 & 0.798 & 0.594 & 0.626
& 0.164 & 0.695 & 0.567 & 0.582
& 0.182 & 0.763 & 0.568 & 0.591 \\

& Max-Prob-Gap
& 0.215 & 0.712 & 0.590 & 0.608
& 0.117 & 0.631 & 0.551 & 0.559
& 0.201 & 0.700 & 0.584 & 0.601 \\

& Min-0\% Prob
& 0.131 & 0.669 & 0.554 & 0.565
& 0.082 & 0.669 & 0.533 & 0.541
& 0.118 & 0.703 & 0.546 & 0.559 \\

& Min-10\% Prob
& 0.173 & 0.705 & 0.570 & 0.586
& 0.118 & 0.704 & 0.546 & 0.559
& 0.160 & 0.761 & 0.559 & 0.580 \\

& Min-20\% Prob
& 0.192 & 0.726 & 0.577 & 0.596
& 0.122 & 0.662 & 0.551 & 0.561
& 0.169 & 0.714 & 0.567 & 0.585 \\

& MaxRényi-Max-0\%
& 0.243 & 0.739 & 0.599 & 0.622
& 0.032 & 0.568 & 0.515 & 0.516
& 0.175 & 0.634 & 0.580 & 0.588 \\

& MaxRényi-Max-10\%
& 0.268 & 0.772 & 0.606 & 0.634
& 0.122 & 0.568 & 0.560 & 0.561
& 0.184 & 0.719 & 0.573 & 0.592 \\

& MaxRényi-Max-100\%
& 0.274 & 0.797 & 0.604 & 0.637
& 0.176 & 0.661 & 0.577 & 0.588
& 0.187 & 0.764 & 0.570 & 0.594 \\

& ModRényi
& 0.249 & 0.775 & 0.596 & 0.625
& 0.137 & 0.702 & 0.554 & 0.569
& 0.177 & 0.746 & 0.567 & 0.589 \\

& Min-0\%++
& 0.142 & 0.615 & 0.565 & 0.571
& 0.029 & 0.152 & 0.553 & 0.515
& -0.004 & 0.645 & 0.499 & 0.498 \\

& Min-10\%++
& 0.177 & 0.672 & 0.576 & 0.588
& 0.015 & 0.302 & 0.513 & 0.508
& 0.125 & \textbf{0.889} & 0.538 & 0.563 \\

& Min-20\%++
& 0.193 & 0.770 & 0.572 & 0.596
& 0.093 & 0.632 & 0.540 & 0.547
& 0.165 & 0.740 & 0.563 & 0.583 \\

\midrule
\multirow{2}{*}{Black-Box}
& Image-only Inference
& 0.076 & 0.484 & 0.543 & 0.538
& 0.120 & 0.596 & 0.556 & 0.560
& -0.164 & 0.307 & 0.393 & 0.418 \\

& \cellcolor{oursblue}CSA-MIA (Ours)
& \cellcolor{oursblue}\textbf{0.386}
& \cellcolor{oursblue}\textbf{0.887}
& \cellcolor{oursblue}\textbf{0.640}
& \cellcolor{oursblue}\textbf{0.693}
& \cellcolor{oursblue}\textbf{0.297}
& \cellcolor{oursblue}\textbf{0.828}
& \cellcolor{oursblue}\textbf{0.610}
& \cellcolor{oursblue}\textbf{0.649}
& \cellcolor{oursblue}\textbf{0.353}
& \cellcolor{oursblue}0.858
& \cellcolor{oursblue}\textbf{0.630}
& \cellcolor{oursblue}\textbf{0.677} \\

\midrule
\multirow{3}{*}{Type} &
\multirow{3}{*}{Method} &
\multicolumn{12}{c}{\textbf{VL-MIA/Flickr\_10k}} \\
\cmidrule(lr){3-14}
& &
\multicolumn{4}{c}{LLaVA-1.5} &
\multicolumn{4}{c}{Minigpt4} &
\multicolumn{4}{c}{LLaMA Adapter v2} \\
\cmidrule(lr){3-6} \cmidrule(lr){7-10} \cmidrule(lr){11-14}
& & Adv & Recall & Precision & Acc
& Adv & Recall & Precision & Acc
& Adv & Recall & Precision & Acc \\
\midrule

\multirow{13}{*}{Gray-Box}
& Aug-KL
& 0.125 & 0.632 & 0.556 & 0.563
& -0.040 & 0.359 & 0.474 & 0.480
& -0.055 & 0.329 & 0.462 & 0.473 \\

& Perplexity
& 0.257 & 0.808 & 0.595 & 0.629
& 0.157 & 0.757 & 0.558 & 0.579
& 0.205 & 0.759 & 0.578 & 0.602 \\

& Max-Prob-Gap
& 0.233 & 0.721 & 0.597 & 0.617
& 0.128 & 0.590 & 0.561 & 0.564
& 0.175 & 0.712 & 0.570 & 0.588 \\

& Min-0\% Prob
& 0.148 & 0.673 & 0.562 & 0.574
& 0.048 & 0.567 & 0.522 & 0.524
& 0.111 & 0.708 & 0.542 & 0.555 \\

& Min-10\% Prob
& 0.194 & 0.725 & 0.578 & 0.597
& 0.081 & 0.680 & 0.532 & 0.540
& 0.169 & 0.718 & 0.567 & 0.585 \\

& Min-20\% Prob
& 0.211 & 0.742 & 0.583 & 0.606
& 0.112 & 0.608 & 0.551 & 0.556
& 0.192 & 0.709 & 0.578 & 0.596 \\

& MaxRényi-Max-0\%
& 0.240 & 0.733 & 0.599 & 0.620
& 0.066 & 0.628 & 0.528 & 0.533
& 0.155 & 0.683 & 0.564 & 0.578 \\

& MaxRényi-Max-10\%
& 0.263 & 0.749 & 0.607 & 0.631
& 0.126 & 0.674 & 0.552 & 0.563
& 0.188 & 0.665 & 0.582 & 0.594 \\

& MaxRényi-Max-100\%
& 0.284 & 0.814 & 0.606 & 0.642
& 0.211 & 0.729 & 0.585 & 0.605
& 0.209 & 0.784 & 0.577 & 0.604 \\

& ModRényi
& 0.263 & 0.788 & 0.601 & 0.631
& 0.148 & 0.658 & 0.563 & 0.574
& 0.212 & 0.716 & 0.587 & 0.606 \\

& Min-0\%++
& 0.098 & 0.733 & 0.536 & 0.549
& 0.042 & 0.182 & 0.565 & 0.521
& -0.011 & 0.636 & 0.496 & 0.495 \\

& Min-10\%++
& 0.151 & 0.823 & 0.551 & 0.576
& 0.044 & 0.168 & 0.576 & 0.522
& 0.081 & 0.833 & 0.526 & 0.540 \\

& Min-20\%++
& 0.197 & 0.772 & 0.574 & 0.598
& 0.107 & 0.757 & 0.538 & 0.554
& 0.130 & 0.794 & 0.545 & 0.565 \\

\midrule
\multirow{2}{*}{Black-Box}
& Image-only Inference
& 0.112 & 0.530 & 0.559 & 0.556
& -0.038 & 0.444 & 0.479 & 0.481
& -0.133 & 0.322 & 0.413 & 0.434 \\

& \cellcolor{oursblue}CSA-MIA (Ours)
& \cellcolor{oursblue}\textbf{0.351}
& \cellcolor{oursblue}\textbf{0.887}
& \cellcolor{oursblue}\textbf{0.640}
& \cellcolor{oursblue}\textbf{0.693}
& \cellcolor{oursblue}\textbf{0.293}
& \cellcolor{oursblue}\textbf{0.800}
& \cellcolor{oursblue}\textbf{0.613}
& \cellcolor{oursblue}\textbf{0.646}
& \cellcolor{oursblue}\textbf{0.337}
& \cellcolor{oursblue}\textbf{0.841}
& \cellcolor{oursblue}\textbf{0.626}
& \cellcolor{oursblue}\textbf{0.668} \\

\bottomrule
\end{tabular}
}
\end{table*}

\section{Additional Results on Prompts}
\label{app:prompt}
We provide the extended prompt sensitivity results across three models. Table \ref{tab:llava_tpr} reports the TPR metrics for LLaVA-1.5. Tables \ref{tab:minigpt_auc} and \ref{tab:minigpt_tpr} present the results for MiniGPT-4, while Tables \ref{tab:llama_auc} and \ref{tab:llama_tpr} cover LLaMA Adapter v2. Across all these settings, CSA-MIA consistently achieves superior performance.
\begin{table*}[!htbp]
\centering

\begin{minipage}[t]{0.48\linewidth}
    \centering
    \caption{Comparison of TPR@5\%FPR performance on LLaVA-1.5.}
    \label{tab:llava_tpr}
    \resizebox{\linewidth}{!}{
        \begin{tabular}{ll ccc}
        \toprule
        \textbf{Type} & \textbf{Method} & \textbf{Prompt 1} & \textbf{Prompt 2} & \textbf{Prompt 3} \\
        \midrule
        \multirow{13}{*}{Gray-Box}
        & Aug\_KL              & 0.073 & 0.110 & 0.070 \\
        & Perplexity           & 0.190 & 0.163 & 0.180 \\
        & Max\_Prob\_Gap       & 0.213 & 0.150 & 0.150 \\
        & Min\_0\% Prob        & 0.113 & 0.117 & 0.120 \\
        & Min\_10\% Prob       & 0.113 & 0.127 & 0.110 \\
        & Min\_20\% Prob       & 0.153 & 0.093 & 0.140 \\
        & MaxRényi\_Max\_0\%   & 0.177 & 0.147 & 0.200 \\
        & MaxRényi\_Max\_10\%  & 0.207 & \textbf{0.280} & 0.250 \\
        & MaxRényi\_Max\_100\% & \underline{0.223} & 0.207 & \underline{0.290} \\
        & ModRényi             & 0.170 & 0.163 & 0.180 \\
        & Min-0\% ++           & 0.057 & 0.100 & 0.080 \\
        & Min-10\% ++          & 0.127 & 0.120 & 0.180 \\
        & Min-20\% ++          & 0.203 & 0.217 & 0.170 \\
        \midrule
        \multirow{2}{*}{Black-Box}
        & Image-only Inference & 0.096 & 0.120 & 0.087 \\
        & \cellcolor{oursblue}\textbf{CSA-MIA (Ours)} 
        & \cellcolor{oursblue}\textbf{0.327} & \cellcolor{oursblue}\underline{0.263} & \cellcolor{oursblue}\textbf{0.303} \\
        \bottomrule
        \end{tabular}
    }
\end{minipage}
\hfill
\begin{minipage}[t]{0.48\linewidth}
    \centering
    \caption{Comparison of AUC performance on MiniGPT-4.}
    \label{tab:minigpt_auc}
    \resizebox{\linewidth}{!}{
        \begin{tabular}{ll ccc}
        \toprule
        \textbf{Type} & \textbf{Method} & \textbf{Prompt 1} & \textbf{Prompt 2} & \textbf{Prompt 3} \\
        \midrule
        \multirow{13}{*}{Gray-Box}
        & Aug\_KL              & 0.467 & 0.520 & 0.493 \\
        & Perplexity           & 0.625 & 0.661 & 0.604 \\
        & Max\_Prob\_Gap       & 0.661 & 0.659 & 0.663 \\
        & Min\_0\% Prob        & 0.550 & 0.556 & 0.530 \\
        & Min\_10\% Prob       & 0.600 & 0.602 & 0.570 \\
        & Min\_20\% Prob       & 0.620 & 0.637 & 0.592 \\
        & MaxRényi\_Max\_0\%   & 0.631 & 0.633 & 0.621 \\
        & MaxRényi\_Max\_10\%  & \underline{0.708} & 0.675 & 0.634 \\
        & MaxRényi\_Max\_100\% & 0.667 & \textbf{0.750} & 0.669 \\
        & ModRényi             & 0.611 & 0.647 & 0.591 \\
        & Min-0\% ++           & 0.653 & 0.628 & \underline{0.721} \\
        & Min-10\% ++          & 0.664 & 0.650 & \underline{0.721} \\
        & Min-20\% ++          & 0.660 & 0.658 & 0.665 \\
        \midrule
        \multirow{2}{*}{Black-Box}
        & Image-only Inference & 0.569 & 0.552 & 0.580 \\
        & \cellcolor{oursblue}\textbf{CSA-MIA (Ours)} 
        & \cellcolor{oursblue}\textbf{0.744} & \cellcolor{oursblue}\underline{0.741} & \cellcolor{oursblue}\textbf{0.751} \\
        \bottomrule
        \end{tabular}
    }
\end{minipage}

\vspace{1.5em}

\begin{minipage}[t]{0.48\linewidth}
    \centering
    \caption{Comparison of TPR@5\%FPR performance on MiniGPT-4.}
    \label{tab:minigpt_tpr}
    \resizebox{\linewidth}{!}{
        \begin{tabular}{ll ccc}
        \toprule
        \textbf{Type} & \textbf{Method} & \textbf{Prompt 1} & \textbf{Prompt 2} & \textbf{Prompt 3} \\
        \midrule
        \multirow{13}{*}{Gray-Box}
        & Aug\_KL              & 0.063 & 0.077 & 0.057 \\
        & Perplexity           & 0.120 & 0.167 & 0.107 \\
        & Max\_Prob\_Gap       & 0.167 & 0.180 & 0.177 \\
        & Min\_0\% Prob        & 0.100 & 0.087 & 0.083 \\
        & Min\_10\% Prob       & 0.093 & 0.090 & 0.073 \\
        & Min\_20\% Prob       & 0.090 & 0.123 & 0.117 \\
        & MaxRényi\_Max\_0\%   & 0.140 & 0.063 & 0.127 \\
        & MaxRényi\_Max\_10\%  & \underline{0.183} & 0.130 & 0.127 \\
        & MaxRényi\_Max\_100\% & 0.177 & \underline{0.197} & 0.183 \\
        & ModRényi             & 0.117 & 0.157 & 0.103 \\
        & Min-0\% ++           & 0.087 & 0.077 & \underline{0.187} \\
        & Min-10\% ++          & 0.093 & 0.083 & \underline{0.187} \\
        & Min-20\% ++          & 0.137 & 0.090 & 0.150 \\
        \midrule
        \multirow{2}{*}{Black-Box}
        & Image-only Inference & 0.113 & 0.056 & 0.097 \\
        & \cellcolor{oursblue}\textbf{CSA-MIA (Ours)} 
        & \cellcolor{oursblue}\textbf{0.187} & \cellcolor{oursblue}\textbf{0.227} & \cellcolor{oursblue}\textbf{0.233} \\
        \bottomrule
        \end{tabular}
    }
\end{minipage}
\hfill
\begin{minipage}[t]{0.48\linewidth}
    \centering
    \caption{Comparison of AUC performance on LLaMA Adapter v2.}
    \label{tab:llama_auc}
    \resizebox{\linewidth}{!}{
        \begin{tabular}{ll ccc}
        \toprule
        \textbf{Type} & \textbf{Method} & \textbf{Prompt 1} & \textbf{Prompt 2} & \textbf{Prompt 3} \\
        \midrule
        \multirow{13}{*}{Gray-Box}
        & Aug\_KL              & 0.518 & 0.456 & 0.441 \\
        & Perplexity           & 0.585 & 0.652 & 0.596 \\
        & Max\_Prob\_Gap       & 0.588 & 0.638 & 0.572 \\
        & Min\_0\% Prob        & 0.573 & 0.628 & 0.568 \\
        & Min\_10\% Prob       & 0.591 & 0.664 & 0.618 \\
        & Min\_20\% Prob       & 0.592 & 0.662 & 0.619 \\
        & MaxRényi\_Max\_0\%   & 0.576 & 0.596 & 0.590 \\
        & MaxRényi\_Max\_10\%  & 0.646 & 0.672 & \underline{0.669} \\
        & MaxRényi\_Max\_100\% & \underline{0.654} & \underline{0.738} & 0.663 \\
        & ModRényi             & 0.572 & 0.637 & 0.584 \\
        & Min-0\% ++           & 0.518 & 0.601 & 0.554 \\
        & Min-10\% ++          & 0.584 & 0.607 & 0.594 \\
        & Min-20\% ++          & 0.602 & 0.624 & 0.611 \\
        \midrule
        \multirow{2}{*}{Black-Box}
        & Image-only Inference & 0.475 & 0.469 & 0.481 \\
        & \cellcolor{oursblue}\textbf{CSA-MIA (Ours)} 
        & \cellcolor{oursblue}\textbf{0.788} & \cellcolor{oursblue}\textbf{0.800} & \cellcolor{oursblue}\textbf{0.785} \\
        \bottomrule
        \end{tabular}
    }
\end{minipage}

\vspace{1.5em}

\begin{minipage}[t]{0.48\linewidth}
    \centering
    \caption{Comparison of TPR@5\%FPR performance on LLaMA Adapter v2.}
    \label{tab:llama_tpr}
    \resizebox{\linewidth}{!}{
        \begin{tabular}{ll ccc}
        \toprule
        \textbf{Type} & \textbf{Method} & \textbf{Prompt 1} & \textbf{Prompt 2} & \textbf{Prompt 3} \\
        \midrule
        \multirow{13}{*}{Gray-Box}
        & Aug\_KL              & 0.040 & 0.030 & 0.030 \\
        & Perplexity           & 0.150 & 0.223 & 0.143 \\
        & Max\_Prob\_Gap       & 0.170 & 0.177 & 0.160 \\
        & Min\_0\% Prob        & 0.063 & 0.150 & 0.090 \\
        & Min\_10\% Prob       & 0.117 & 0.197 & 0.133 \\
        & Min\_20\% Prob       & 0.160 & 0.173 & 0.133 \\
        & MaxRényi\_Max\_0\%   & 0.080 & 0.120 & 0.093 \\
        & MaxRényi\_Max\_10\%  & 0.147 & 0.167 & 0.150 \\
        & MaxRényi\_Max\_100\% & \underline{0.260} & \textbf{0.307} & \textbf{0.263} \\
        & ModRényi             & 0.130 & 0.210 & 0.140 \\
        & Min-0\% ++           & 0.057 & 0.057 & 0.060 \\
        & Min-10\% ++          & 0.103 & 0.103 & 0.090 \\
        & Min-20\% ++          & 0.143 & 0.167 & 0.147 \\
        \midrule
        \multirow{2}{*}{Black-Box}
        & Image-only Inference & 0.050 & 0.057 & 0.023 \\
        & \cellcolor{oursblue}\textbf{CSA-MIA (Ours)} 
        & \cellcolor{oursblue}\textbf{0.287} & \cellcolor{oursblue}\underline{0.283} & \cellcolor{oursblue}\underline{0.243} \\
        \bottomrule
        \end{tabular}
    }
\end{minipage}
\hfill
\begin{minipage}[t]{0.48\linewidth}
\end{minipage}

\end{table*}

\section{Robustness Examples}
\label{sec:robustness_examples}
To provide a clear visualization of the image processing operations discussed in this study, Table~\ref{tab:robustness_appendix_optimized} displays examples of the nine types of perturbations. These examples are categorized into three severity levels: Marginal, Moderate, and Severe. 

\newcommand{\imgcell}[1]{%
  \includegraphics[width=0.17\textwidth,keepaspectratio]{#1}%
}

\begin{table*}[!t]
\centering
\caption{Examples of corrupted images under different perturbations and severity levels.}
\label{tab:robustness_appendix_optimized}

\setlength{\tabcolsep}{2pt} 
\renewcommand{\arraystretch}{0.8} 
\scriptsize 

\resizebox{\textwidth}{!}{%
\begin{tabular}{l c c c c c}
\toprule
\textbf{Severity} & \textbf{Original} & \textbf{Blur} & \textbf{Noise} & \textbf{JPEG} & \textbf{Brightness} \\
\midrule
\textbf{Marginal} &
\imgcell{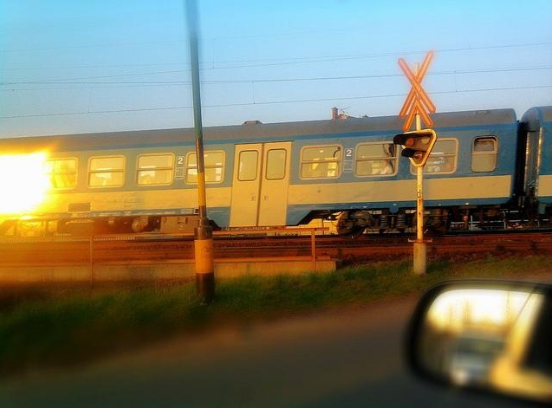} &
\imgcell{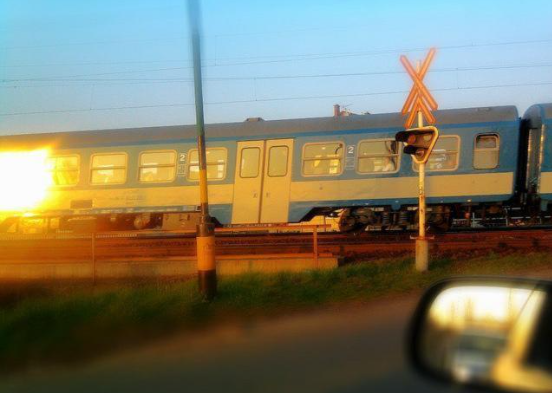} &
\imgcell{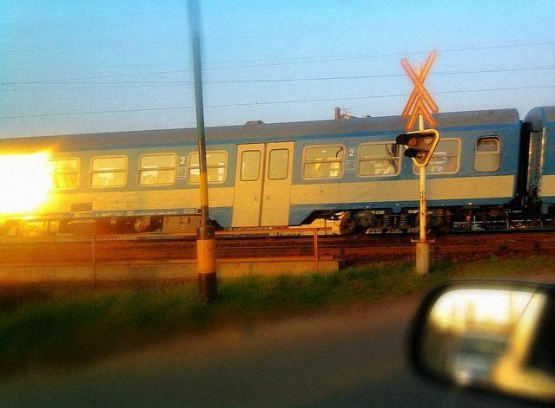} &
\imgcell{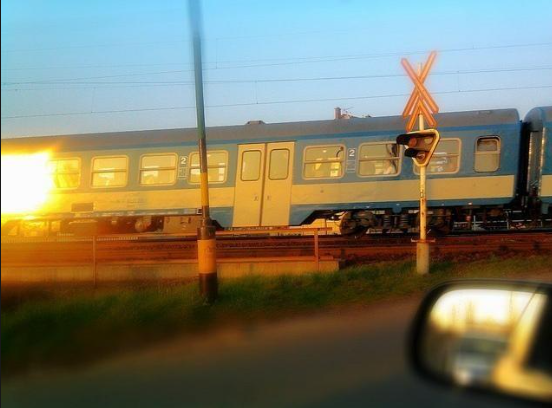} &
\imgcell{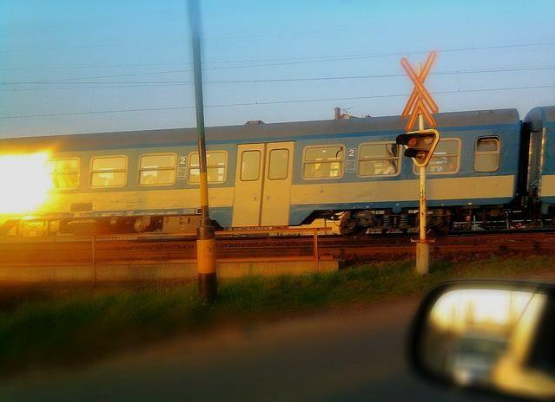} \\
\textbf{Moderate} &
\imgcell{pic/robustness_examples/original.png} &
\imgcell{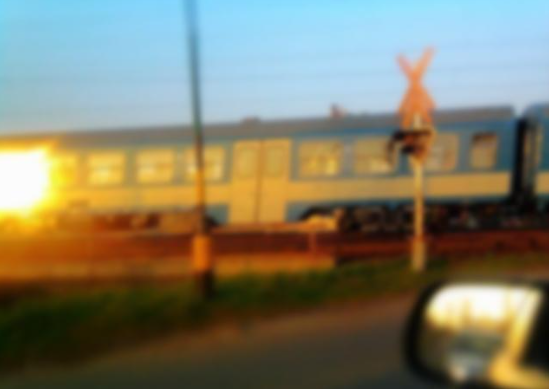} &
\imgcell{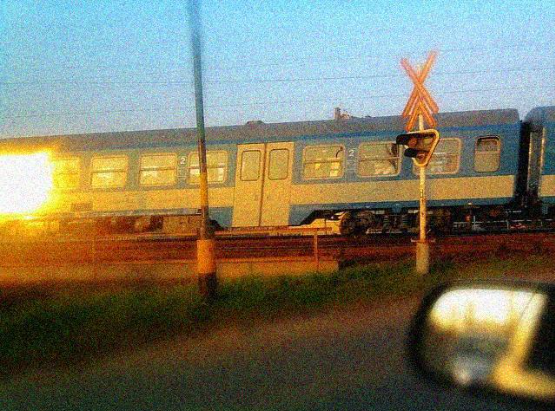} &
\imgcell{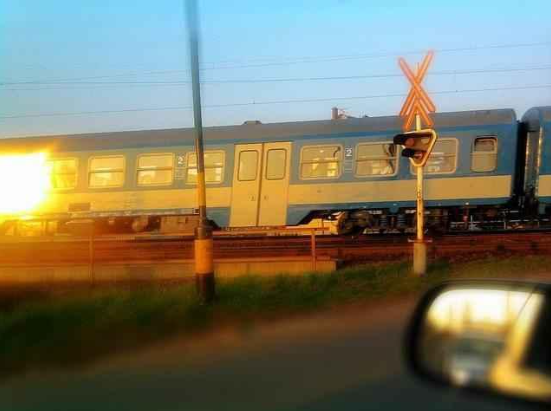} &
\imgcell{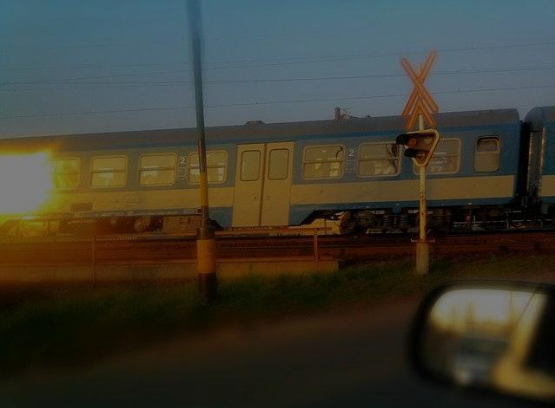} \\
\textbf{Severe} &
\imgcell{pic/robustness_examples/original.png} &
\imgcell{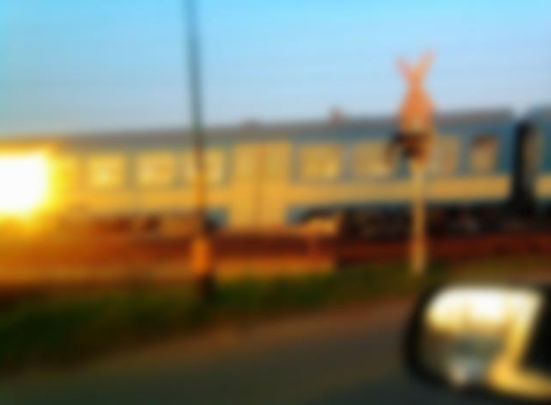} &
\imgcell{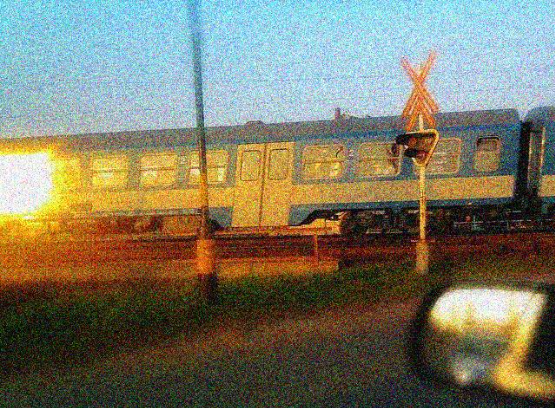} &
\imgcell{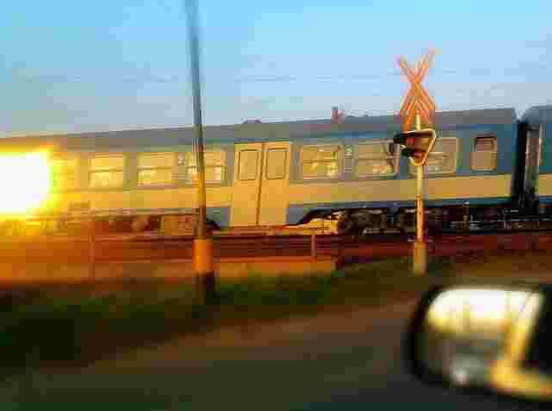} &
\imgcell{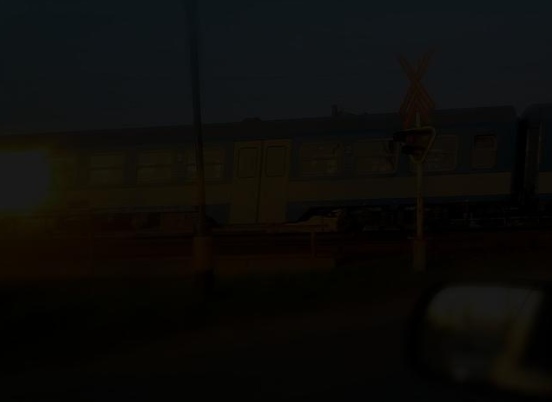} \\
\bottomrule
\end{tabular}
}

\vspace{5pt} 

\resizebox{\textwidth}{!}{%
\begin{tabular}{l c c c c c}
\toprule
\textbf{Severity} & \textbf{Contrast} & \textbf{Center crop} & \textbf{Random crop} & \textbf{Watermark} & \textbf{Shadow} \\
\midrule
\textbf{Marginal} &
\imgcell{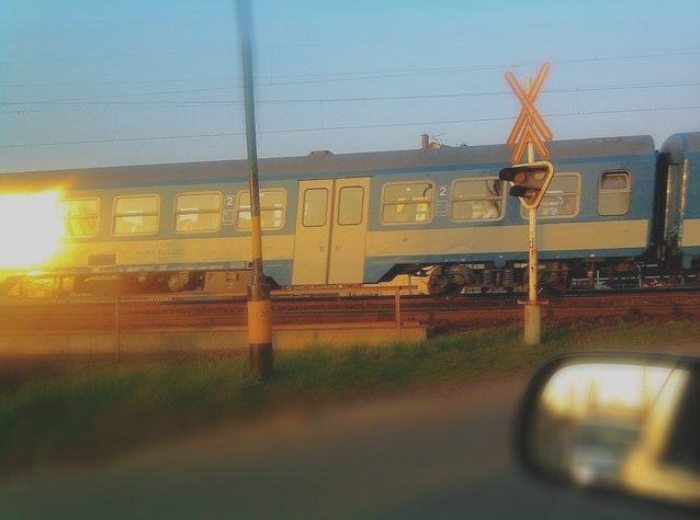} &
\imgcell{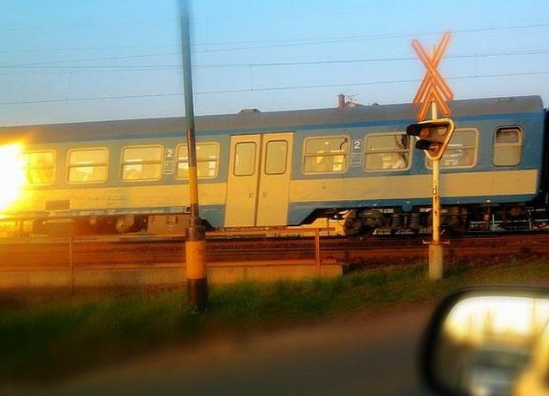} &
\imgcell{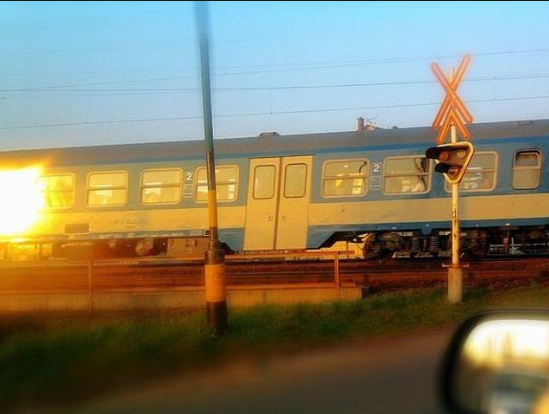} &
\imgcell{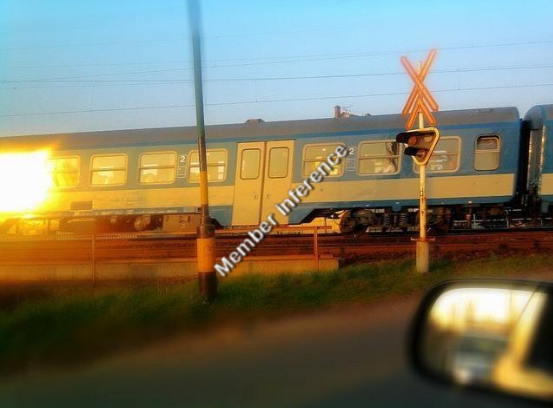} &
\imgcell{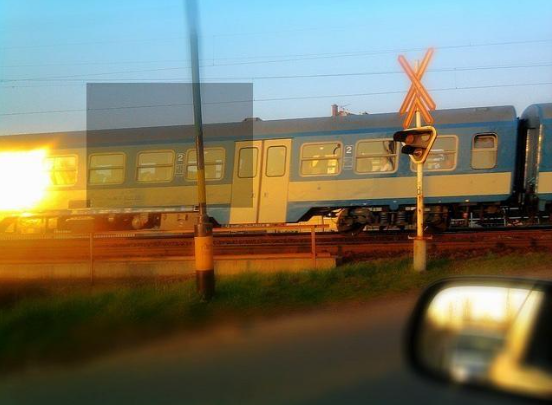} \\
\textbf{Moderate} &
\imgcell{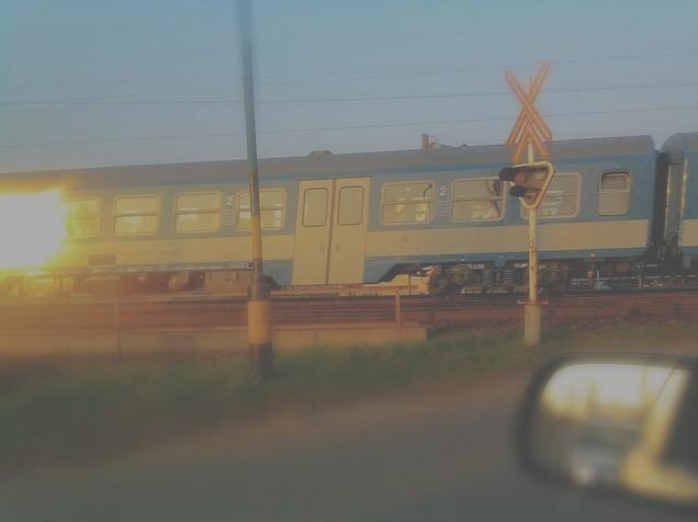} &
\imgcell{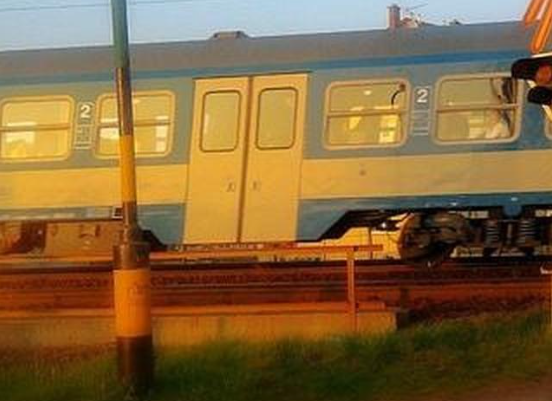} &
\imgcell{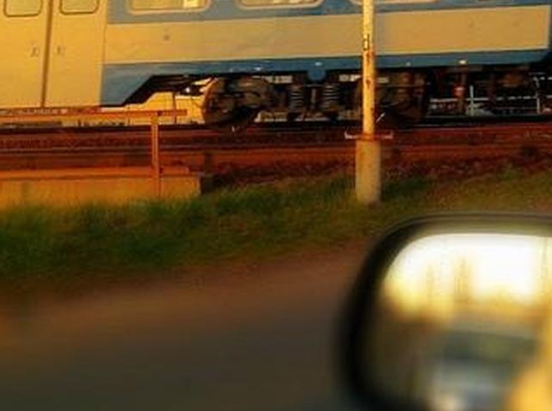} &
\imgcell{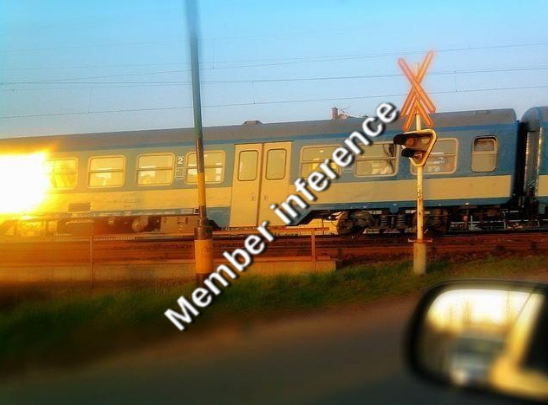} &
\imgcell{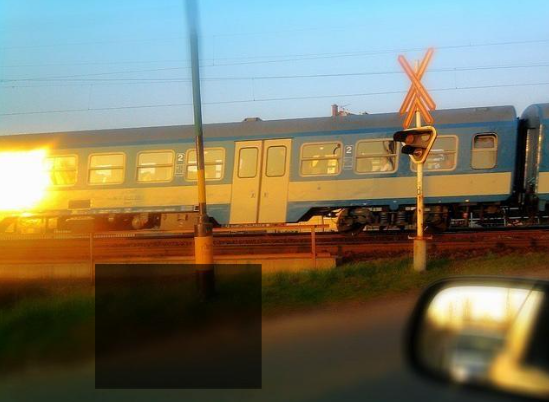} \\
\textbf{Severe} &
\imgcell{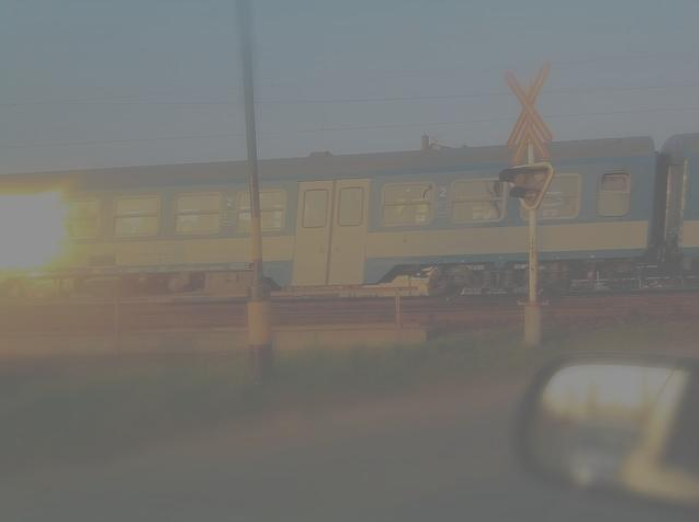} &
\imgcell{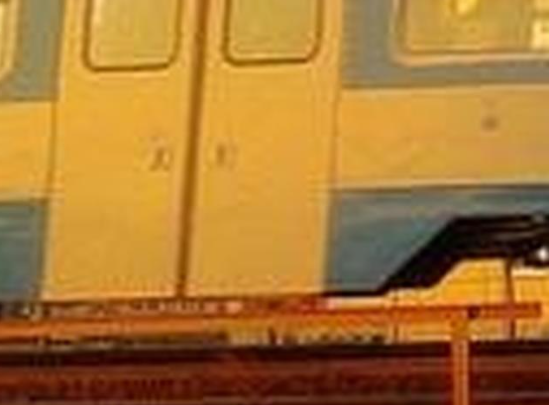} &
\imgcell{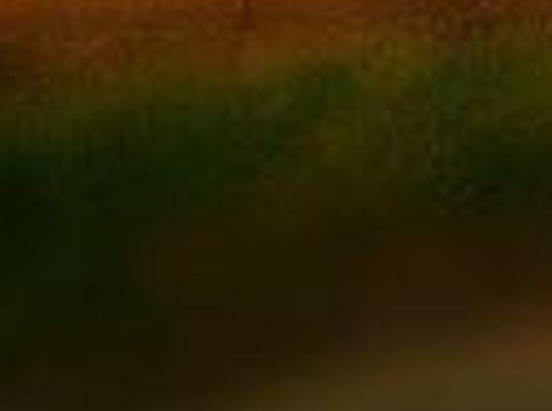} &
\imgcell{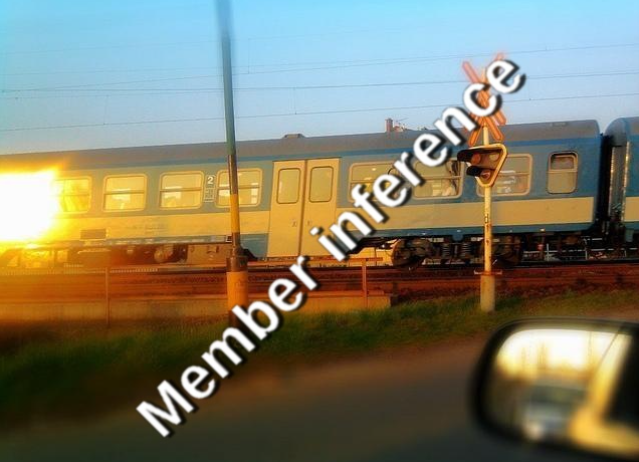} &
\imgcell{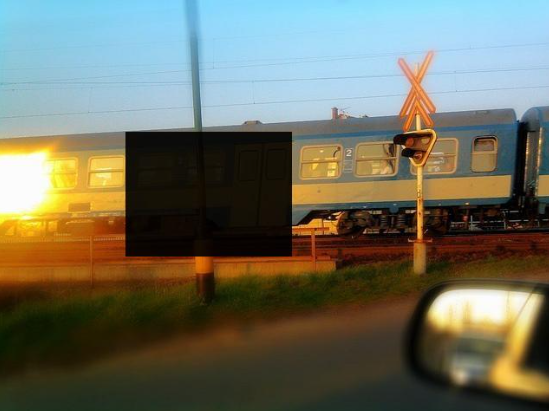} \\
\bottomrule
\end{tabular}
}

\end{table*}

\end{document}